\documentclass[10pt,journal,compsoc]{IEEEtran}
\ifCLASSOPTIONcompsoc
\usepackage[nocompress]{cite}
\else
\usepackage{cite}
\fi
\ifCLASSINFOpdf
\else
\fi
\usepackage{array}
\usepackage{microtype}
\usepackage{graphicx}
\usepackage{bm}
\usepackage{algorithm}
\usepackage{algorithmic}
\usepackage{multirow}
\usepackage{amsthm}
\usepackage{amsmath}
\usepackage{subfigure}
\usepackage{amssymb}
\usepackage{bbding}
\usepackage{pifont}
\usepackage{wasysym}
\usepackage{booktabs} % for professional tables
\usepackage{makecell}
\usepackage{subfigure}
\usepackage{bbding}
\usepackage{balance}
\usepackage{float}
\usepackage{subeqnarray}
\usepackage{cases}
\usepackage{tabularx}
\usepackage{color}
\usepackage{colortbl,url}
\usepackage{threeparttable}
\newcommand{\cmark}{\ding{51}}%
\newcommand{\xmark}{\ding{55}}%

\usepackage{xcolor}

\renewcommand{\arraystretch}{1.2}

\newtheorem{problem}{Problem}

\begin{document}
	
	\title{Learning with Constraint Learning:\\ New Perspective, Solution Strategy and Various Applications}
	
	\author{Risheng~Liu,~\IEEEmembership{Member,~IEEE,}
		Jiaxin~Gao, 
		Xuan~Liu,
		and~Xin~Fan,~\IEEEmembership{Senior Member,~IEEE}
		%Deyu~Meng,~\IEEEmembership{Member,~IEEE,}
		%and~Zhouchen~Lin,~\IEEEmembership{Fellow,~IEEE}% <-this % stops a space
		\IEEEcompsocitemizethanks{\IEEEcompsocthanksitem  R. Liu is with DUT-RU International School of Information Science \& Engineering, Dalian University of Technology, and also with the Key Laboratory for Ubiquitous Network and Service Software of Liaoning Province, Dalian, Liaoning, P.R., China. R. Liu is also with the Peng Cheng Laboratory, Shenzhen, Guangdong, P.R. China. (Corresponding author, e-mail: rsliu@dlut.edu.cn).
			%R. Liu and J. Gao are with the DUT-RU International School of Information Science \& Engineering, Dalian University of Technology, and also with the Key Laboratory for Ubiquitous Network and Service Software of Liaoning Province, Dalian 116024, China.
			%E-mail: rsliu@dlut.edu.cn, jiaxinn.gao@outlook.com. R. Liu is the corresponding author. \protect  
%			\IEEEcompsocthanksitem J. Zhang is with the Department of Mathematics, Southern University of Science and Technology, and National Center for Applied Mathematics Shenzhen, China, E-mail: zhangj9@sustech.edu.cn.\protect
%			
\IEEEcompsocthanksitem J. Gao and X. Liu are with DUT-RU International School of Information Science \& Engineering, School of Software Technology, Dalian University of Technology, Dalian, 116024, China. (e-mail: jiaxinn.gao@outlook.com, liuxuan\_16@126.com). \protect

\IEEEcompsocthanksitem X. Fan is with the DUT-RU International School of Information Science \& Engineering, Dalian University of Technology, Dalian, 116024, China. (email: xin.fan@dlut.edu.cn). \protect
%			\IEEEcompsocthanksitem Z. Lin is with the Key Laboratory of Machine Perception (Ministry of Education), School of Electronics Engineering and Computer Science, Peking University, Beijing 100871, China, and also with the Cooperative Medianet Innovation Center, Shanghai Jiao Tong University, Shanghai 200240, China. E-mail: zlin@pku.edu.cn.\protect 
		} 
		\thanks{Manuscript received April 19, 2005; revised August 26, 2015.}}
	

	\markboth{Journal of \LaTeX\ Class Files,~Vol.~14, No.~8, August~2015}%
	{Shell \MakeLowercase{\textit{et al.}}: Bare Demo of IEEEtran.cls for Computer Society Journals}
	
	\IEEEtitleabstractindextext{%
		\begin{abstract}

The complexity of learning problems, such as Generative Adversarial Network (GAN) and its variants, multi-task and meta-learning, hyper-parameter learning, and a variety of real-world vision applications, demands a deeper understanding of their underlying coupling mechanisms. Existing approaches often address these problems in isolation, lacking a unified perspective that can reveal commonalities and enable effective solutions. Therefore, in this work, we proposed a new framework, named Learning with Constraint Learning (LwCL), that can holistically examine challenges and provide a unified methodology to tackle all the above-mentioned complex learning and vision problems. Specifically, LwCL is designed as a general hierarchical optimization model that captures the essence of these diverse learning and vision problems. Furthermore, we develop a gradient-response based fast solution strategy to overcome optimization challenges of the LwCL framework. Our proposed framework efficiently addresses a wide range of applications in learning and vision, encompassing three categories and nine different problem types. Extensive experiments on synthetic tasks and real-world applications verify the effectiveness of our approach. The LwCL framework offers a comprehensive solution for tackling complex machine learning and computer vision problems, bridging the gap between theory and practice.			
		\end{abstract}

		}

	\maketitle
	
	\IEEEdisplaynontitleabstractindextext
	
	\IEEEpeerreviewmaketitle
	
	\IEEEraisesectionheading{\section{Introduction}\label{sec:introduction}}

	\IEEEPARstart{I}{n} recent years, a plethora of endeavors have emerged to tackle contemporary intricate problems such as GAN and its variants~\cite{goodfellow2014generative,zhong2019adversarial}, multi-task and meta learning~\cite{rajeswaran2019meta,li2017meta}, hyper-parameter learning~\cite{franceschi2017forward,shaban2019truncated}, and various challenging real-world vision applications~\cite{DBLP:conf/miccai/CicekALBR16,zhang2021plug}. 	
	In contrast to these conventional learning paradigms that solely focus on a single learning objective (e.g., classification and regression), these aforementioned modern complex problems often necessitate the simultaneous handling of multiple interrelated learning tasks. For instance, the  well-known generative adversarial networks often require discriminatorassisted branches in the process of adversarial game~\cite{arjovsky2017wasserstein,zhu2017unpaired}. Similarly,  multi-task and meta learning introduce task-specific classifiers as supportive sub-tasks to facilitate the acquisition of enhanced generalization representations of meta-features~\cite{collins2020task,NEURIPS2020_84c578f2}. Hyper-parameter learning involves the construction of simple classifiers as interconnected tasks, aiding the base model in attaining optimal hyper-parameters~\cite{pedregosa2016hyperparameter,lorraine2020optimizing}. 	
	Despite the presence of diverse motivations and mechanisms, all these issues encounter the challenge of simultaneously addressing multiple interrelated tasks with coupled structures. This hierarchical coupling induces complexity in the learning process and constitutes the fundamental factor exacerbating the difficulties encountered in problem-solving.

While considerable advancements have been achieved, the present state-of-the-art techniques employed to tackle these modern complex problems still encounter numerous challenges. On the one hand, certain approaches, which typically involve constructing task-specific methodologies tailored to specific scenarios, heavily rely on extensive models and datasets. Often, these task-specific methods exhibit limited transferability to other tasks, resulting in subpar generalization capabilities. On the other hand, accurately characterizing the coupling relationships between the primary task and multiple related learning tasks proves challenging due to the empirical and trial-and-error nature of most learning strategy designs. For instance, simplistic alternating iterative learning strategies that optimize one aspect while keeping another fixed overlook the potential coupling constraints and dynamic game states among the multiple interdependent tasks. Therefore, it becomes imperative to integrate diverse modeling approaches into a unified framework and explore the inherent connections among multiple tasks.

In the following, this paper endeavors to establish a unified and coherent optimization perspective that explores the intrinsic relationships of these modern complex problems, considering their potential coupling. Termed as Learning with Constraint Learning (LwCL) in this paper, this perspective offers a comprehensive framework for understanding and formulating these problems. 
Essentially, the LwCL problem can be formulated as follows:

\begin{problem}

\textbf{Learning with Constraint Learning (LwCL)} represents an innovative learning mechanism, distinguished by a hierarchical arrangement of two interconnected learning tasks. Within LwCL, the fulfillment of the primary objective task (referred to as the Objective Learner or OL) relies upon the successful completion of a lower-level learning task (referred to as the Constraint Learner or CL)\footnote{The detailed concepts and applications of this framework will be presented in Sections~\ref{sec:sec2} and~\ref{sec:sec3}, respectively.}. This hierarchical structure endows the learning process with added depth, as the CL acts as a constraint that must be satisfied, effectively guiding and shaping the optimization process towards the attainment of the overarching objective. Through this nested hierarchy, LwCL enhances the learning process, fostering a more structured and directed approach to achieving the desired learning outcome.
 \label{pro:pro}
\end{problem} 

In essence, LwCL embodies a nested hierarchy of learning tasks, where each subtask contributes to the accomplishment of the overarching objective. This intricate nested framework adds complexity to the learning process and requires a more sophisticated approach to problem-solving. It can be challenging because the optimization process must balance the competing objectives of completing the subtasks at each layer while also optimizing the overall objective of the entire system. Nevertheless, by leveraging the hierarchical structure of the problem, LwCL can improve performance on complex tasks and enable efficient transfer learning.

Based on the above analysis, the primary objective of this paper is to present a unified perspective, termed LwCL, which aims to reinterpret and elucidate the underlying mechanisms of modern complex problems. 
Building upon this foundation, we have developed a generic hierarchical optimization framework, encompassing reformulation and algorithmic components, to unveil the potential coupling constraints among multiple tasks. Additionally, leveraging the concept of dynamic best response, we have employed an outer-product-based Hessian approximation technique to devise a rapid solving strategy from the standpoint of implicit gradients. This approach enables accurate tracing of the gradient feedback dynamics between the OL and the CL, thereby yielding unprecedented advancements in training stability and performance. Importantly, our proposed method exhibits remarkable flexibility and adaptability, as it can be seamlessly integrated into a diverse range of contemporary complex learning problems, owing to the inherent tolerance of the constraint learning paradigm towards the requirements of the objective function. We also demonstrate that our proposed framework can efficiently address a wide range of LwCL applications in the fields of learning and vision, including three categories of problems, with a total of nine different types. 

Our contributions can be summarized as follows:
\begin{itemize}

	\item From a comprehensive and in-depth  point of view, we introduce a unified perspective, termed as Learning with Constraint Learning (LwCL), to analyze, reformulate, and address a wide array of complex learning problems that exhibit underlying coupled relationships in the domains of machine learning and computer vision. 
	
	\item 

	We propose a hierarchical optimization framework that effectively formulates the potential dependencies and uncovers the inherent coupling among multiple tasks within LwCLs. This framework facilitates precise optimization of the two learning tasks through a synergistic and interactive approach, incorporating the proposed gradient-response feedback.
	\item To alleviate high computational complexity issues  associated with naive learning strategy, we design an implicit gradient scheme with outer-product Hessian approximation as fast solution strategy to efficiently solve the nested optimization process, which is more computation-friendly and suitable for diverse high-dimensional large-scale real-world applications. 
	%\item 
	\item We demonstrate that LwCL can efficiently address a wide range of modern complex learning and vision applications, including three categories of problems, with a total of nine different types. The versatility and effectiveness of our proposed LwCL framework is verified through extensive experiments on both synthetic tasks and real-world applications.
\end{itemize}

	\section{Review of Related Works within LwCL}

	Based on Problem \ref{pro:pro}, we now proceed to comprehensively understand and (re)formulate existing modern works from the unified perspective of LwCL. Specifically, we categorize these works into three classes, including Adversarial Learning (AL), Auxiliary with Related Tasks (ART), and Task Divide and Conquer (TDC), utilizing the lens of LwCL. 
%	Tab.~\ref{tab:relatedworks} provides a summary of the relevant works, indicating their corresponding OL and CL tasks.
	
	\textbf{AL-type Applications.}
	As one of the most popular LwCTs, AL-based methods exhibit a
	strong ability to model specific data distribution by addressing the assisted discriminative tasks via a dynamic adversarial game. OL and CL can be regarded as generator learning and discriminator learning, respectively.	
	For example, vanilla GAN~\cite{goodfellow2014generative},  as a discriminator-assisted learning problem, can be interpreted as a dynamic adversarial game that greedily finds the solution of the minmax formula through an alternating iterative strategy. 	
	Metz \textit{et al.}~\cite{metz2016unrolled} proposed the idea that gradients can be back-propagated through the unrolled optimization procedure in a differentiable way to address the challenges of unstable optimization. 
	%Farnia et al.~\cite{farnia2020gans} introduced the concept of proximal equilibrium as a solution concept for zero-sum games and proposes proximal training paradigm.  
	Accordingly, various variants of loss types and regulation (i.e., second-order gradient loss~\cite{metz2016unrolled}, Hinge loss~\cite{brock2018large}, and Lipschitz penalty~\cite{petzka2017regularization}) also appear in the optimization objectives of GAN-variants. Nonetheless, these methods still suffer from poor generation quality, training oscillations, and other challenges that have not been fully addressed.  
	In addition to the narrowly defined GANs and their variants, a series of adversarial vision tasks in cutting-edge areas,  such as image generation~\cite{arjovsky2017wasserstein,farnia2020gans,gao2018deep,engel2017latent}, style transfer~\cite{zhu2017unpaired,azadi2018multi} and imitation learning~\cite{fu2017learning,arulkumaran2021pragmatic} have risen in recent years. These problems employ different task-specific losses and strategies to mitigate the reconstruction discrepancy between different image domains in training process. 
	For example, Zhu \textit{et al.}~\cite{jiang2021enlightengan} constructed bi-directional cycle generative-discriminative architecture and cycle consistency loss to complete cross-domain style transfer.
	In reinforcement learning, Pfau \textit{et al.}~\cite{pfau2016connecting} treated the discriminator as a regression task providing scalar values rather than a binary classifier and opened up new avenues of research by treating adversarial learning programs as an actor-critic problem. 	
	Despite the good intentions of these approaches, existing AL-based strategies still suffer from several major challenges, such as training instability, oscillations and mode collapse.
	% due to the highly complex training mechanisms caused by the introduced adversarial tasks. 
	The underlying reason is the inability of learning mechanisms that rely on alternating iterations to accurately portray the intrinsically complex dynamics between the considered task and the introduced adversarial task. Therefore, we proceed to present new mathematical tools to model and solve such AL-type problems.

	\textbf{ART-type Applications.} 
	In recent decades, a category of typical learning tasks towards sophisticated applications have addressed considered learning tasks with related auxiliary learning devices, named auxiliary with related tasks, such as medical image analysis (i.e., medical image registration and segmentation~\cite{DBLP:conf/miccai/XuN19,he2022learning,DBLP:conf/wacv/TomarBLVRT22,DBLP:conf/cvpr/ZhaoBDGD19} and low-light image enhancement~\cite{liu2022learning,xue2022best}) and hyper-parameter learning~\cite{franceschi2017forward,liu2021towards,pedregosa2016hyperparameter,lorraine2020optimizing,liu2021general}.   OL and CL can be regarded as objective learning task and auxiliary learning task, respectively.
	For example, in the spirit that medical image registration can provide more label information for one-shot image segmentation, Xu \textit{et al.}~\cite{DBLP:conf/miccai/CicekALBR16} developed a joint model for simultaneous image registration and segmentation. For low-light scenes, Xue \textit{et al.}~\cite{xue2022best} proposed to introduce additional detection and segmentation models to assist the low-light enhancement task. Actually, these approaches often generally rely on naive empirical strategies (i,e., alternate learning) to solve the problem and often suffer from disadvantages such as low training efficiency, low performance, and poor generalization. Similarly, in order to assist the base model for obtaining optimal hyper-parameters, hyper-parameter learning usually introduces simple classifiers that contain only a few fully connected layers as auxiliary tasks, with first-order gradient based algorithms~\cite{franceschi2017forward,liu2021towards}.  However, such methods possess a high algorithmic complexity and are usually limited to low-dimensional data scenarios. Overall, an in-depth exploration focusing on how to uniformly and efficiently  formulate and address these ART-type tasks is essential and necessary.

	\textbf{TDC-type Applications.}
	There is another class of learning task construction ideas, i.e., dividing a complete learning process into multiple subtask learning processes, called task divide and conquer,  with typical applications such as  image deblur~\cite{zhang2021plug,zhang2017learning} and multi-task meta learning~\cite{collins2020task,rajeswaran2019meta,collins2020task,li2017meta,nichol2018first,shaban2019truncated}.  OL and CL can be regarded as  meta-feature/prior learning and task-specific classifier/fidelity learning, respectively.	
	For example, meta-feature learning methods~\cite{rajeswaran2019meta,li2017meta} generally separated the network structure into meta-feature extraction modules and task-specific modules. The latter guides better learning of generalization representations of meta-features by constructing losses for multiple different tasks. Admittedly, the above methods are usually confined to small-scale, low-dimensional simulation scenarios, and still suffer from various unsolved challenges such as training instability and computational inefficiency for real-world high-dimensional applications. For image deblur task, Zhang \textit{et al.}~\cite{franceschi2017bridge} proposed semi-quadratic split-based deep unrolling method to enhance deblurred images,  divided into the fidelity learning and a prior learning subproblems. Among them, the prior learning task introduce a plug-and-play denoiser. Unfortunately, the fixed pre-trained denoiser, lacks generalization applicability to various complex scenarios. 
    In the subsequent sections, we will develop a deeper understanding, modeling and solving such TDC-type problems from the LwCL perspective.

	\begin{figure*}[tbp]
		\centering 	
		\begin{tabular}{c}
			\includegraphics[width=18cm]{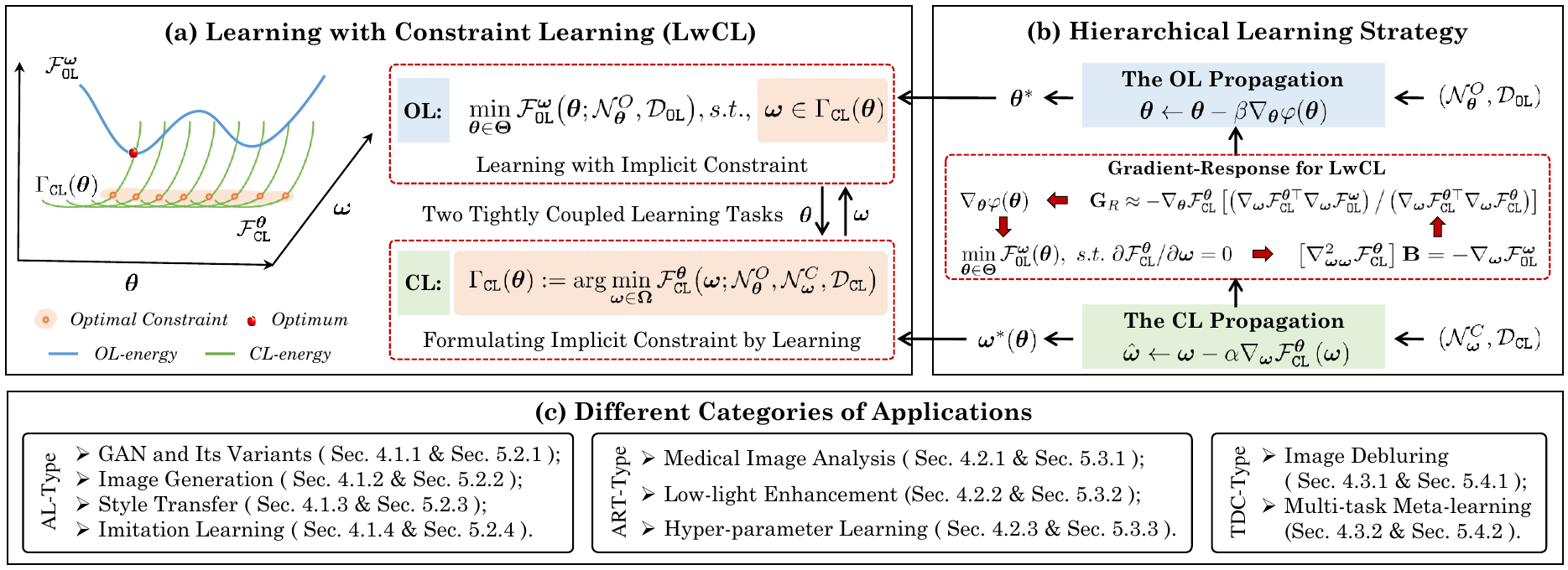}\\
		\end{tabular}%
		\caption{Overview of the proposed method. In (a), we present a novel LwCL perspective to comprehensively investigate and address these contemporary complex problems with underlying coupling relationships in a unified manner. To effectively resolve the nested optimization process, (b) introduces a specially designed dynamic Gradient-Response solution strategy tailored for LwCL. Lastly, (c) provides a comprehensive enumeration of nine major problem categories spanning three broad application areas that our method can effectively tackle. 
		Please notice that the diagram in (a) is just a simplified illustration for our LwCL mechanism. However, in real applications, both OL and CL tasks will pose significantly higher complexity and challenges, such as non-convex learning problems.
		}\label{fig:fig1}
		\vspace{-3mm}
	\end{figure*}% 

	\section{Learning with Constraint Learning} \label{sec:sec2}   
	In this section, we endeavor to establish a comprehensive hierarchical optimization framework and a dynamic best response-based fast solution strategy that enable a unified formulation and resolution of various types of LwCL problems. 	
	The overall framework is depicted in Fig.~\ref{fig:fig1}. 	
	In (a), we introduce a novel perspective to comprehensively investigate and address contemporary complex problems (i.e., AL, ART and TDC) in a unified manner. To effectively handle the nested optimization process, (b) illustrates a comprehensive hierarchical optimization framework that aims to redefine and reformulate LwCLs, and proposes a specially designed  gradient-response solution strategy tailored specifically for LwCL (as denoted by the dashed rectangle). Finally, (c) provides a comprehensive enumeration of nine major problem categories across three broad application areas that our method can effectively tackle.

	\subsection{Hierarchical Optimization Formulation for LwCL}  
  Our investigation is centered around contemporary and intricate LwCL problems, as delineated in Problem~\ref{pro:pro}. Specifically, we regard that the essence of LwCL lies in the construction of an OL, denoted as $\mathcal{N}_{\bm{\theta}}^{O}$ with parameterization $\bm{\theta}$, which aims to optimize the performance of the desired objective learning task, such as generator learning or meta-feature learning. This endeavor can be expressed as an optimization problem with respect to $\bm{\theta}$, whereby the OL energy $\mathcal{F}_{\mathtt{OL}}$ encapsulates the optimization objective base on the OL dataset $\mathcal{D}_{\mathtt{OL}}$. 
  
  Given the inherent complexities in directly solving the OL, it is customary to introduce multiple interrelated learning tasks in the form of auxiliary CL, denoted as $\mathcal{N}_{\bm{\omega}}^{C}$ with parameterization $\bm{\omega}$, to assist the overarching objective of OL. 
  To elucidate the aforementioned notion, we generally introduce the more abstract formulation of energy-constrained learning, which is expressed as 
%     \begin{equation} \label{eq:up} 
%\min\limits_{\bm{\theta}\in\bm{\Theta}}\mathcal{F}_{\mathtt{OL}}\big[\bm{\theta},\bm{\omega}(\bm{\theta});\mathcal{N}_{\bm{\theta}}^{O},\mathcal{N}_{\bm{\omega}}^{C},\mathcal{D}_{\mathtt{OL}}\big], s.t.,~
%  		\bm{\omega}(\bm{\theta}) \in \Gamma_{\mathtt{CL}}(\bm{\theta}), 
%  \end{equation}
     \begin{equation} \label{eq:up} 
	\min\limits_{\bm{\theta}\in\bm{\Theta}}\mathcal{F}_{\mathtt{OL}}^{\bm{\omega}}\big(\bm{\theta};\mathcal{N}_{\bm{\theta}}^{O},\mathcal{D}_{\mathtt{OL}}\big), s.t.,~
	\bm{\omega} \in \Gamma_{\mathtt{CL}}(\bm{\theta}), 
\end{equation}
where $\Gamma_{\mathtt{CL}}(\bm{\theta})$ denotes the optimal constraint with respect to $\bm{\omega}$. 
  Due to the tight coupling and potential dependency between the two variables, combined with the difficulty in accurately determining $\Gamma_{\mathtt{CL}}(\bm{\theta})$, the aforementioned problem in Eq.~\eqref{eq:up} becomes extremely complex.
  
  Given this, we introduce a learning modeling approach with constraints that characterizes the optimization process concerning the variable $\bm{\omega}$. This auxiliary learning task can likewise be framed as an optimization problem, wherein the CL energy $\mathcal{F}^{\bm{\theta}}_{\mathtt{CL}}$ characterizes the performance of CL with variable $\bm{\omega}$,  formulated as 
  \begin{equation} \label{eq:ll} 
  	\Gamma_{\mathtt{CL}}(\bm{\theta}):=\arg\min\limits_{\bm{\omega}\in\bm{\Omega}}\mathcal{F}^{\bm{\theta}}_{\mathtt{CL}}\big(\bm{\omega};\mathcal{N}_{\bm{\theta}}^{O},\mathcal{N}_{\bm{\omega}}^{C},\mathcal{D}_{\mathtt{CL}}\big).
  \end{equation}    
  By combining Eqs.~\eqref{eq:up} and \eqref{eq:ll}, we observe that, the intrinsic hierarchical relationship between the two learners (i.e., $\mathcal{N}_{\bm{\theta}}^{O}$ and $\mathcal{N}_{\bm{\omega}}^{C}$) is explicitly encoded by the task-specific energy functions (i.e., $\mathcal{F}^{\bm{\omega}}_{\mathtt{OL}}$ and $\mathcal{F}^{\bm{\theta}}_{\mathtt{CL}}$)\footnote{The detailed design of $\mathcal{F}^{\bm{\omega}}_{\mathtt{OL}}$ and $\mathcal{F}^{\bm{\theta}}_{\mathtt{CL}}$ in specific applications will be presented in Section~\ref{sec:sec3}.}. Furthermore, it is worth noting that the energy functions $\mathcal{F}^{\bm{\omega}}_{\mathtt{OL}} (or~\mathcal{F}^{\bm{\theta}}_{\mathtt{CL}}):\mathbb{R}^m\times\mathbb{R}^n\to\mathbb{R}$ are continuous, while the set $\bm{\Omega}\subseteq\mathbb{R}^n$ represents a nonempty feasible region, and $\bm{\Theta}\subseteq\mathbb{R}^m$ denotes the feasible set for the variables. In this context, we refer to $\bm{\theta}$ and $\bm{\omega}$ as the outer-level (or OL-level) variables and $\mathcal{F}^{\bm{\omega}}_{\mathtt{OL}}$ and $\mathcal{F}^{\bm{\theta}}_{\mathtt{CL}}$ as the outer-level (or OL-level) energy and inner-level (or CL-level) energy, respectively.   
%  Drawing from these insights, we posit that LwCL can be abstracted as comprising two hierarchical levels of learning tasks, imbued with an embedded structure, wherein OL serves as the primary task while CL operates as a nested constraint.  
  %Drawing on the idea of Stackelberg game theory~\cite{dempe2020bilevel}, we attempt to provide a powerful hierarchical optimization framework as a new mathematical tool to explicitly inscribe the coupling dependencies of two players (i.e., OL and CL).
  Building upon the principles of Stackelberg game theory~\cite{dempe2020bilevel}, we strive to present a robust hierarchical optimization framework, serving as a novel mathematical instrument to explicitly inscribe the coupling dependencies of two key players.

Inherently, a notable asymmetry is observed between the two levels of learning tasks, where $\mathcal{F}^{\bm{\theta}}_{\mathtt{CL}}$ assumes the role of a constraint upon $\mathcal{F}^{\bm{\omega}}_{\mathtt{OL}}$, facilitating the derivation of optimal feedback $\bm{\omega}(\bm{\theta})$ to be passed onto the core optimization objective. This dynamic constrained learning process entails a high degree of interdependence between the variables $\bm{\theta}$ and $\bm{\omega}$, such that every incremental update of $\bm{\theta}$ is inevitably influenced by the state of $\bm{\omega}$. Notably, this framework can also be interpreted as a more encompassing bilevel optimization problem. To address practical applications in high-dimensional real-world scenarios, we proceed to propose a rapid and efficient solution strategy characterized by dynamic best response.

   \subsection{Solution Strategy with Gradient-Response} 
   %Moving one step forward, 
   Commencing from the  dynamic gradient-response, we further define the value-function $\varphi(\bm{\theta})$, leading us to the minimization problem, $
   \min\limits_{\bm{\theta}} \varphi(\bm{\theta}):=\mathcal{F}^{\bm{\omega}}_{\mathtt{OL}}(\bm{\theta},\bm{\omega}(\bm{\theta})).$ 	
    Progressing further, the key to resolving this problem lies in computing the gradient of the OL optimization objective: 
	%\begin{small}	
	\begin{equation}
	\nabla_{\bm{\theta}} \varphi(\bm{\theta})=\nabla_{\bm{\theta}} \mathcal{F}^{\bm{\omega}}_{\mathtt{OL}}(\bm{\theta},\bm{\omega}(\bm{\theta})) +   \underbrace{\nabla_{\bm{\omega}}\mathcal{F}^{\bm{\omega}}_{\mathtt{OL}}(\bm{\theta},\bm{\omega}(\bm{\theta}))\nabla_{\bm{\theta}}\bm{\omega}(\bm{\theta})}_{\mathbf{G}_R}. \label{eq:bilevel-gradient0}
	\end{equation}
	%\end{small}
	Notably, the direct gradient term $\nabla_{\bm{\theta}} \mathcal{F}^{\bm{\omega}}_{\mathtt{OL}}(\bm{\theta},\bm{\omega})$ showcases a straightforward reliance on the OL variable $\bm{\theta}$ and can be readily obtained through simple computations in practice. Conversely, the second gradient term, denoted as $\mathbf{G}_R$ poses challenges in its calculation due to the varying rate of $\bm{\omega}(\bm{\theta})$ with respect to $\bm{\theta}$.  
	Nonetheless, $\mathbf{G}_R$ accurately captures the best response gradient (in connection with $\nabla_{\bm{\omega}} \mathcal{F}^{\bm{\theta}}_{\mathtt{CL}}$ and $\nabla_{\bm{\theta}} \mathcal{F}^{\bm{\theta}}_{\mathtt{CL}}$) between the two learning tasks and assumes a pivotal role in optimizing LwCLs. Essentially, equipped with Eq.~\eqref{eq:bilevel-gradient0}, the gradient of $\mathcal{N}_{\bm{\omega}}^{C}$ can be dynamically and accurately back-propagated to $\mathcal{N}_{\bm{\theta}}^{O}$ at each iteration, effectively aiding in the optimization of its parameters.
		
	Undoubtedly, evaluating the exact best response gradient of Eq.~\eqref{eq:bilevel-gradient0} poses significant computational challenges for most existing strategies, particularly when the dimensions of $\bm{\omega}$ and $\bm{\theta}$ are high. To address this challenge, we leverage implicit methods, which offer a direct and precise estimation of the optimal gradient. Inspired by the implicit function theorem, we derive the following equation based on the best response gradient, i.e., $ \partial \mathcal{F}^{\bm{\theta}}_{\mathtt{CL}}/ \partial\bm{\omega}=0$.  In contrast to the solution strategy, the best response  gradient is then substituted with an implicit equation, wherein $\nabla_{\bm{\theta}} \bm{\omega}(\bm{\theta})$ is derived as: 
	\begin{equation}
	\nabla_{\bm{\theta}} \bm{\omega}(\bm{\theta})
	= -\left[ \nabla_{\bm{\omega}}^2 \mathcal{F}^{\bm{\theta}}_{\mathtt{CL}}(\bm{\theta},\bm{\omega}(\bm{\theta}))  \right] ^{-1}
	\nabla_{\bm{\omega}\bm{\theta}}^2 \mathcal{F}^{\bm{\theta}}_{\mathtt{CL}}(\bm{\theta},\bm{\omega}(\bm{\theta})). \label{eq:ift0}
	\end{equation} 	
	Considering the formidable challenges associated with computing the Hessian and its inverse, we propose a fast solution strategy by simplifying the second derivative to the first derivative, enabling the calculation of the best response Jacobian. This strategy involves two key computational steps: implicit gradient estimation and outer-product approximation.
	
	\textbf{Implicit Gradient Estimation.} 
	To circumvent the direct calculation of multiple Hessian products and their inversions, we introduce a linear solver system $\mathbf{B}$ based on Eq.~\eqref{eq:ift0}, allowing us to avoid the complexity associated with computing $\mathbf{G}_{R}$. Consequently, the indirect response gradient can be reformulated as:
	\begin{equation}
	\mathbf{G}_{R}=\left[\nabla_{\bm{\omega}\bm{\theta}}^2 \mathcal{F}^{\bm{\theta}}_{\mathtt{CL}}  \right]^{\top}\mathbf{B}, \ \mbox{where} \ \left[\nabla_{\bm{\omega}\bm{\omega}}^2 \mathcal{F}^{\bm{\theta}}_{\mathtt{CL}} \right]\mathbf{B}=- \nabla_{\bm{\omega}} \mathcal{F}^{\bm{\omega}}_{\mathtt{OL}},\label{eq:adjoint-equation0}
	\end{equation}
	where $(\cdot)^{\top}$ denotes the transposition operation. Through this formulation, $\mathbf{G}_{R}$ is solely dependent on the first-order condition, effectively decoupling the computational burden from the solution trajectory of the constrained sub-task. This decoupling greatly alleviates the pressure of propagating the backward gradient in the constrained dynamics.
	%Nonetheless, it can be recognized that the calculation of second-order derivatives in $\mathbf{G}_{R}$ is still intractable. 
	%The urgent requirements from  approximating the repeated computation of two Hessian matrix $\nabla_{\bm{\omega}\bm{\omega}}^2 \mathcal{F}_{\mathtt{CL}}$ and  $\nabla_{\bm{\omega}\bm{\theta}}^2 \mathcal{F}_{\mathtt{CL}}$  spawned the following concept of outer-product approximation.
	
	However, it is worth noting that the calculation of second-order derivatives in $\mathbf{G}_{R}$ remains intractable. The pressing need to approximate the repeated computation of two Hessian matrices, $\nabla_{\bm{\omega}\bm{\omega}}^2 \mathcal{F}^{\bm{\theta}}_{\mathtt{CL}}$ and  $\nabla_{\bm{\omega}\bm{\theta}}^2 \mathcal{F}^{\bm{\theta}}_{\mathtt{CL}}$, has led to the emergence of the concept of outer-product approximation.
	
		\textbf{Outer-Product Approximation.} To further suppress the complexity of constrained optimization, we consider replacing the original Hessian operation with the Gauss-Newton strategy and introduce two corresponding outer products, as follows:
	\begin{equation}
	\nabla_{\bm{\omega}\bm{\omega}}^2 \mathcal{F}^{\bm{\theta}}_{\mathtt{CL}} \approx \nabla_{\bm{\omega}}  \mathcal{F}^{\bm{\theta}}_{\mathtt{CL}} \nabla_{\bm{\omega}}  \mathcal{F}_{\mathtt{CL}}^{\bm{\theta}\top},\
	\nabla_{\bm{\omega}\bm{\theta}}^2 \mathcal{F}^{\bm{\theta}}_{\mathtt{CL}}  \approx \nabla_{\bm{\omega}} \mathcal{F}^{\bm{\theta}}_{\mathtt{CL}} \nabla_{\bm{\theta}} \mathcal{F}_{\mathtt{CL}}^{\bm{\theta}\top}.\label{eq:outer-product0}
	\end{equation} 
	By separating the gradient, this approach converts the highly complex second-order derivative into a simple product operation involving the first-order derivative, which significantly reduces the algorithm's complexity, especially in terms of memory consumption. 
	By combining Eqs.~\eqref{eq:adjoint-equation0}-\eqref{eq:outer-product0}, we establish the nonlinear least squares problem by approximating the Gauss-Newton formula. Plugging into the Eq.~\eqref{eq:adjoint-equation0}, thus we can obtain  
%	\begin{small}
%	\begin{equation*}
%	\left[\left(\nabla_{\bm{\omega}} \mathcal{F}_{\mathtt{CL}} \nabla_{\bm{\omega}} \mathcal{F}_{\mathtt{CL}}^{\top}\right)^{\top}\nabla_{\bm{\omega}} \mathcal{F}_{\mathtt{CL}} \nabla_{\bm{\omega}} \mathcal{F}_{\mathtt{CL}}^{\top} \right]\mathbf{B}\approx -\left(\nabla_{\bm{\omega}} \mathcal{F}_{\mathtt{CL}} \nabla_{\bm{\omega}} \mathcal{F}_{\mathtt{CL}}^{\top} \right)^{\top} \nabla_{\bm{\omega}} \mathcal{F}^{\bm{\omega}}_{\mathtt{OL}}.
%	\end{equation*} 
%	\end{small}  
\begin{equation}
	\left(\nabla_{\bm{\omega}} \mathcal{F}^{\bm{\theta}}_{\mathtt{CL}} \nabla_{\bm{\omega}} \mathcal{F}_{\mathtt{CL}}^{\bm{\theta}\top} \right)^2 \mathbf{B} \approx -\left(\nabla_{\bm{\omega}} \mathcal{F}^{\bm{\theta}}_{\mathtt{CL}} \nabla_{\bm{\omega}} \mathcal{F}_{\mathtt{CL}}^{\bm{\theta}\top} \right)^{\top} \nabla_{\bm{\omega}} \mathcal{F}^{\bm{\omega}}_{\mathtt{OL}}.
\end{equation}
	To over simplify, we can express $\mathbf{B}$ with $\mathbf{M}$ as follows $\mathbf{B}=\mathbf{M} \left(\nabla_{\bm{\omega}} \mathcal{F}_{\mathtt{CL}}^{\bm{\theta}\top} \nabla_{\bm{\omega}} \mathcal{F}^{\bm{\theta}}_{\mathtt{CL}} \right)\nabla_{\bm{\omega}} \mathcal{F}^{\bm{\omega}}_{\mathtt{OL}},$  
%	\begin{equation*}
%	\mathbf{B}=\mathbf{M} \left(\nabla_{\bm{\omega}} \mathcal{F}_{\mathtt{CL}}^{\bm{\theta}\top} \nabla_{\bm{\omega}} \mathcal{F}^{\bm{\theta}}_{\mathtt{CL}} \right)\nabla_{\bm{\omega}} \mathcal{F}^{\bm{\omega}}_{\mathtt{OL}}, 
%	\end{equation*} 
    where $\mathbf{M}=\left[\nabla_{\bm{\omega}} \mathcal{F}^{\bm{\theta}}_{\mathtt{OL}} \left(\nabla_{\bm{\omega}} \mathcal{F}_{\mathtt{CL}}^{\bm{\theta}\top} \nabla_{\bm{\omega}} \mathcal{F}^{\bm{\theta}}_{\mathtt{CL}}\right) \nabla_{\bm{\omega}} \mathcal{F}_{\mathtt{CL}}^{\bm{\theta}\top} \right]^{-1}.$ 
	Ultimately, we obtain an approximate representation of the response gradient $\mathbf{G}_R$ as follows:
	\begin{small}
	\begin{equation}
	\mathbf{G}_R\approx-\nabla_{\bm{\theta}} \mathcal{F}^{\bm{\theta}}_{\mathtt{CL}} \left[ \left(\nabla_{\bm{\omega}}  \mathcal{F}_{\mathtt{CL}}^{\bm{\theta}\top} \nabla_{\bm{\omega}}  \mathcal{F}^{\bm{\omega}}_{\mathtt{OL}} \right) \big/  \left(\nabla_{\bm{\omega}}  \mathcal{F}_{\mathtt{CL}}^{\bm{\theta}\top} \nabla_{\bm{\omega}}  \mathcal{F}^{\bm{\theta}}_{\mathtt{CL}} \right)\right].\label{eq:approximate-G0}
	\end{equation}
    \end{small}	
%	Based on the above derivations, we can summarize the complete solution strategy in Alg.~\ref{alg:fast0}.  In the training process, with the current parameters $\bm{\theta}$, we first optimize $\bm{\omega}$ according to the objective $\mathcal{F}_{\mathtt{CL}}$ in several steps to approximate the dynamic best response, i.e., $\hat{\bm{\omega}}(\bm{\theta}) \approx \bm{\omega}(\bm{\theta})$. Then, $\bm{\omega}(\bm{\theta})$ is  back-propagated to $\mathcal{F}^{\bm{\omega}}_{\mathtt{OL}}$, $\mathbf{G}_R$ is calculated based on our proposed implicit gradient strategy (i.e., Eq.~\eqref{eq:approximate-G0}), which in turn updates  $\bm{\theta}$ until convergence. 		
	Drawing upon the aforementioned derivations, we can succinctly outline the comprehensive solution strategy as Algorithm \ref{alg:fast0}. During the training phase, given the current parameter set $\bm{\theta}$, we initiate the optimization of $\bm{\omega}$ based on the objective function $\mathcal{F}^{\bm{\theta}}_{\mathtt{CL}}$, iteratively refining it to approximate the dynamic best response, denoted as $\hat{\bm{\omega}}(\bm{\theta}) \approx \bm{\omega}(\bm{\theta})$. Subsequently, we propagate the updated $\bm{\omega}(\bm{\theta})$ to $\mathcal{F}^{\bm{\omega}}_{\mathtt{OL}}$, where we calculate the response gradient $\mathbf{G}_R$ utilizing our proposed implicit gradient strategy (i.e., Eq. \eqref{eq:approximate-G0}). This computed gradient, in turn, facilitates the iterative update of $\bm{\theta}$ until convergence is attained.

		\begin{algorithm}[ht]
		\caption{Fast Solution Strategy for LwCL.}\label{alg:fast0}
		\begin{algorithmic}[1]
			
			\REQUIRE Initialization of $\bm{\theta}, \bm{\omega}$, energy functions $\mathcal{F}^{\bm{\omega}}_{\mathtt{OL}}$ and $\mathcal{F}^{\bm{\theta}}_{\mathtt{CL}}$, and other essential  hyper-parameters. 
			\ENSURE  The optimal parameters $\bm{\theta}^*$ and $\bm{\omega}^*$.
			\WHILE{not converge}
			\STATE \% $CL-level\ variable\ probe$ 
			\STATE Obtain an approximation $\hat{\bm{\omega}}$ to $\bm{\omega}$ by updating $\hat{\bm{\omega}}\leftarrow \bm{\omega}- \alpha \nabla_{\bm{\omega}} \mathcal{F}^{\bm{\theta}}_{\mathtt{CL}} \left(\bm{\omega} \right)$  ($\alpha$: learning rate).
			%\STATE Obtain approximation $\hat{\bm{\omega}}$ to $\bm{\omega}$ by solving $\varUpsilon(\bm{\theta})$.
			\STATE \% $OL-level\ variable\ probe$ 
			\STATE Calculate the response gradient  $\mathbf{G}_{R}$ with $\hat{\bm{\omega}}$ and the current $\bm{\theta}$, according to Eq.~\eqref{eq:approximate-G0}.
			\STATE Calculate $\nabla_{\bm{\theta}} \varphi(\bm{\theta})$ with $\mathbf{G}_{R}$ by Eq.~\eqref{eq:bilevel-gradient0}.
			\STATE Update $\bm{\theta} \leftarrow \bm{\theta} - \beta \nabla_{\bm{\theta}} \varphi(\bm{\theta})$ ($\beta$: learning rate).
			\ENDWHILE
			\RETURN  $\bm{\theta}^*$ and $\bm{\omega}^*$.
		\end{algorithmic}
	\end{algorithm}

   \subsection{Discussion} 
%    Next, we discuss the superiority of our learning framework over previous methods from two aspects: model reformulation and solving strategy.
    %\textbf{Naive Alternating Learning v.s. }
    By fundamentally elucidating the underlying coupling relationships among multiple learning tasks in complex problems, our methodology provides a comprehensive understanding of their intricate interplay. Moreover, the proposed dynamic best response based solution strategy not only showcases scalability, adaptability, and generalizability but also empowers its application across a broad spectrum of large-scale, high-dimensional real-world scenarios.
    % In summary, our approach not only uncovers the fundamental connections between learning tasks but also offers an efficient and versatile solution for addressing complex challenges in various practical contexts.
    
    	%\textbf{Unified Perspective for Redefining Modern Complex Problems: Hierarchical Optimization and Flexible Task Constraint Integration.} 
%    	\textbf{Unifying Complex Learning Problems with LwCL.}
    	\textbf{Unveiling the Intrinsic Coupling Relationships for Complex Learning Problems.}
%    	In contrast to previous methods that design modeling approaches based on specific tasks, we provide a unified perspective (i.e., LwCL) to redefine the underlying mechanism of these modern complex problems. On the one hand, through a hierarchical optimization framework, we can delve deep into the potential coupling relationships among multiple tasks and accurately characterize them. On the other hand, our framework allows for the flexible integration of various relevant task constraints in practical application scenarios and can be easily embedded into the majority of contemporary complex learning problems, resulting in consistent performance improvements. More details on the comparative mechanisms can be found in Sec.~\ref{sec:application}.
    	Traditional approaches in the past have often relied on task-specific methodologies, limiting their generalization capabilities and hindering their transferability to different tasks. Moreover, accurately capturing the interdependencies between multiple related learning tasks has proven challenging due to the empirical nature of designing learning strategies. Hence, the need for a unified framework arises, one that can reconcile diverse modeling approaches and explore the inherent connections among these tasks.   	
    	The LwCL framework tackles these challenges by explicitly considering the nested structure of learning tasks. Its hierarchical optimization framework provides a profound understanding of the potential coupling relationships among tasks, allowing for accurate characterization. Additionally, the framework offers flexibility in integrating various task constraints, rendering it suitable for addressing a wide array of complex learning problems. By leveraging the hierarchical structure, the LwCL framework not only enhances performance on intricate tasks but also enables efficient transfer learning across different domains.

    	\textbf{Implicitly-derived Fast and Efficient Solution Strategy.} 
		Indeed, the most  straightforward idea towards real-world vision applications  is to employ alternating iterative algorithms, where one component is fixed while the other is optimized. While the alternating iterative mechanism exhibits sound design principles, it often leads to a fragmentation between the two learning tasks in practical implementations. Specifically, in Eq~\eqref{eq:bilevel-gradient0}, conventional alternating methods directly overlook the computation of the coupling term $\mathbf{G}_R$, thereby disregarding the gradient feedback from CL to OL during the back-propagated process. In contrast, our proposed method addresses this limitation by emphasizing the collaborative synergy between the CL and OL, which is unattainable in traditional alternating approaches.  Our proposed solution strategy accurately computes the optimal gradient-response during each iteration of the back-propagation, ensuring a more stable and expeditious convergence in the learning process. In addition to the aforementioned intuitive solutions, the development of bilevel optimization methods capable of performing gradient-based explicit and implicit algorithms through automatic differentiation holds significance~\cite{liu2021investigating}. Nevertheless, classical first-order gradient-based algorithms typically suffer from high complexity and low operational efficiency due to the computation of the recursive or Hessian gradient for $\mathbf{G}_R$, rendering them impractical for high-dimensional complex real-world scenarios.
    	
    	We would like to emphasize that our proposed fast solution strategy accurately computes the gradient-response $\mathbf{G}_R$. Furthermore, it exhibits significant superiority over state-of-the-art gradient-based bilevel optimization methods, particularly in terms of convergence speed and computational complexity. 		
    	In regards to computational complexity, Alg.~\ref{alg:fast0} circumvents the need for unfolded recurrent iterations or Hessian inversions for $\mathbf{G}_R$, thereby avoiding any computations involving Hessian- or Jacobian-vector products. The complexity of our strategy primarily stems from computing the first-order gradient.		
    	Considering that the calculation of the function's first derivative and the Hessian-vector product share similar time and space complexity, our proposed approximate method simplifies the process of computing the gradient-response $\mathbf{G}_R$ to evaluating the first-order derivative a few fixed times. 
    	In the subsequent experimental section, we undertake a comprehensive set of numerical and real experiments to compare various traditional gradient-based bilevel optimization methods and our fast solution strategy.  Through these experiments, we aim to substantiate the exceptional performance of our strategy, specifically in terms of convergence speed and memory utilization. More details on the comparative mechanisms can be found in Sec.~\ref{sec:application}.

	\section{Applications of LwCL} \label{sec:sec3}  
	In this section,we provide an elaborate discussion on the versatility of our proposed framework as a general methodology, which can be seamlessly applied to a diverse array of LwCL applications, spanning domains such as AL, ART and TDC within the realms of vision and learning. By applying our framework to these diverse applications, we aim to demonstrate its broad applicability and effectiveness across different domains and tasks.

	\subsection{AL-type Applications} 
	In the realm of  AL-type LwCL applications, our focus lies on the introduction various discriminator learning tasks  $\mathcal{N}_{\bm{\omega}}^{C}$ to assist with the generator learning tasks  $\mathcal{N}_{\bm{\theta}}^{O}$. We emphasize that CL entails the incorporation of discriminators (potentially multiple), classifiers, and critics, each equipped with specialized architectures designed to address diverse applications within the realms of vision and learning.  As for the constraint energy function $\mathcal{F}^{\bm{\theta}}_{\mathtt{CL}}$, its fundamental concept lies in establishing the relationship between the output distribution of  $\mathcal{N}_{\bm{\theta}}^{O}$ and the distribution of the desired solution for $\mathcal{N}_{\bm{\omega}}^{C}$.  In the subsequent discussion, we primarily delve into four prominent categories of representative applications, namely  vanilla GAN, image generation, style transfer and imitation learning, which serve as exemplars to showcase the versatility of our framework and its efficacy in these domains.

	%we introduce the CL function $f$ by constructing the relation between the distribution of outputs for OL network and the distribution of the desired solution. 
	%In the following we first introduce the most representative GANs and then explore three applications with introducing various  discriminative constraints, i.e., \textit{image generation, style transfer and imitation learning.}
	
	\subsubsection{GAN and Its Variants}   
%	Formally, the learning process of vanilla GAN can be abstracted as the $``\mathcal{\mathtt{Distance}}"$ minimization between the generated distribution $P_{G}$ and data distribution $P_{data}$, i.e.,  $
%		\min``\mathcal{\mathtt{Distance}}"(P_{G},P_{data}).\label{eq:loss}$ 
%	Vanilla GAN dynamics under the standard definition advocate to introduce an auxiliary discriminator $D$ to assist the generator $G$  through alternating learning strategy in the divergence minimization $\min_{G} \max_{D}\mathcal{V}(G,D)$.  
	Formally, the learning process of GAN can be conceptualized as the minimization of a distance metric, denoted as $``\mathcal{\mathtt{Distance}}"$, between the generated distribution $P_{G}$ and the data distribution $P_{data}$, represented as $
	\min``\mathcal{\mathtt{Distance}}"(P_{G},P_{data}).$ Under the standard definition, vanilla GAN dynamics advocate the incorporation of an auxiliary discriminator $D$ to facilitate the training of the generator $G$ through an alternating learning strategy, seeking to minimize the divergence in the objective $\min_{G} \max_{D}\mathcal{V}(G,D)$. In essence, as the most representative instance of LwCL, it can be formulated as a dynamically coupled game process, expressed as  $\mathcal{V}\big[G(\bm{\theta}),D(\bm{\omega})\big]=\mathcal{F}^{\bm{\omega}}_{\mathtt{OL}}(\bm{\theta})=-\mathcal{F}^{\bm{\theta}}_{\mathtt{CL}}(\bm{\omega}).$
		
		By employing an alternating direction iteration strategy, the original learning strategy  establishes two gradient flows using gradient descent:  $\bm{\omega}_{t+1}\leftarrow\bm{\omega}_t-  \nabla_{\bm{\omega}} \mathcal{F}^{\bm{\theta}}_{\mathtt{CL}} \left(\bm{\omega}_t \right)$ and $\bm{\theta}_{t+1}\leftarrow\bm{\theta}_t-  \nabla_{\bm{\theta}}  \mathcal{F}^{\bm{\omega}}_{\mathtt{OL}} \left(\bm{\theta}_t\right).$ This leads to two independent optimization branches for $\bm{\omega}$ and $\bm{\theta}$ that proceed in parallel. However, the optimization of the generator depends on the discriminator's parameters from the previous step, rather than the current step. This inaccurate approximation fails to capture the coupled best response gradient term depicted in Eq.~\eqref{eq:bilevel-gradient0}. 		
	In contrast, our LwCL framework accurately formulates and characterizes the potential dependency of $\bm{\theta}$ on the current $\bm{\omega}$. Consequently, the optimization of the discriminator can be described by an exact estimated dynamic best response, which is then dynamically back-propagated to the optimization process for the generator dynamics. 
	In light of these considerations, our proposed framework fundamentally circumvents the occurrence of training instabilities and mitigates mode collapse issues. To validate its effectiveness in addressing these challenges, we conduct a comprehensive set of experiments in Sec.~\ref{sec:application}, which showcase the results of these experiments and provide compelling support for the effectiveness of our framework.

	\subsubsection{Image Generation} 		
	Image generation aims to generate intricate and diverse images from compact seed inputs.  Existing research focuses on developing diverse generative models, but they often encounter challenges during model training and require manual tuning techniques to mitigate mode collapse issues.
	With the versatility of our proposed framework, we apply our learning strategy to state-of-the-art generative model architectures. Specifically, we introduce different constrained objectives, referred to as CL energy functions $\mathcal{F}^{\bm{\theta}}_{\mathtt{CL}}$, such as binary cross-entropy loss, least squares loss and 1-Lipschitz limit-loss with spectral norm. These objectives correspond to discriminators with different network structures. Remarkably, our LwCL framework seamlessly integrates into various advanced GAN variants without necessitating architectural modifications or loss selection changes, thereby consistently enhancing performance.	In Sec.~\ref{sec:application}, we present comprehensive experimental results to demonstrate the effectiveness and efficiency of our LwCL framework.  These results showcase more stable training performance and improved generalization capabilities, substantiating the practical benefits of our approach.

	\subsubsection{Style Transfer}

	Style transfer aims to impose style constraints by optimizing the adversarial loss between two distinct datasets, while ensuring content preservation through reverse transformations. 	
	Drawing inspiration from the circular generative adversarial architecture proposed in Zhu \textit{et al}.~\cite{zhu2017unpaired}, we establish a bidirectional adversarial learning framework to guide the style transfer task. Specifically, by creating a cyclic mapping between two domains, denoted as $X$ and $Y$, we introduce two generators, namely $G_1$ and $G_2$, along with two discriminators, denoted as $D_X$ and $D_Y$. To capture the complexity of unsupervised learning, we design two components for our loss functions:  the least squares loss $\mathcal{L}_{GAN}$ and cycle consistency loss $\mathcal{L}_{cyc}$, which account for the original input and the output after inverse transformation. Within our LwCL framework, we define the objective learner $\mathcal{N}_{\bm{\theta}}^{O}$ and the constraint learner $\mathcal{N}_{\bm{\omega}}^{C}$ as the two generators with parameters $\bm{\theta}$ and the two discriminative classifiers with parameters $\bm{\omega}$, respectively. The OL and CL objectives can then be expressed as $\mathcal{F}^{\bm{\omega}}_{\mathtt{OL}}=-\mathcal{F}^{\bm{\theta}}_{\mathtt{CL}}=\mathcal{L}(G_1,G_2,D_X,D_Y)=\mathcal{L}_{GAN}(G_1,D_Y,X,Y)+\mathcal{L}_{GAN}(G_2,D_X,Y,X)+ \mathcal{L}_{cyc}(G_1,G_2)$. More details on the setup of the forward and backward cyclic consistency functions can be found in the experimental section.
	
	\subsubsection{Imitation Learning}
	Imitation learning endeavors to achieve optimal decision-making by interacting with the environment and acquiring knowledge from experiences. 
	The objective of imitation learning is to simultaneously learn a state-action value function, denoted as $Q^{\pi}$, which predicts the expected discounted cumulative reward, and an optimal policy that aligns with the value function. Formally, we have: 
$Q^{\pi}(s,a)=\mathbb{E}_{s_{i+j}\sim\mathcal{P},r_{i+j}\sim\mathcal{R},a_{i+j}\sim\pi}
	(\sum_{k=0}^{\infty}\gamma^jr_{i+j}|s_i=s,a_i=a),\label{eq:eqac}$
	where $\mathcal{P}$ and $\mathcal{R}$ denote dynamics of the environment and reward function, respectively. Here, $s$ and $a$ are the state and action, $i$ and $j$ refer to the i-th and j-th steps in the learning process.
	Within our LwCL paradigm, the actor and critic correspond to the objective learner  $\mathcal{N}_{\bm{\theta}}^{O}$ and constraint learner $\mathcal{N}_{\bm{\omega}}^{C}$, respectively. Let $\bm{\theta}$ denote the parameters of the state-action value-function and $\bm{\omega}$ denote the parameters of the policy $\pi$. The expressions for $\mathcal{F}^{\bm{\omega}}_{\mathtt{OL}}$ and $\mathcal{F}^{\bm{\theta}}_{\mathtt{CL}}$ are given by: $\mathcal{F}^{\bm{\omega}}_{\mathtt{OL}}:= \mathbb{E}{s_{i},a_{i}\sim\pi}[\mathtt{div}(\mathbb{E}_{s_{i+1},a_{i+1},r_{i+1}} 
	\left(D_Q\right) \Vert Q(s_{i},a_{i}))],$ and $\mathcal{F}^{\bm{\theta}}_{\mathtt{CL}}:=-\mathbb{E}_{s_{0}\sim p_{0},a_{0}\sim\pi} Q^{\pi}(s_{0},a_{0}),$ where $\mathtt{div}(\cdot\Vert\cdot)$ represents any divergence and $D_Q=r_{i+1}+\gamma Q(s_{i+1},a_{i+1})$.  
	For specific settings of the state-action value function, please refer to the experimental section.
	
	\subsection{ART-type Applications} 
	 As mentioned previously, ART-type LwCL tasks solving sophisticated applications have introduced related tasks as auxiliary CL devices to augment the considered OL tasks. 
	 In the subsequent subsections, we delve into three specific applications that leverage auxiliary task constraints: medical image analysis, low-light image enhancement and hyper-parameter learning.

	\subsubsection{Medical Image Analysis}

	Medical image analysis involves the extraction of anatomical structures or lesions from medical images.
	Drawing inspiration from the concept that learning registration can provide additional pseudo-labeled training data to assist segmentation~\cite{DBLP:conf/miccai/XuN19}, we leverage our LwCL framework to dynamically address inter-task dependencies. In our framework, the registration process serves as the objective learner $\mathcal{N}_{\bm{\theta}}^{O}$, while the segmentation process acts as the constraint learner $\mathcal{N}_{\bm{\omega}}^{C}$. 	
Building upon a base model~\cite{DBLP:conf/miccai/CicekALBR16}, we incorporate a semantic consistency constraint into the segmentation task. Under our LwCL framework, the OL procedures for the registration task can be formulated as $\mathcal{F}^{\bm{\omega}}_{\mathtt{OL}}:=\min_{\theta}\mathcal{L}_{reg}(\mathcal{N}_R(I_{mov},I_{fix},\theta))$,  where $\mathcal{N}_R$ represents the registration network with learnable parameters $\bm{\theta}$, and $I_{mov}$ and $I_{fix}$ denote the moving image and fixed image, respectively. Subsequently, by obtaining the warped image $I_{war}$, the CL procedures for the segmentation task can be formulated as $\mathcal{F}^{\bm{\theta}}_{\mathtt{CL}}:=\min_{\theta}\mathcal{L}_{seg}(\mathcal{N}_S(\mathcal{N}_R(I_{war},I_{fix},\theta));\omega)$, where $\mathcal{N}_S$ represents the segmentation network with learnable parameters $\bm{\omega}$. Please refer to the experimental section in Sec.~\ref{sec:sec5} for specific details on the settings of the loss functions.

	\subsubsection{Low-light Image Enhancement}

	Low-light image enhancement aims to reveal hidden information in dark areas to improve overall image quality.
	Drawing on the principles of the Retinex theory, we delve into the impact of downstream perceptual tasks, such as object detection, on the performance of upstream enhancement tasks. Guided by this concept, we construct a low-light enhancement network, inspired by recent advancements~\cite{ma2022toward}, as our objective learner $\mathcal{N}_{\bm{\theta}}^{O}$. Furthermore, we introduce a face detector proposed by~\cite{li2019dsfd} as an auxiliary constraint learner $\mathcal{N}_{\bm{\omega}}^{C}$. 	
	Within our LwCL framework, we employ the unsupervised illumination learning loss $\mathcal{F}^{\bm{\omega}}_{\mathtt{OL}}:=\mathcal{L}_p+\mathcal{L}_s$  as the OL function. Here, $\mathcal{L}_p$ and $\mathcal{L}_s$ represent the pixel fidelity term and smoothness term, respectively, which evaluate the performance of the upstream enhancement task. For the energy function $\mathcal{F}^{\bm{\theta}}_{\mathtt{CL}}$ of the constraint learner, we introduce the anchor-based multi-task loss and progressive anchor loss to assess the detection performance, defined as: $\mathcal{F}^{\bm{\theta}}_{\mathtt{CL}}:=\mathcal{L}_{SSL}(a)+\mathcal{L}_{SSL}(sa)$. Detailed configurations of the training loss and hyper-parameters can be found in the experimental Sec.~\ref{sec:sec5}.

	\subsubsection{Hyper-parameter Learning} \label{sec:hpl}

	Hyper-parameter learning aims to determine the optimal configuration of hyper-parameters for a given machine learning task. Hyper-parameters are parameters that remain fixed during the training process of a machine learning model. 	
	In essence, the goal of hyper-parameter learning is to find a set of hyper-parameters that minimizes the loss or maximizes the accuracy of the objective learning task. Mathematically, it can be expressed as  $\bm{\theta}^*=\arg\min E_{(\mathcal{D}_{\mathtt{tr}},\mathcal{D}_{\mathtt{val}})\sim \mathcal{D}}L(g_{\bm{\omega}}(\cdot),\bm{\theta},\mathcal{D}_{\mathtt{tr}},\mathcal{D}_{\mathtt{val}})$, where $L$ represents the objective function, $g_{\bm{\omega}}(\cdot)$ denotes the learning algorithm applied to the hyper-parameters $\bm{\theta}$, and the model is trained on the training dataset $\mathcal{D}_{\mathtt{tr}}$ and validated on the validation dataset $\mathcal{D}_{\mathtt{val}}$.  
	Within our LwCL framework, the objective learner $\mathcal{N}_{\bm{\theta}}^{O}$ aims to minimize the loss on the validation set $\mathcal{F}^{\bm{\omega}}_{\mathtt{OL}}(\bm{\theta},\bm{\omega};\mathcal{D}_{\mathtt{val}})$ with respect to the hyper-parameters $\bm{\theta}$, which include parameters such as learning rate, batch size, optimizer, and loss weights. On the other hand, the constraint learner $\mathcal{N}_{\bm{\omega}}^{C}$ is responsible for generating a learning algorithm by minimizing the training loss $\mathcal{F}^{\bm{\theta}}_{\mathtt{CL}}(\bm{\theta},\bm{\omega};\mathcal{D}_{\mathtt{tr}})$ with respect to the model parameters $\bm{\omega}$, which encompass weights and biases.

	\subsection{TDC-type Applications} \label{sec:tdc}

     As mentioned previously, TDC-type LwCL tasks, inspired by the concept of  ``divide and conquer'' aim to analyze and formulate the coupling relationship. This approach involves decomposing the overall learning task into two distinct components: the objective learner $\mathcal{N}_{\bm{\theta}}^{O}$ and the constraint learner $\mathcal{N}_{\bm{\omega}}^{C}$.   In the subsequent sections, we delve into the practical implementations of two notable application types, namely image deblurring and multi-task meta-learning learning and 
     These applications serve as representative examples to showcase the efficacy and versatility of the the LwCL paradigm in addressing diverse learning challenges. 
     
     \begin{figure}[htp]
     	\begin{center}
     		\begin{tabular}{c@{\extracolsep{0.3em}}c@{\extracolsep{0.3em}}}
     			\includegraphics[width=0.47\linewidth,clip]{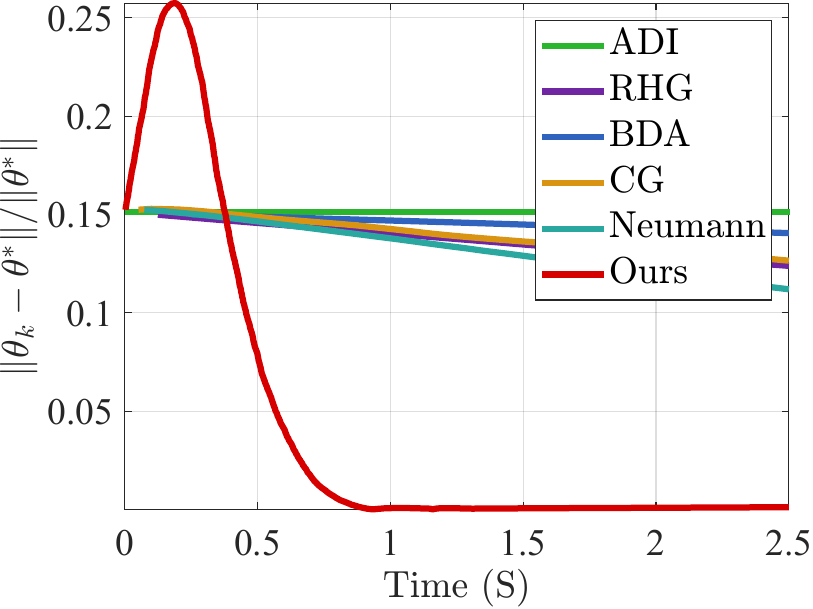}&\includegraphics[width=0.47\linewidth]{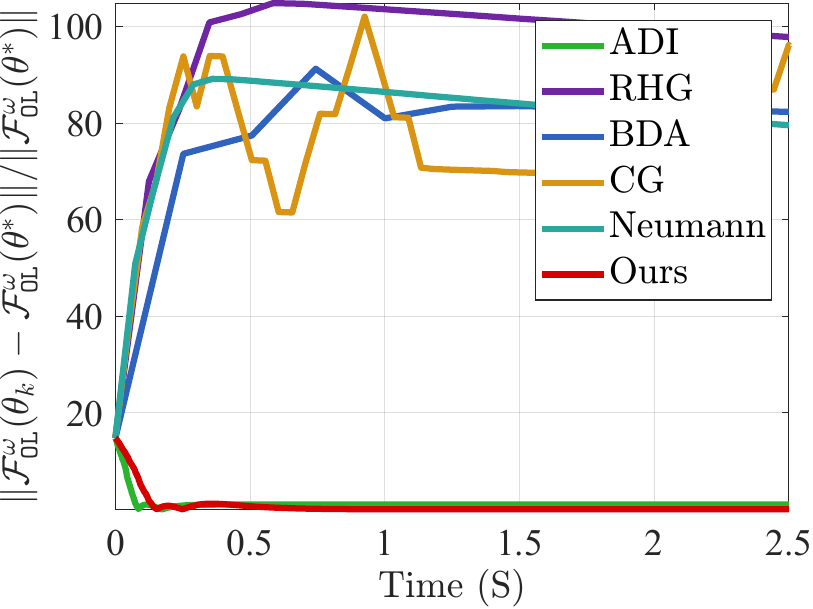}\\
     			%(a)&(b)\\
     			%\multicolumn{2}{c|}{Convergence with the initial point $(\bm{\theta},\bm{\omega})=(0,0)$ and $(\bm{\theta},\bm{\omega})=(1,1)$}\\
     		\end{tabular}
     	\end{center}
     	%\vspace{-0.25cm}
     	\caption{
     		%The convergence behavior of naive alternating learning strategy  (aptly abbreviated AL) and our LwCL methodology under different initial points $(\bm{\theta}_0,\bm{\omega}_0)=(0,0)$ and $(\bm{\theta}_0,\bm{\omega}_0)=(1,1)$. 
     		%The subfigures (a) and (b) respectively show the errors of $\bm{\theta}$  variable (i.e., $\Vert{\bm{\theta}_k}-{\bm{\theta}}^*\Vert/\Vert{\bm{\theta}}^*\Vert$) and objective $F$, i.e., $\Vert F({\bm{\theta}}_k,{\bm{\omega}}_k)-F({\bm{\theta}^*},{\bm{\omega}^*})\Vert/ \Vert F({\bm{\theta}^*},{\bm{\omega}^*})\Vert$. 
     		%The legend is only plotted in the subfigure (a).
     		Illustrating the convergence curve of $\Vert{\bm{\theta}_k}-{\bm{\theta}}^*\Vert/\Vert{\bm{\theta}}^*\Vert$ and $\Vert \mathcal{F}^{\bm{\omega}}_{\mathtt{OL}}({\bm{\theta}}_k)-\mathcal{F}^{\bm{\omega}}_{\mathtt{OL}}({\bm{\theta}^*})\Vert/ \Vert \mathcal{F}^{\bm{\omega}}_{\mathtt{OL}}({\bm{\theta}^*})\Vert$ among series of mainstream bilevel optimization strategies. 
     	}
     	\label{fig:syn_convergence}
     \end{figure} 	
     
     \begin{table}[htp]
     	\begin{center} \footnotesize
     		\renewcommand{\arraystretch}{1.2}
     		\caption{Comparison of the time complexity (Hour) and space complexity (MB) on high-dimensional numerical scenarios under a given convergence criterion (i.e., $\Vert {\bm{\theta}}_k-{\bm{\theta}}_{k-1}\Vert/\Vert{\bm{\theta}}_k\Vert \leq 10^{-5}$). Note that ``N/A'' means that the calculation time exceeds 1200 hours and the calculation memory exceeds 1024 MB. ``N/C'' indicates that the convergence criterion is not met.  ``T'' and ``M'' represent ``Time'' and ``Memory'', respectively. ``$n$'' denotes the dimension of of $\bm{\omega}$. } 
     		%\vskip -0.1in
     		\label{table:Tnet} 
     		\setlength{\tabcolsep}{1mm}{
     			\begin{tabular}{|c|c|c|c|c|c|c|c|}				
     				%\hline
     				%\footnotesize Dataset  & \footnotesize Metric & \footnotesize RHG   & \footnotesize BDA   & \footnotesize CG    & \footnotesize Neumann & \footnotesize IGA \\
     				%\hline   
     				%				\hline
     				%				\multirow{2}{*}{\footnotesize 2D MOG}& T &1083.97& 1512.74 & 317.96 & \textcolor{blue}{\textbf{311.16}}&\textcolor{red}{\textbf{107.77}} \\
     				%				\cline{2-7} 
     				%				& M  &94.85&106.31& 45.56&  \textcolor{blue}{\textbf{43.92}}&\textcolor{red}{\textbf{16.34}} \\
     				%				\hline 
     				%				\multirow{2}{*}{\footnotesize 3D MOG}&  T &1031.49& 1537.86 & 373.62 & \textcolor{blue}{\textbf{309.21}}&\textcolor{red}{\textbf{94.98}} \\
     				%				\cline{2-7}
     				%				& M  &74.90&96.85& 23.16&  \textcolor{blue}{\textbf{21.05}}&\textcolor{red}{\textbf{12.56}} \\
     				%				\hline 
     				\hline 
     				\footnotesize $n$  & \footnotesize Metric & \footnotesize ADI & \footnotesize RHG   & \footnotesize BDA   & \footnotesize CG    & \footnotesize NS & \footnotesize Ours \\
     				\hline	
     				\hline					
     				%\multirow{2}{*}{$10^{2}$} &T & & 5.995 & 7.014 & \textcolor{blue}{\textbf{4.467}} & 4.802 & \textcolor{red}{\textbf{1.249}} \\
     				\multirow{2}{*}{$10^{2}$} &T & \multirow{2}{*}{N/C} & 0.984 & 1.126 & 5.292 & \textcolor{blue}{\textbf{0.492}} & \textcolor{red}{\textbf{0.150}} \\
     				%\cline{2-2} \cline{4-8}
     				& M & & 9.961 & 10.552 & 2.032 &\textcolor{blue}{\textbf{1.983}} & \textcolor{red}{\textbf{0.065}} \\
%     				\hline					
     				%\multirow{2}{*}{$10^{3}$}& T & & 114.7   & 116.8 & 78.01 & \textcolor{blue}{\textbf{77.03}} &\textcolor{red}{\textbf{11.85}} \\
     				\multirow{2}{*}{$10^{3}$}& T & \multirow{2}{*}{N/C}  & 18.829  & 18.688 & 92.420 & \textcolor{blue}{\textbf{9.286}} &\textcolor{red}{\textbf{1.428}} \\
     				% \cline{2-2} \cline{4-8}
     				& M &  &98.552   &98.823 & 20.076 & \textcolor{blue}{\textbf{19.585}} &\textcolor{red}{\textbf{0.552}} \\
%     				\hline
     				%\multirow{2}{*}{$10^{4}$} &T & & 2015  & 2037  & 1652  & \textcolor{blue}{\textbf{1602}}  & \textcolor{red}{\textbf{119.0}}\\
     				\multirow{2}{*}{$10^{4}$} &T & \multirow{2}{*}{N/C}  & 330.795  & 325.92  & 1957.161  & \textcolor{blue}{\textbf{193.13}}  & \textcolor{red}{\textbf{14.346}}\\
     				%\cline{2-2} \cline{4-8}
     				& M & & 1004.681  & 1012.842  & 200.560  & \textcolor{blue}{\textbf{195.642}}  & \textcolor{red}{\textbf{5.310}}\\
%     				\hline
     				%\multirow{2}{*}{$10^{5}$} & T & &N/A   & N/A   & N/A & N/A & \textcolor{red}{\textbf{1198}} \\
     				\multirow{2}{*}{$10^{5}$} & T & \footnotesize \multirow{2}{*}{N/C}  & \footnotesize \multirow{2}{*}{N/A}  & \multirow{2}{*}{N/A}   & \multirow{2}{*}{N/A}  & \multirow{2}{*}{N/A}  & \textcolor{red}{\textbf{144.425}} \\
     				%\cline{2-2} \cline{8-8}
     				& M & &   &  & &  & \textcolor{red}{\textbf{53.030}} \\					
     				\hline
     			\end{tabular}\label{tab:time2}}
     	\end{center} 
     \end{table}

     \subsubsection{Image Deblurring}

     Image deblurring aims to recover a latent clear image $\textbf{u}$ from a blurred counterpart $\textbf{b}$. The physical model governing this process is represented by $\textbf{b}=\textbf{K}\ast \textbf{u}+\textbf{n}$, where $\textbf{K}$, $\textbf{n}$, and $\ast$ denote the blur kernel, additional noise, and the two-dimensional convolution operator, respectively. Typically, image deblurring entails two main tasks: deconvolution, involving the estimation of sharp images from blurred observations, and denoising.     
     Drawing inspiration from the plug-and-play framework, which leverages semi-quadratic splitting, the deblurring problem is decomposed into alternating iterations of two sub-problems concerning $\textbf{u}$ and an auxiliary variable $\textbf{z}$. Within our LwCL framework, we define the constraint learner $\mathcal{N}_{\bm{\omega}}^{C}$  as the fidelity learning sub-problem, addressing $\textbf{u}$ and $\textbf{z}$. This can be formulated as $\{\textbf{u}, \textbf{z}\}=\arg\min_{\textbf{u},\textbf{z}}\{\mathcal{F}^{\bm{\theta}}_{\mathtt{CL}}(\textbf{u},\textbf{z}):=\|\textbf{K}\ast \textbf{u}-\textbf{b}\|_2^2+\mu \|\textbf{u}-\textbf{z}\|^2+k\|\textbf{Wu}\|_1\},$  where $\mu$ represents a penalty parameter, and $\textbf{W}$ denotes the wavelet transform matrix.     
     The objective learner $\mathcal{N}_{\bm{\theta}}^{O}$ can be viewed as the prior learning process, governed by a denoiser $\mathtt{Net}_{\bm{\theta}}(\textbf{u})$, with regard to the variable $\bm{\theta}$. Mathematically, it can be expressed as $\bm{\theta}=\arg\min_{\textbf{z}}\mathcal{F}^{\bm{\omega}}_{\mathtt{OL}}(\mathtt{Net}{\bm{\theta}}(\textbf{u}), \bm{\theta})$. For further details on the parameter configurations, please refer to Sec.~\ref{sec:sec5}.

    \subsubsection{Multi-task Meta-learning} \label{sec:mt-ml}

    Multi-task meta-learning represents a formidable challenge that revolves around swiftly adapting to novel tasks with limited examples. Among these tasks, meta-feature learning stands out as a prominent representative of multi-task and meta-learning, with the objective of acquiring a shared meta feature representation that encompasses all tasks. This is achieved by bifurcating the network architecture into two distinct components: the meta feature extraction part, responsible for generating the cross-task intermediate representation layers (parameterized by $\bm{\theta}$), and the task-specific part, characterized by the multinomial logistic regression layer (parameterized by $\bm{\omega}^j$). This framework allows for building accurate machine learning models utilizing a smaller training dataset, especially in the context of few-shot classification tasks, which are widely recognized in the field.     
    Within the LwCL framework we propose, the intermediate representation layers that produce the meta-features can be viewed as the objective learner $\mathcal{N}_{\bm{\theta}}^{O}$ for multiple task-specific assignments. Consequently, we optimize the performance of these meta-features using the validation set through the defined loss function $\mathcal{F}^{\bm{\omega}^j}_{\mathtt{OL}}(\bm{\theta};\mathcal{D}_{\mathtt{val}}^j)$. Additionally, the forward propagation of the classification layers at the network's end serves as the constraint learner $\mathcal{N}_{\bm{\omega}}^{C}$, wherein the training set loss $\mathcal{F}^{\bm{\theta}}_{\mathtt{CL}}(\bm{\omega}^j;\mathcal{D}_{\mathtt{tr}}^j)$ is utilized to guide the learning process.

\section{Experimental Results}	\label{sec:application}  
In this section,  we first evaluate the learning mechanism of our proposed framework through a meticulous examination of numerical examples. This comprehensive evaluation will facilitate a profound understanding of the framework's underlying principles and intricacies of its learning processes. Subsequently,  we proceed to conduct a series of rigorous and extensive experiments, aimed at meticulously scrutinizing the efficacy and viability of our proposed framework across a diverse array of learning paradigms and visual applications.  All of these experiments are carried out on a high-performance computing system comprising an Intel Core i7-7700 CPU operating at a frequency of 3.6 GHz, 32GB of RAM, and an NVIDIA GeForce RTX 2060 GPU with 6GB of dedicated memory.
\subsection{Mechanism Evaluation} 	\label{sec:sec5-1}
First and foremost,  we commence by assessing the convergence performance and computational complexity of our proposed algorithm on a numerical example.   Specifically, we introduce a challenging toy example~\cite{liu2021value} wherein the CL problem is formulated as a non-convex optimization task:
\begin{equation}
	\label{eq:non-convexExperiment}
	\begin{aligned}
		& \min_{ \bm{\theta} \in \mathbb{R}, \bm{\omega} \in \mathbb{R}^n} \Vert \bm{\theta}-a\Vert^2+\Vert \bm{\omega}-a-\mathbf{c}\Vert^2, \\
		& \text{\ s.t.\ }\;  [\bm{\omega}]_i \in \underset{ [\bm{\omega}]_{i} \in \mathbb{R}}{\mathrm{argmin}}\; \sin( \bm{\theta}+ [\bm{\omega}]_i-[\mathbf{c}]_i), \forall \ i,
	\end{aligned}
\end{equation}
where $[\bm{\omega}]_i$ denotes the $i$-th component of $\bm{\omega}$, 
$a \in \mathbb{R}$ and $\mathbf{c}\in \mathbb{R}^n$ denote adjustable parameters. For this particular numerical example, we set $a=2$ and $[\mathbf{c}]_i=2, \text{ for any }i = 1,2,\cdots,n$. The optimal solution for this numerical example is as follows $$
\bm{\theta}=\frac{(1-n) a+n C}{1+n},\ \text { and } \ [\bm{\omega}]_{i}=C+[\mathbf{c}]_i-\bm{\theta}, \forall \ i,
$$ 
where
$ 
C=\operatorname{argmin}_C\left\{\|C-2 a\| : C=-\pi/2+2 k \pi, k \in \mathbb{Z}\right\} ,
$ 
and the optimal value is $(C-2 a)^{2}-(C-2 a)^{2}/(1+n)$. 
To evaluate the convergence properties and computational complexity, we conducted two sets of experiments: one in low-dimensional simple scenarios with $n=1$, and the other in high-dimensional scenarios with larger values of $n$.

\begin{figure}[tpb]
	\begin{center}
		\begin{tabular}{c@{\extracolsep{1.7em}}c@{\extracolsep{1.7em}}}
			\includegraphics[height=3.4cm,trim=0 0 0 0,clip]{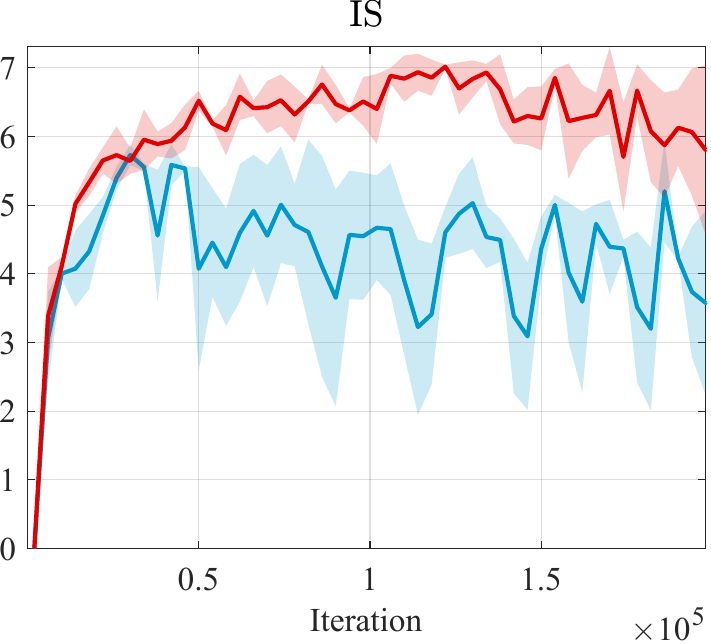} &\includegraphics[height=3.4cm,trim=0 0 0 0,clip]{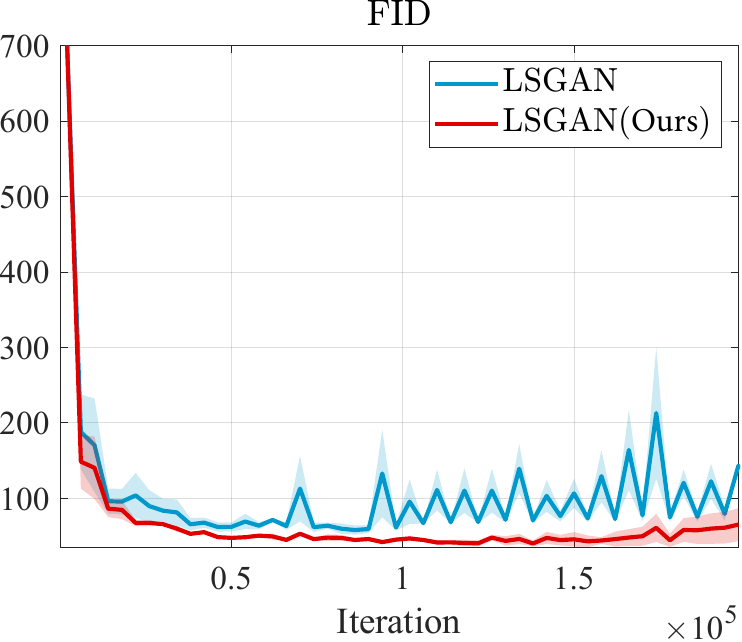} \\
			\multicolumn{2}{c}{\footnotesize CIFAR10 }\\
			\includegraphics[height=3.4cm,trim=0 0 0 0,clip]{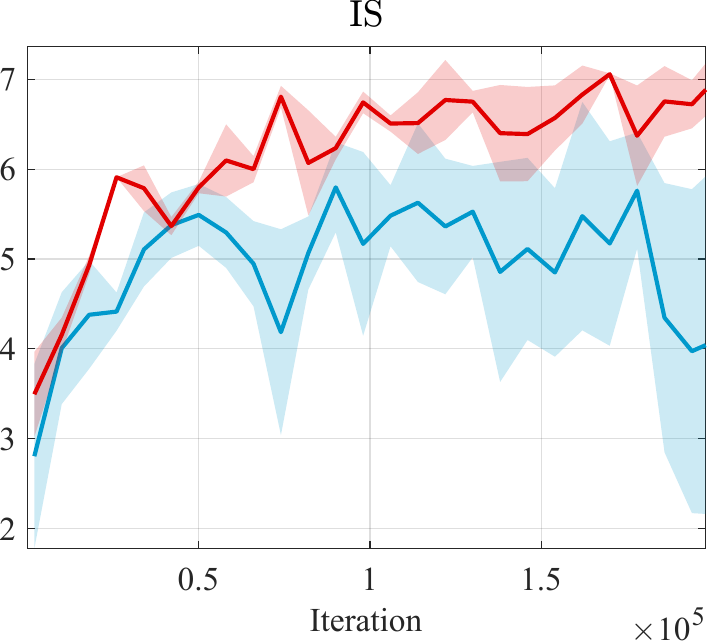} &\includegraphics[height=3.4cm,trim=0 0 0 0,clip]{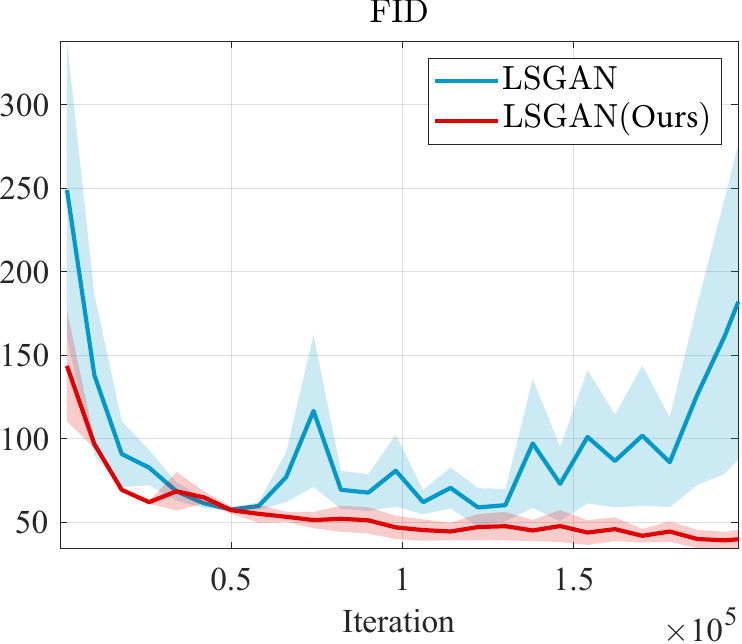}\\	
			\multicolumn{2}{c}{\footnotesize CIFAR100}\\
		\end{tabular}
	\end{center}
	%\vspace{-0.2cm}
	\caption{Comparison of training efficiency results measured by FID and IS on CIFAR10 dataset (\textit{listed top}) and CIFAR100 datasets (\textit{listed bottom}). More stable training convergence curves and better metric scores can be obtained by our LwCL framework.}
	\label{fig:real_is}
\end{figure}

In the case of one dimension, we initialize the point at  $(\bm{\theta},\bm{\omega})=(3,3)$ and the optimal solution is $(\bm{\theta}^*,\bm{\omega}^*)=(3/4\pi,3/4\pi-2)$.  Fig.~\ref{fig:syn_convergence} compares the convergence curves of $\Vert{\bm{\theta}_k}-{\bm{\theta}}^*\Vert/\Vert{\bm{\theta}}^*\Vert$ and $\Vert \mathcal{F}^{\bm{\omega}}_{\mathtt{OL}}({\bm{\theta}}_k)-\mathcal{F}^{\bm{\omega}}_{\mathtt{OL}}({\bm{\theta}^*})\Vert/ \Vert \mathcal{F}^{\bm{\omega}}_{\mathtt{OL}}({\bm{\theta}^*})\Vert$ among series of mainstream bilevel optimization strategies, including Alternating Direction Iteration (ADI), CG~\cite{pedregosa2016hyperparameter}, Neumann~\cite{lorraine2020optimizing}, RHG~\cite{franceschi2017forward}, and BDA~\cite{liu2021general}. It can be observed that these methods either deviate from the optimal solution throughout the iteration process or fail to achieve fast convergence. In contrast, our proposed LwCL algorithm converges to the optimal solution more rapidly, showcasing its superiority. 
%To evaluate computational efficiency, we further compare time and space complexity among our algorithm and current mainstream algorithms in high-dimensional data scenarios.
%As can be seen from the Tab.~\ref{tab:time2}, since the inner constrained objective function is non-convex, the ADI always fails to converge in the high-dimensional scenario. Some implicit gradient methods (i.e. CG and Neumann) keep the computational complexity lower compared to the explicit gradient methods (i.e. RHG and BDA) because they are efficient to avoid the computationally expensive Hessian inverse. Conversely, our LwCL strategy outperforms these algorithms and can guarantee less time consumption and less memory footprint. When the dimension of the variable reaches $n=10^{5}$, all algorithms except ours cause time and memory complexity to skyrocket, or even crash. 
To assess computational efficiency, we compare the time and space complexity between our algorithm and current mainstream algorithms in high-dimensional data scenarios. As depicted in Tab.~\ref{tab:time2}, due to the non-convex nature of the inner constrained objective function, ADI fails to converge in high-dimensional scenarios. Implicit gradient methods such as CG and Neumann exhibit lower computational complexity compared to explicit gradient methods like RHG and BDA, as they efficiently avoid the computationally expensive Hessian inverse. Conversely, our LwCL strategy surpasses these algorithms in terms of time consumption and memory footprint. When the dimension of the variable reaches $n=10^{5}$, all algorithms except ours lead to a sharp increase in time and memory complexity, potentially resulting in crashes.

Furthermore, we conducted experiments on real-world datasets to validate the superiority of our learning strategy. We employed the LSGAN~\cite{mao2017least} as the foundational network architecture to validate the stability of our learning mechanism.
Fig.~\ref{fig:real_is} reports the score comparison of FID and JS in each training iteration of LSGAN on the CIFAR10~\cite{krizhevsky2010convolutional} and CIFAR100~\cite{krizhevsky2009learning}. The results clearly demonstrate that when combined with our LwCL framework, LSGAN exhibits enhanced training stability and achieves superior FID and Inception Score (IS) performance compared to directly applying alternating learning strategies.

\begin{figure*}[!t]
	%\vspace{-0.1cm}
	\begin{center}
		\begin{tabular}{c@{\extracolsep{1em}}c@{\extracolsep{1em}}c@{\extracolsep{1em}}c@{\extracolsep{1em}}c@{\extracolsep{1em}}c@{\extracolsep{1em}}c@{\extracolsep{1em}}}
			\hline \toprule[0.65pt] \specialrule{0em}{1pt}{1pt} 
			\multirow{5}{*}{\footnotesize \textbf{Target}} & \footnotesize \textbf{VGAN} & \footnotesize \textbf{WGAN} & \footnotesize \textbf{ProxGAN} & \footnotesize \textbf{LCGAN} & & \multirow{7}{*}{\textbf{NAL}}\\
			\vspace{-0.4cm}
			\multirow{10}{*}{\includegraphics[height=2.2 cm,trim=0 0 0 0,clip]{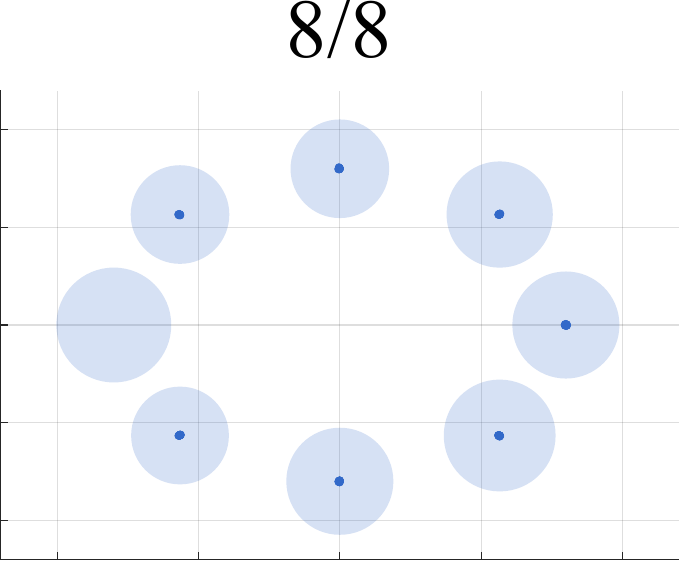}}& &&& & & \\
			&\includegraphics[height=1.9cm]{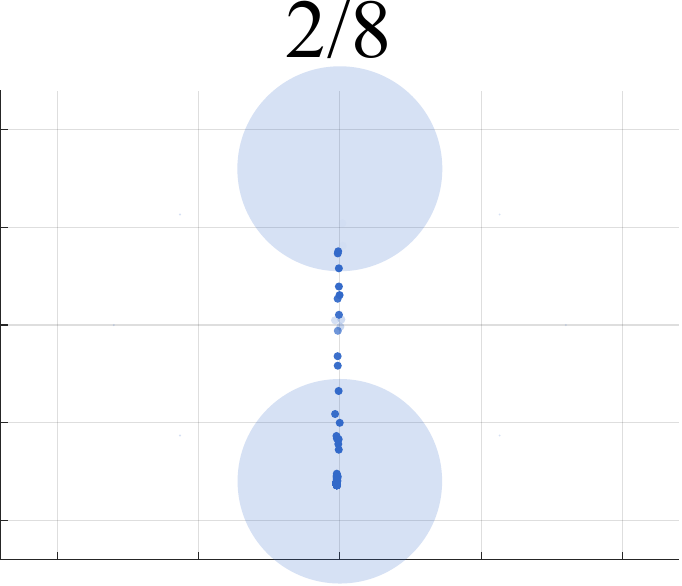}& \includegraphics[height=1.9cm]{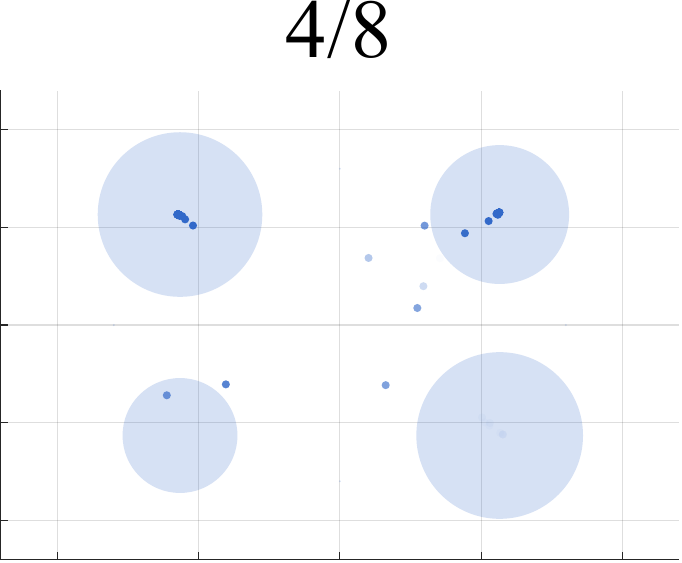} & \includegraphics[height=1.9cm]{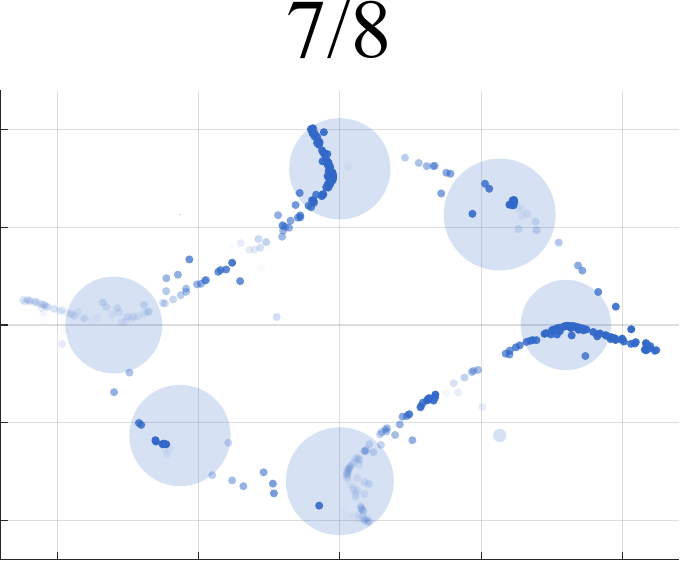} & \includegraphics[height=1.9cm]{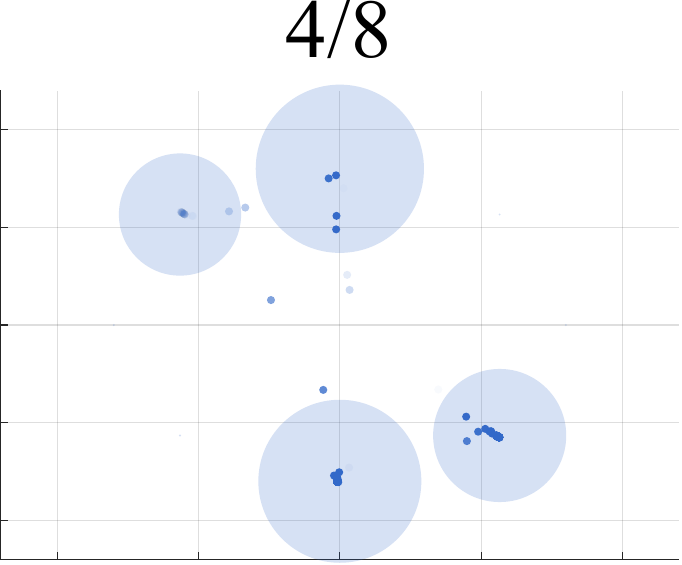}&& \multirow{7}{*}{\textbf{LwCL}}\\
			& \includegraphics[height=1.9cm]{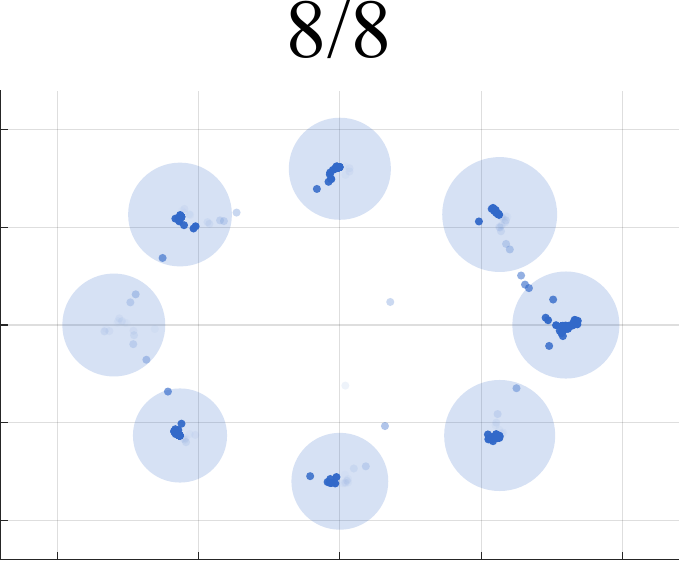} & \includegraphics[height=1.9cm]{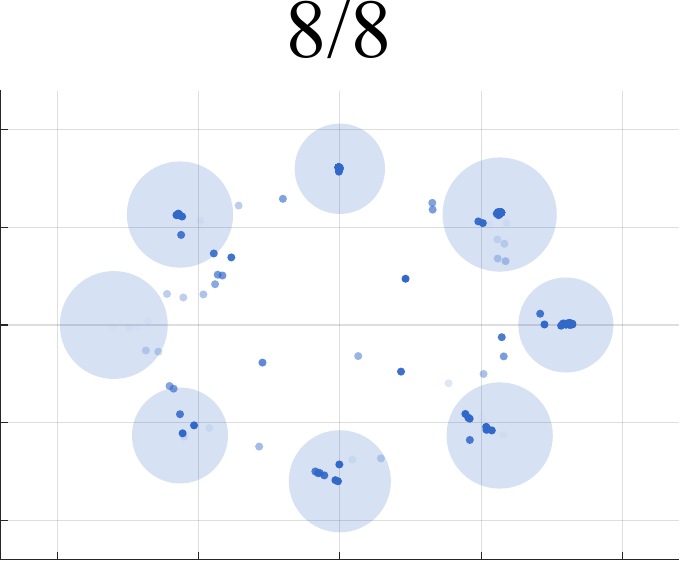} &\includegraphics[height=1.9cm]{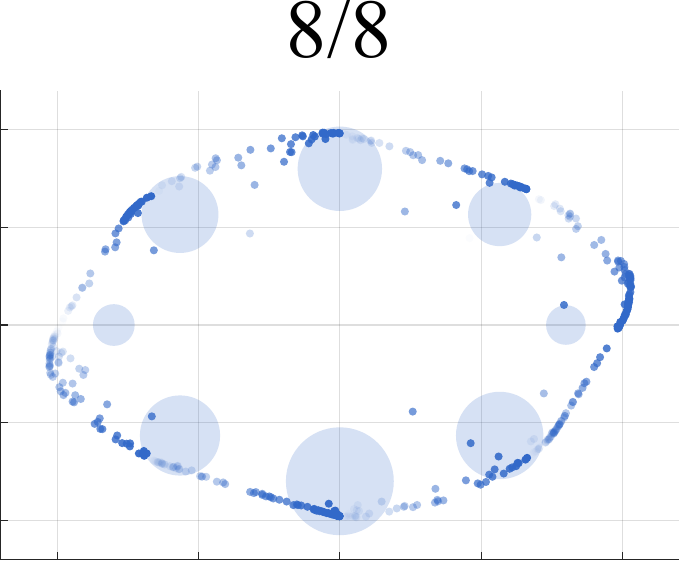} &  \includegraphics[height=1.9cm]{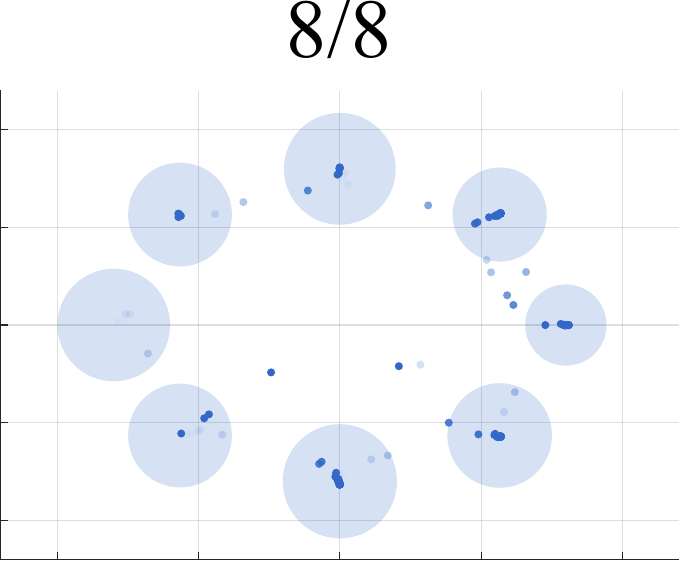} && \\
			\multicolumn{7}{c}{\textbf{2D Ring}} \\ \specialrule{0em}{1pt}{1pt} 
			\hline \specialrule{0em}{1pt}{1pt} 
			\vspace{0.1cm} 
			\multirow{8}{*}{\footnotesize \textbf{Target}} & \footnotesize \textbf{VGAN} & \footnotesize \textbf{WGAN} & \footnotesize \textbf{ProxGAN} & \footnotesize \textbf{LCGAN} & & \multirow{7}{*}{\textbf{NAL}} \\
			
			%\multirow{10}{*}{\includegraphics[height=2.2 cm,trim=0 0 0 0,clip]{fig/rand/target10div}}& &&& & & \\
			%&\includegraphics[height=1.9cm]{fig/grid/GAN10div}& \includegraphics[height=1.9cm]{fig/grid/WGAN10div} & \includegraphics[height=1.9cm]{fig/grid/prox10div} & \includegraphics[height=1.9cm]{fig/grid/LC10div}&& \multirow{7}{*}{\textbf{DCL}}\\
			%& \includegraphics[height=1.9cm]{fig/grid/GN10div} & \includegraphics[height=1.9cm]{fig/grid/WGANGN10div} &\includegraphics[height=1.9cm]{fig/grid/proxGN10div} &  \includegraphics[height=1.9cm]{fig/grid/LCGN10div} && \\
			%\multicolumn{7}{c}{\textbf{2D Grid}} \\ \specialrule{0em}{1pt}{1pt} 
			%\hline \specialrule{0em}{1pt}{1pt} 
			%\vspace{0.1cm} 
			%\multirow{8}{*}{\footnotesize \textbf{Target}} & \footnotesize \textbf{VGAN} & \footnotesize \textbf{WGAN} & \footnotesize \textbf{ProxGAN} & \footnotesize \textbf{LCGAN} & & \multirow{7}{*}{\textbf{Alt.}} \\
			
			\vspace{-0.4cm}			%\multirow{2}{*}{\includegraphics[height=1.9cm]{fig/rand/target10div}}&\includegraphics[height=1.9cm]{fig/rand/GAN10div}& \includegraphics[height=1.9cm]{fig/rand/WGAN10div} & \includegraphics[height=1.9cm]{fig/rand/prox10div} & \includegraphics[height=1.9cm]{fig/rand/LC10div}& w/o\\
			%			& \includegraphics[height=1.9cm]{fig/rand/GN10div} & \includegraphics[height=1.9cm]{fig/rand/WGANGN10div} &\includegraphics[height=1.9cm]{fig/rand/proxGN10div} &  \includegraphics[height=1.9cm]{fig/rand/LCGN10div}& w/\\
			%			\multicolumn{6}{c}{ 2D Rand} \\ 
			%			\multirow{2}{*}{\includegraphics[height=1.9cm]{fig/grid/target10div}}&\includegraphics[height=1.9cm]{fig/grid/GAN10div}& \includegraphics[height=1.9cm]{fig/grid/WGAN10div} & \includegraphics[height=1.9cm]{fig/grid/prox10div} & \includegraphics[height=1.9cm]{fig/grid/LC10div}& w/o\\
			%			& \includegraphics[height=1.9cm]{fig/grid/GN10div} & \includegraphics[height=1.9cm]{fig/grid/WGANGN10div} &\includegraphics[height=1.9cm]{fig/grid/proxGN10div} &  \includegraphics[height=1.9cm]{fig/grid/LCGN10div} & w/ \\
			%			\multicolumn{6}{c}{ 2D Grid} \\ 
			\multirow{12}{*}{\includegraphics[height=1.9cm,trim=0 0 0 0,clip]{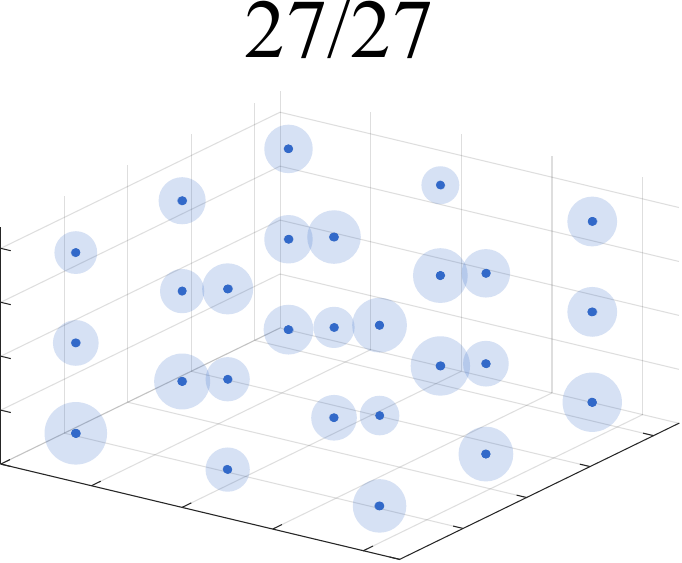}}& &&& & & \\	 
			&\includegraphics[height=1.9cm,trim=0 0 0 0,clip]{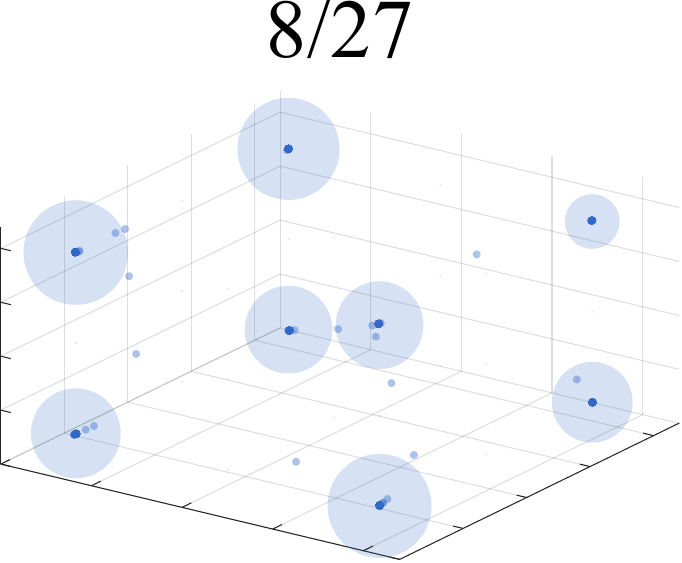}& \includegraphics[height=1.9cm,trim=0 0 0 0,clip]{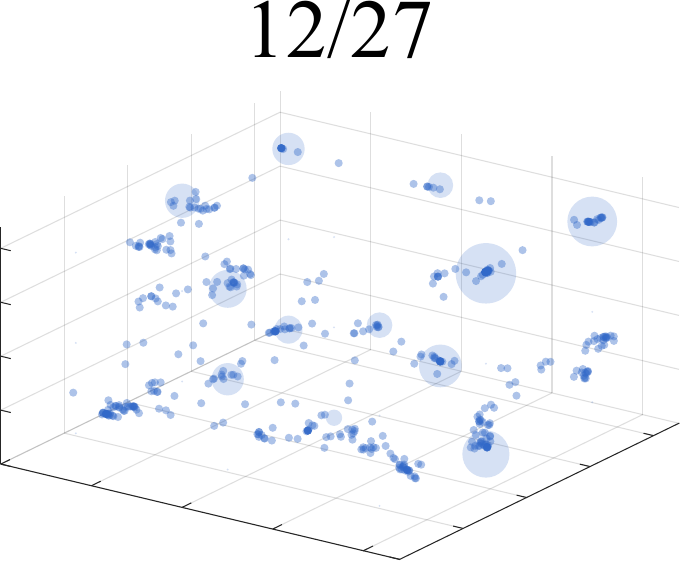} & \includegraphics[height=1.9cm,trim=0 0 0 0,clip]{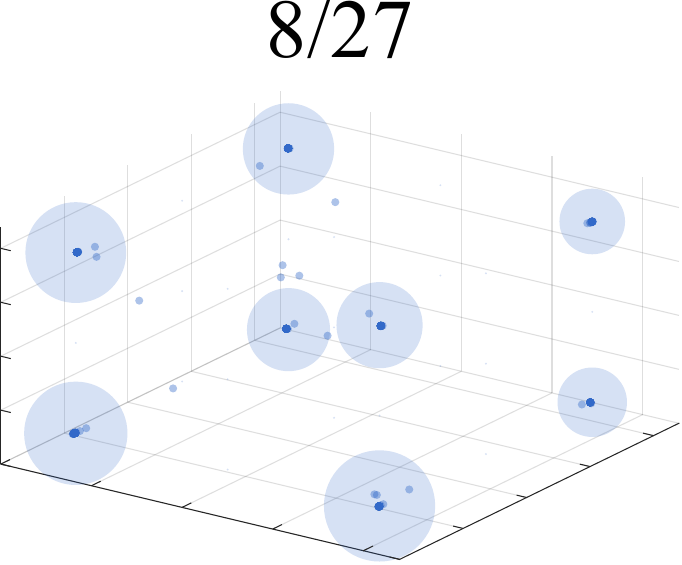} & \includegraphics[height=1.9cm,trim=0 0 0 0,clip]{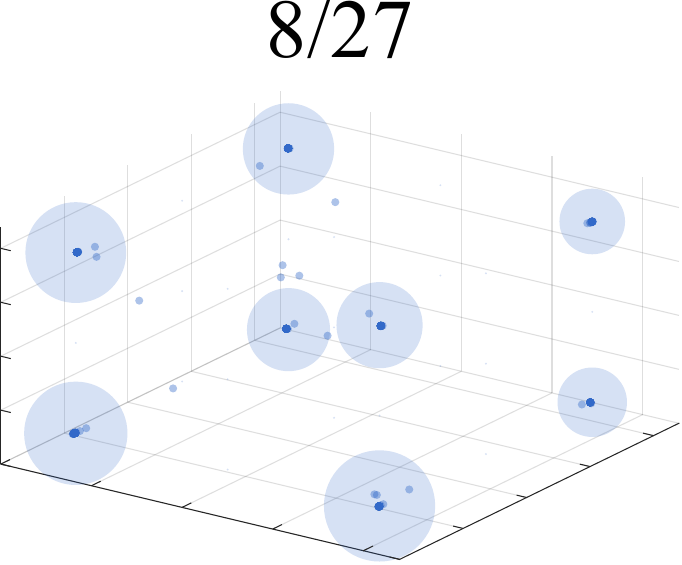} & & \multirow{6}{*}{\textbf{LwCL}}\\
			& \includegraphics[height=1.9cm,trim=0 0 0 0,clip]{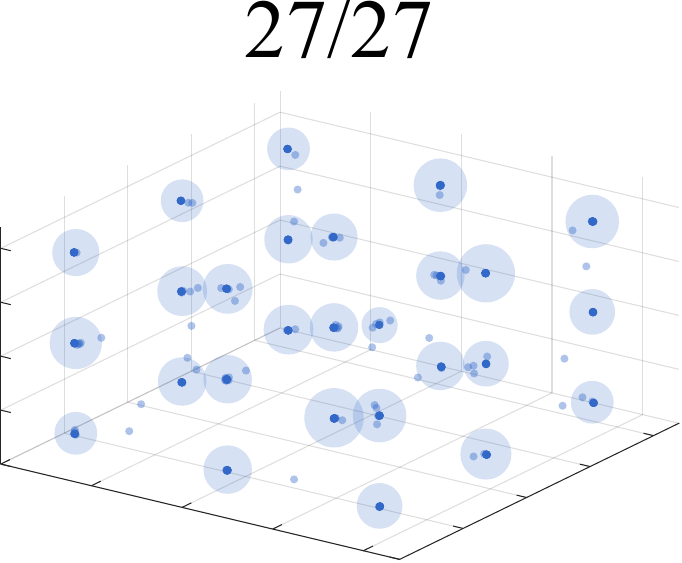} & \includegraphics[height=1.9cm,trim=0 0 0 0,clip]{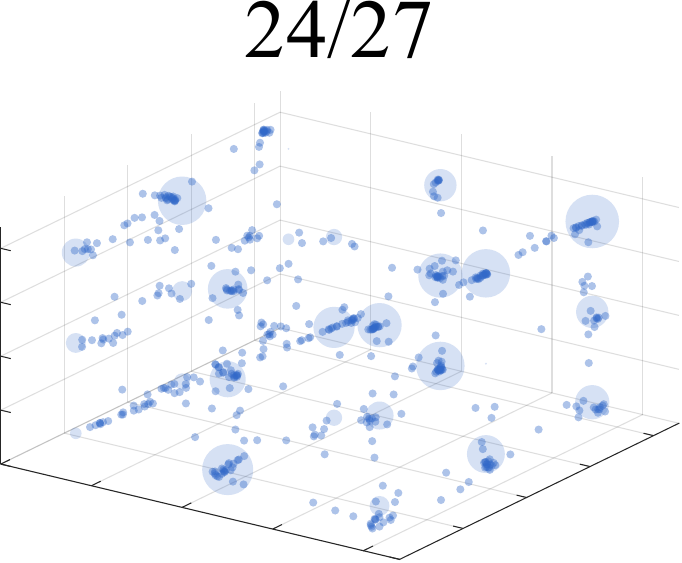} &\includegraphics[height=1.9cm,trim=0 0 0 0,clip]{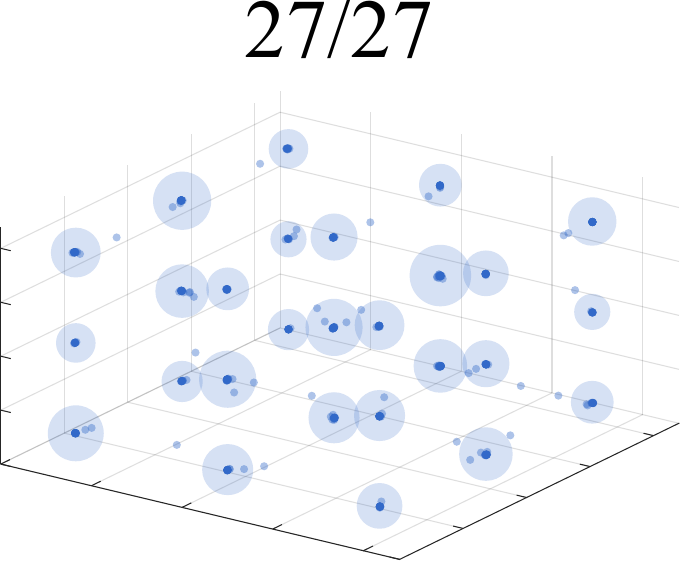} &  \includegraphics[height=1.9cm,trim=0 0 0 0,clip]{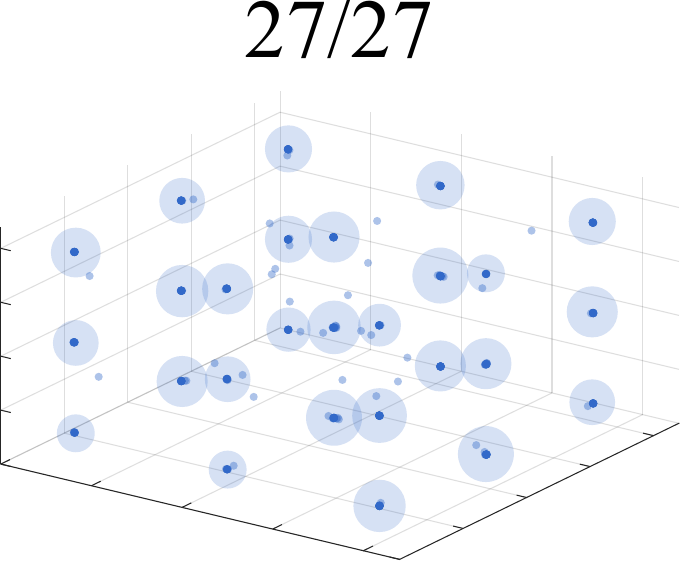} & & \\ 
			\multicolumn{7}{c}{\textbf{3D Cube}} \\ 
			\hline \toprule[0.65pt]
		\end{tabular}
	\end{center}
	\vspace{-0.2cm}
	\caption{Comparison results  among four mainstream GAN methods on two synthetic MOG distribution datasets with naive alternating learning strategy and our LwCL methodology (aptly abbreviated NAL or LwCL). Heatmaps for all distributions are generated through the same iterative steps. The diversity of generated samples (number of generated classes/number of target classes) is listed at the top. All experiments were terminated when their performance stabilized. The shading of the dots represents the density of the final distribution, with darker dots representing greater density.}
	\label{fig:syn_mode}
\end{figure*}  

\begin{table*}[htbp]
	\begin{center}  \footnotesize
		\renewcommand{\arraystretch}{1.2}
		\caption{Comparison results on mainstream GANs (i.e., VGAN, LCGAN, WGAN, and ProxGAN) with or without LwCL. We report the average FID$\downarrow$, JSD$\downarrow$ and Mode$\uparrow$ scores on four MOG synthesized datasets, i.e., 2D Ring, 2D Rand, 2D Grid and 3D Cube. The best result is in red whereas the second best one is in blue.}
		\label{tab:syn_metric}%
		\setlength{\tabcolsep}{4.5mm}{
			\begin{tabular}{|c|c|c|c|c|c|c|c|}
				\hline
				\multirow{2}[0]{*}{Method} & \multirow{2}[0]{*}{LwCL} &\multicolumn{3}{c|}{2D Ring (max mode=8)}&	\multicolumn{3}{c|}{2D Random (max mode=10)}\\
				\cline{3-5}
				\cline{6-8}		
				& & FID$\downarrow$   & JS$\downarrow$   & Mode$\uparrow$    & \footnotesize FID$\downarrow$   &  JS$\downarrow$  & Mode$\uparrow$\\		
				\hline
				\hline
				\multirow{2}[0]{*}{\footnotesize VGAN} & \xmark & 193.20$\pm$65.30 & 0.63$\pm$0.12 & 3.50$\pm$1.00& 77.58$\pm$71.49 & 8.74$\pm$0.02 & 3.75$\pm$0.71 \hspace{0.2in}\\
				%\cline{2-8}
				& \cmark & 34.16$\pm$10.23 & \textcolor{blue}{\textbf{0.40$\pm$0.16}} & \textcolor{red}{\textbf{7.50$\pm$1.00}}& 23.29$\pm$10.63 & 3.41$\pm$0.07 & \textcolor{red}{\textbf{7.75$\pm$0.96}}\\
				%\hline
				\multirow{2}[0]{*}{\footnotesize LCGAN} & \xmark & 15.01$\pm$15.30 & 0.63$\pm$0.12 & 3.50$\pm$1.00& 77.58$\pm$11.49 & 0.71$\pm$0.02 & 2.00$\pm$0 \\
				%\cline{2-8}
				& \cmark & 1.19$\pm$1.23 & \textcolor{blue}{\textbf{0.40$\pm$0.16}} & \textcolor{red}{\textbf{7.50$\pm$1.00}}& \textcolor{blue}{\textbf{23.29$\pm$0.63}} & \textcolor{red}{\textbf{0.33$\pm$0.07}} & \textcolor{blue}{\textbf{7.50$\pm$0.71}} \\
				%\hline
				\multirow{2}[0]{*}{WGAN} & \xmark  & \textcolor{blue}{\textbf{0.42$\pm$0.21}} & 0.65$\pm$0.13 & 6.25$\pm$1.53 & 97.98$\pm$0.10 & 0.75$\pm$0 & 1.73$\pm$1.63\\
				%\cline{2-8}
				& \cmark & \textcolor{red}{\textbf{0.16$\pm$0.09}} & \textcolor{red}{\textbf{0.27$\pm$0.25}} & \textcolor{blue}{\textbf{7.00$\pm$0.82}} & 97.30$\pm$1.49 & 0.70$\pm$0.06 &7.33$\pm$0.81\\
				%\hline
				\multirow{2}[0]{*}{ProxGAN} & \xmark & 15.01$\pm$15.30 & 0.63$\pm$0.12 & 3.50$\pm$1.00& 74.05$\pm$8.77&0.75$\pm$0.17&3.50$\pm$1.91\\
				%\cline{2-8}
				& \cmark & 8.82$\pm$17.25 & 0.58$\pm$0.26 & 6.50$\pm$3.00 & \textcolor{red}{\textbf{20.44$\pm$11.28}} & \textcolor{blue}{\textbf{0.36$\pm$0.01}} & \textcolor{blue}{\textbf{7.50$\pm$0.71}} \\
				\hline
				\hline
				\multirow{2}[0]{*}{Method} & \multirow{2}[0]{*}{LwCL} &\multicolumn{3}{c|}{2D Grid (max mode=25)}&	\multicolumn{3}{c|}{3D Cube (max mode=27)}\\
				\cline{3-5}
				\cline{6-8}	
				& & FID$\downarrow$   & JS$\downarrow$   & Mode$\uparrow$  &  FID$\downarrow$   &  JS$\downarrow$ &   Mode$\uparrow$\\
				\hline
				\hline
				\multirow{2}[0]{*}{VGAN} & \xmark &  10.14 $\pm$1.36 & 0.79$\pm$0.24  & 12.50$\pm$1.00    &  12.28$\pm$26.10 & \textcolor{blue}{\textbf{0.48$\pm$0.08}} & 8.50$\pm$1.00\\
				%\cline{2-8}
				& \cmark   & 0.81$\pm$0.24 & \textcolor{blue}{\textbf{0.62$\pm$0.13}}  & \textcolor{blue}{\textbf{18.5$\pm$1.00}}    & \textcolor{red}{\textbf{0.528$\pm$0.16}} & 0.61$\pm$0.30  & \textcolor{blue}{\textbf{23.00$\pm$1.40}}  \\
				%\hline
				\multirow{2}[0]{*}{LCGAN} & \xmark  & 50.73$\pm$47.08 & 0.64$\pm$0.21 & 11.50$\pm$8.1& 68.80$\pm$60.85&0.87$\pm$0.17&6.00$\pm$3.16\\
				%\cline{2-8}
				& \cmark & \textcolor{red}{\textbf{0.42$\pm$0.13}} & 0.62$\pm$0.27 & 17.25$\pm$7.27& 30.5$\pm$45.81 & 0.70$\pm$0.30 & 17.33$\pm$14.15 \\
				%\hline
				\multirow{2}[0]{*}{WGAN} & \xmark  & 171.31$\pm$77.46 & 0.89$\pm$0.29   & 5.00$\pm$1.30 & 12.00$\pm$0.90 & 0.62$\pm$0.07 & 15.00$\pm$0.41 \\
				%\cline{2-8}
				& \cmark & 15.91$\pm$1.43 & 0.68$\pm$0.14 & \textcolor{red}{\textbf{18.50$\pm$0.58}} & \textcolor{blue}{\textbf{0.80$\pm$0.20}} & \textcolor{red}{\textbf{0.21$\pm$0.03}}   & \textcolor{red}{\textbf{24.00$\pm$0.51}}  \\
				%\hline
				\multirow{2}[0]{*}{ProxGAN} & \xmark & 38.29$\pm$47.08 & 0.67$\pm$0.21 & 11.5$\pm$8.10& 111.64$\pm$63.96 & 0.85$\pm$0.03 & 3.67$\pm$2.89 \\
				%\cline{2-8}
				& \cmark & \textcolor{blue}{\textbf{0.46$\pm$0.24}} & \textcolor{red}{\textbf{0.60$\pm$0.23}} & 18.25$\pm$7.63 & 52.10$\pm$86.98 & 0.69$\pm$0.30 & 17.67$\pm$14.47\\			
				\hline
			\end{tabular}%
		}
	\end{center}	
\end{table*}%

\begin{figure*}[htbp]
	\begin{center}
		\begin{tabular}{c@{\extracolsep{0.3em}}c@{\extracolsep{0.3em}}}
			\includegraphics[width=0.45\linewidth,trim=0 0 0 0,clip]{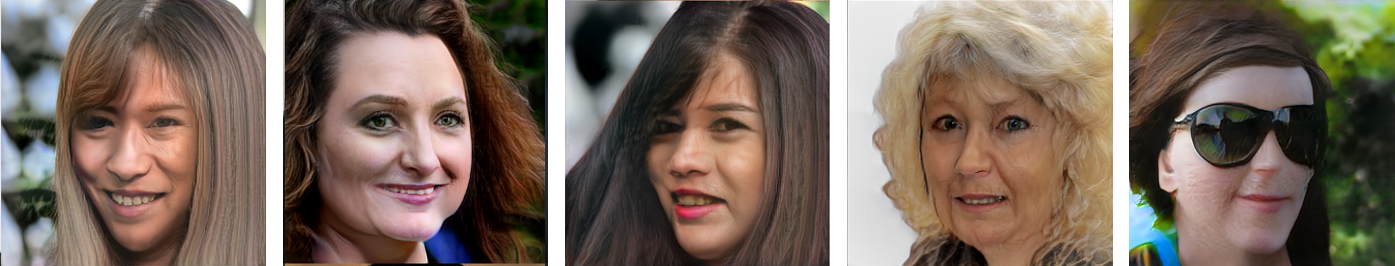}&\includegraphics[width=0.45\linewidth,trim=0 0 0 0,clip]{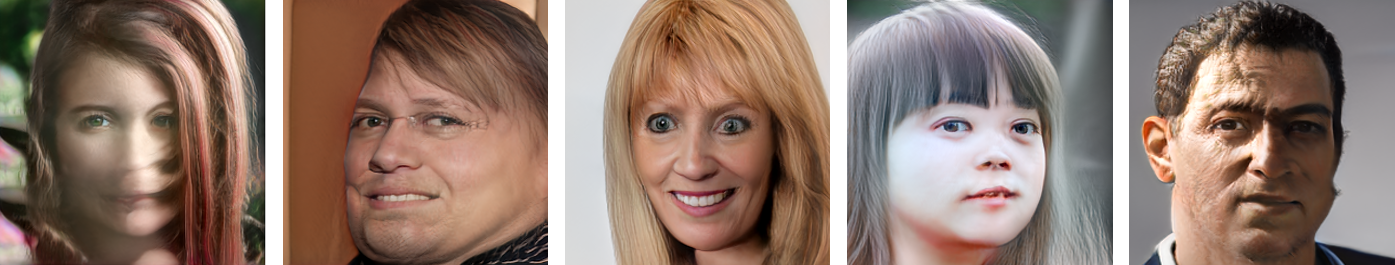}\\\specialrule{0em}{-2pt}{-3.5pt}			
			\multicolumn{2}{c}{\footnotesize StyleGAN} \\ 
			\includegraphics[width=0.45\linewidth,trim=0 0 0 0,clip]{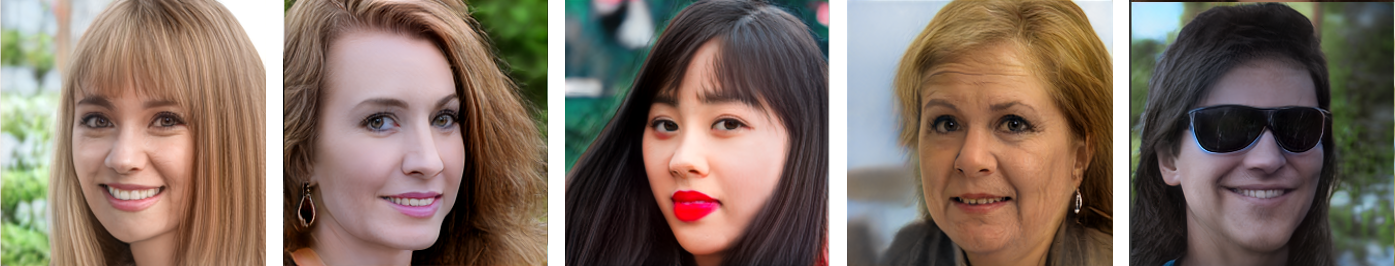}&\includegraphics[width=0.45\linewidth,trim=0 0 0 0,clip]{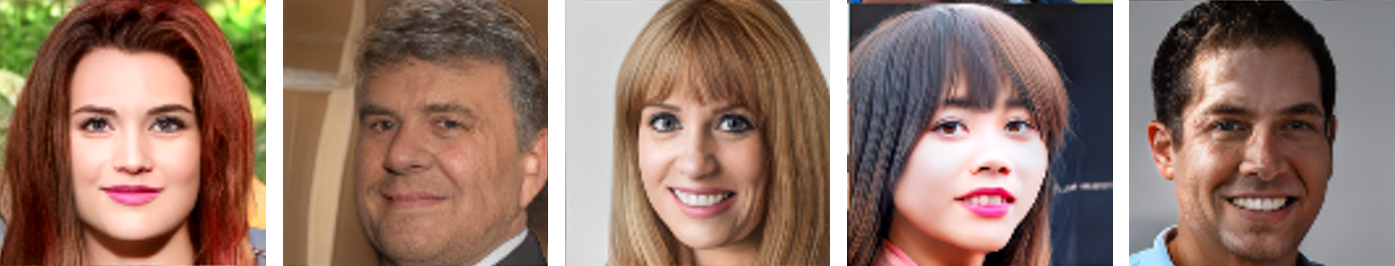}\\\specialrule{0em}{-2pt}{-3.5pt}	
			%\multicolumn{2}{c}{\footnotesize StyleGAN(DCL)}\\
			\multicolumn{2}{c}{\footnotesize Ours}\\
		\end{tabular}
	\end{center}
	\vspace{-0.4cm}
	\caption{Comparison results of face generation on CelebA-HQ dataset~\cite{liu2015deep}. 
		%regarding whether the StyleGAN network is retrained within our DCL framework. 
		Given the same latent feature variables, our method (i.e., StyleGAN with LwCL) is capable of generating more natural and realistic face details, with sharper and well visualized facial contours. }
	\label{fig:real_face}
\end{figure*}

%\subsection{Augment with Auxiliary}
\subsection{AL-type Applications}
%In this section, a series of experiments are conducted on each of the two technical ideas (i.e., \textit{auxiliary with discriminative tasks and auxiliary with related tasks}) to verify the effectiveness and flexibility of our methodology.
%\subsubsection{Auxiliary with Discriminative Tasks}
%In the following, we consider four types of applications (i.e., vanilla GAN, image generation, style transfer and imitation learning) 
%to verify the effectiveness and flexibility of our methodology.  Noteworthy, we demonstrate that even with different motivations and formulations, a variety of AL-type LwCL applications \textit{ALL} can be uniformly improved by our flexible methodology.

In the subsequent analysis, we explore four distinct applications, namely GAN and its variants, image generation, style transfer, and imitation learning, in order to validate the efficacy and versatility of our methodology. Importantly, we showcase that despite their diverse motivations and formulations, a wide range of AL-type LwCL applications, \textit{ALL} can be uniformly improved by our flexible methodology. 

\begin{table}[htbp]
	\begin{center}
		\renewcommand{\arraystretch}{1.2}
		\caption{ Comparison results of FID and IS score on CIFAR10 and CIFAR100 dataset.  The best result is in red whereas the second best one is in blue. }
		\label{tab:real_is}
		\setlength{\tabcolsep}{3.5mm}{
			\begin{tabular}{|c|c|c|c|c|c|}
				\hline  
				\multirow{2}[0]{*}{Method} &\multirow{2}[0]{*}{LwCL}&\multicolumn{2}{c|}{CIFAR10}&  \multicolumn{2}{c|}{CIFAR100}\\ 
				\cline{3-6}   		
				& &IS$\uparrow$&FID$\downarrow$ &IS$\uparrow$&FID$\downarrow$\\
				\hline 
				\hline
				\multirow{2}[0]{*}{DCGAN} & \xmark &6.63&49.03 &6.56&57.37    \\
				%\cline{2-6}  
				&\cmark &7.06&42.23 &6.87  & 44.18    \\
				% \hline 
				\multirow{2}[0]{*}{LSGAN} & \xmark &5.57&66.68 & 3.81&145.54 \\
				%\cline{2-6}  
				&\cmark &\textcolor{blue}{\textbf{7.54}}&32.50 & 7.30 & 35.72 \\
				%\hline 
				\multirow{2}[0]{*}{SNGAN} &\xmark &7.48&\textcolor{blue}{\textbf{26.51}} &\textcolor{blue}{\textbf{7.99}}&\textcolor{blue}{\textbf{25.33}}    \\ 
				%\cline{2-6}  			
				&\cmark  &\textcolor{red}{\textbf{7.58}}&\textcolor{red}{\textbf{22.81}}& \textcolor{red}{\textbf{8.27}}&\textcolor{red}{\textbf{21.37}}   \\
				\hline 
			\end{tabular}%
		}
	\end{center}	
\end{table}%

\begin{figure}[htbp]
	\begin{center}
		\begin{tabular}{c@{\extracolsep{0.3em}}}
			\includegraphics[width=0.95\linewidth,trim=0 0 0 0,clip]{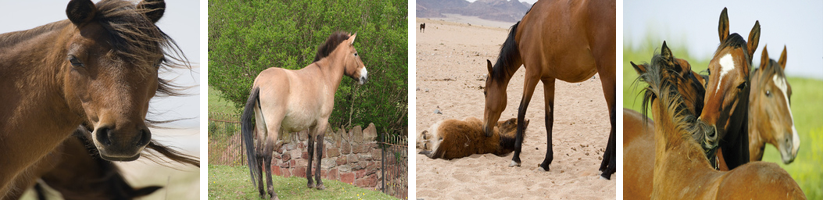}\\\specialrule{0em}{-2pt}{-3pt}
			\footnotesize Target\\
			\includegraphics[width=0.95\linewidth,trim=0 0 0 0,clip]{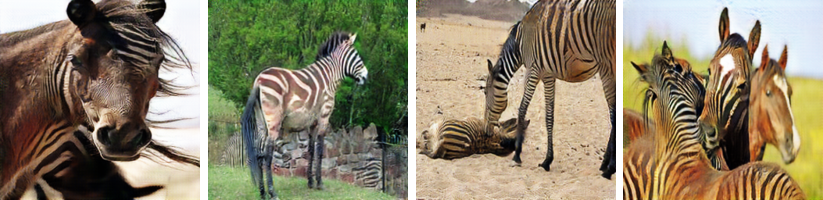}\\\specialrule{0em}{-2pt}{-3pt}
			\footnotesize CycleGAN\\ 
			\includegraphics[width=0.95\linewidth,trim=0 0 0 0,clip]{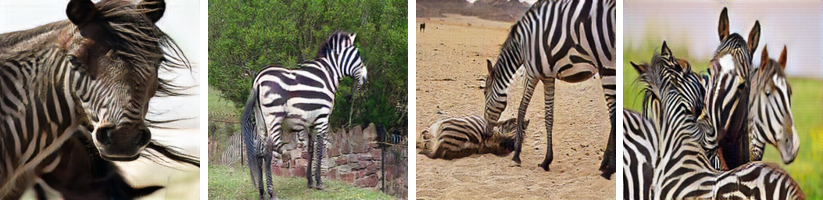}\\\specialrule{0em}{-2pt}{-3pt}
			%\footnotesize CycleGAN(DCL)\\
			\footnotesize Ours\\
		\end{tabular}
	\end{center}
	\vspace{-0.4cm}
	\caption{Comparition results for style transfer (i.e., \textit{horse $\rightarrow$ zebra}) on FFHQ~\cite{karras2019style}. 
		CycleGAN combining with our proposed LwCL methodology could generate more accurate transformation full of realistic textures. }
	\label{fig:real_transfer}
\end{figure}

\begin{figure}[htbp]
	\begin{center}
		\begin{tabular}{c@{\extracolsep{0.3em}}c@{\extracolsep{0.3em}}c@{\extracolsep{0.3em}}c@{\extracolsep{0.3em}}c@{\extracolsep{0.3em}}c@{\extracolsep{0.3em}}}
			\tiny\textit{Epoch 20} & \tiny\textit{Epoch 40} & \tiny\textit{Epoch 60} & \tiny\textit{Epoch 80} & \tiny\textit{Epoch 100} & \tiny\textit{Epoch 120} \\ \specialrule{0em}{-2pt}{-0.5pt}
			\includegraphics[width=0.15\linewidth,trim=0 0 0 0,clip]{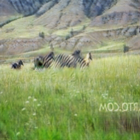}&\includegraphics[width=0.15\linewidth,trim=0 0 0 0,clip]{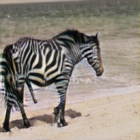}&\includegraphics[width=0.15\linewidth,trim=0 0 0 0,clip]{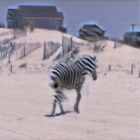}&\includegraphics[width=0.15\linewidth,trim=0 0 0 0,clip]{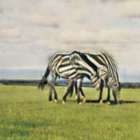}&\includegraphics[width=0.15\linewidth,trim=0 0 0 0,clip]{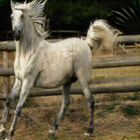}&\includegraphics[width=0.15\linewidth,trim=0 0 0 0,clip]{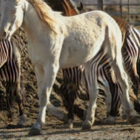}\\\specialrule{0em}{-2pt}{-3pt}
			\multicolumn{6}{c}{\footnotesize CycleGAN} \\ 
			\includegraphics[width=0.15\linewidth,trim=0 0 0 0,clip]{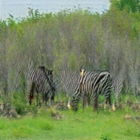}&\includegraphics[width=0.15\linewidth,trim=0 0 0 0,clip]{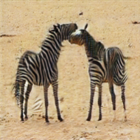}&\includegraphics[width=0.15\linewidth,trim=0 0 0 0,clip]{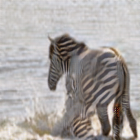}&\includegraphics[width=0.15\linewidth,trim=0 0 0 0,clip]{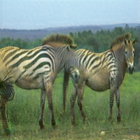}&\includegraphics[width=0.15\linewidth,trim=0 0 0 0,clip]{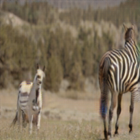}&\includegraphics[width=0.15\linewidth,trim=0 0 0 0,clip]{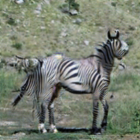}\\\specialrule{0em}{-2pt}{-3.5pt}
			%\multicolumn{6}{c}{ \footnotesize CycleGAN(DCL)} \\ 	
			\multicolumn{6}{c}{ \footnotesize Ours} \\ 		    
		\end{tabular}
	\end{center}
	\vspace{-0.4cm}
	\caption{Visualization comparison during training between standard CycleGAN and our method.  It can be observed that standard CycleGAN is not stable (with or without streaks), while the generation quality after introducing LwCL is significantly improved.}
	\label{fig:real_transfer_epoch}
\end{figure}

\subsubsection{GAN and Its Variants} 
Initially, we conduct extensive experiments on synthesized datasets following a Mixed of Gaussian (MOG) distribution. These experiments aim to provide a quantitative and qualitative evaluation of our algorithm, considering aspects such as mode generation, computational efficiency, and training stability. To establish a performance comparison, we benchmark our approach against several state-of-the-art GAN architectures, including VGAN~\cite{goodfellow2014generative}, WGAN~\cite{arjovsky2017wasserstein}, ProxGAN~\cite{farnia2020gans}, and LCGAN~\cite{engel2017latent}.  
For the synthetic data, we generate four distinct types of MOG distributions: 2D Ring (consisting of 5 or 8 2D Gaussians arranged in a ring), 2D Random (comprising 10 2D Gaussians with random magnitudes and positions), 2D Grid (comprising 25 2D Gaussians arranged in a grid), and 3D Cube (comprising 27 3D Gaussians within a cube). Each Gaussian distribution has a fixed variance of 0.02. During the training phase, we construct training batches with 512 samples from each mixture of Gaussian models, consisting of both real and generated data. Additionally, we sample 512 generated images for testing purposes. 

To optimize the two networks, we uniformly employ the Adam optimizer, with a learning rate of $10^{-4}$ for the discriminator and $10^{-3}$ for the generator. Both the generator and discriminator adopt a 3-layer linear network with a width of 256. The activation function utilized is a leaky ReLU with a threshold of 0.2.
To provide a comprehensive comparison, we employ three well-established metrics: Frechet Inception Distance (FID)~\cite{heusel2017gans}, Jensen-Shannon divergence (JS)~\cite{goodfellow2014generative}, number of Modes (Mode). These metrics serve as a basis for evaluating and contrasting the performance of different approaches. 
%It should be noted that in the synthetic experiments, we directly compute the data itself as a feature to measure FID, which requires the use of a pretrained network for feature extraction. 

\begin{figure}[htbp]
	\centering 
	\begin{tabular}{c@{\extracolsep{0.3em}}c@{\extracolsep{0.3em}}}
		\includegraphics[height=3.3cm,trim=1 0 1 0,clip]{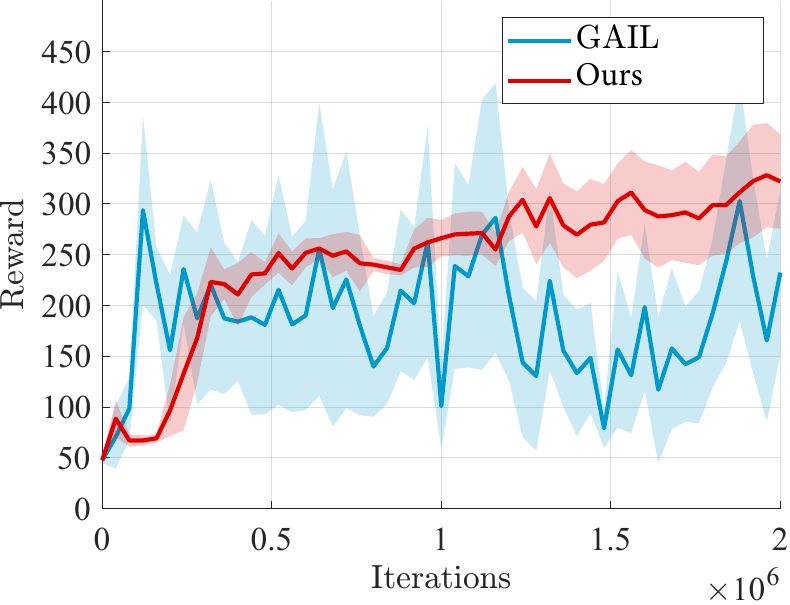} & \includegraphics[height=3.3cm,trim=1 0 0 0,clip]{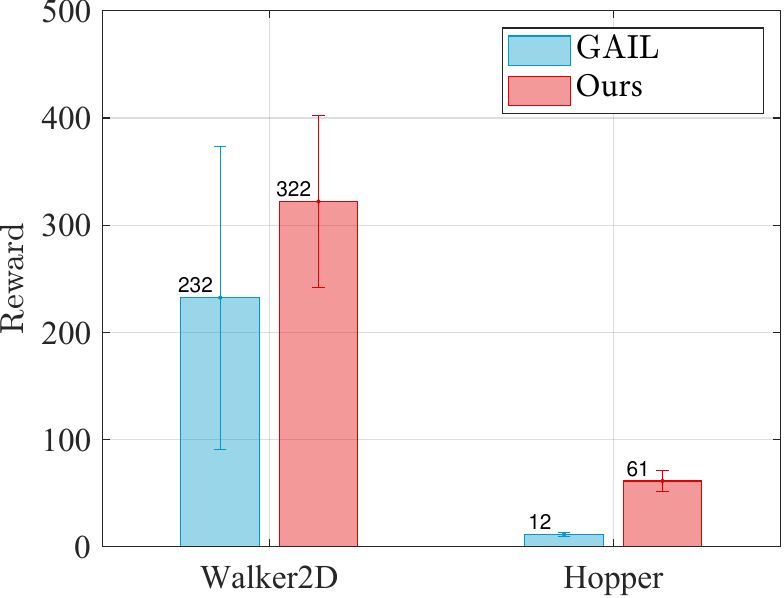}\\
	\end{tabular}%
	\caption{Comparison of reward scores  in terms of the training curve (\textit{left}), and final quantitative performance under two simulated environments, i.e., ``Walker2D'' and ``Hopper'' (\textit{right}). }\label{fig:gail}%
\end{figure}%

Fig.~\ref{fig:syn_mode} presents a comprehensive comparison of the number of generated samples among various advanced GAN methods, both with and without our LwCL methodology. When VGAN and LCGAN are combined with the NAL strategy, they exhibit a mapping of diverse inputs to the same output, resulting in limited capturing of different distributions. 
 It is evident that the original GAN models struggle to capture a significant number of distributions, leading to severe mode collapse and unsatisfactory performance. However, when integrated into our framework, these methods are able to capture a relatively larger number of distributions with the assistance of our approach, ultimately generating a more diverse range of realistic distributions. 
 In the case of the 3D Cube distribution, ProxGAN and LCGAN, when combined with our methodology, demonstrate the ability to fit almost all Gaussian distributions accurately, preserving intricate details. 
Tab.~\ref{tab:syn_metric} further demonstrates the effectiveness of our methodology in alleviating the mode collapse issue. Specifically, WGAN combined with LwCL achieves the lowest FID score in the 2D Ring distribution, and obtains the lowest JS score in the 3D Cube dataset. It is noteworthy that our flexible LwCL methodology uniformly improves a wide range of existing GANs, enhancing their overall performance.

\begin{figure*}[htbp]
	\begin{center}
		\begin{tabular}{c@{\extracolsep{0.3em}}}
			\includegraphics[width=17.5cm]{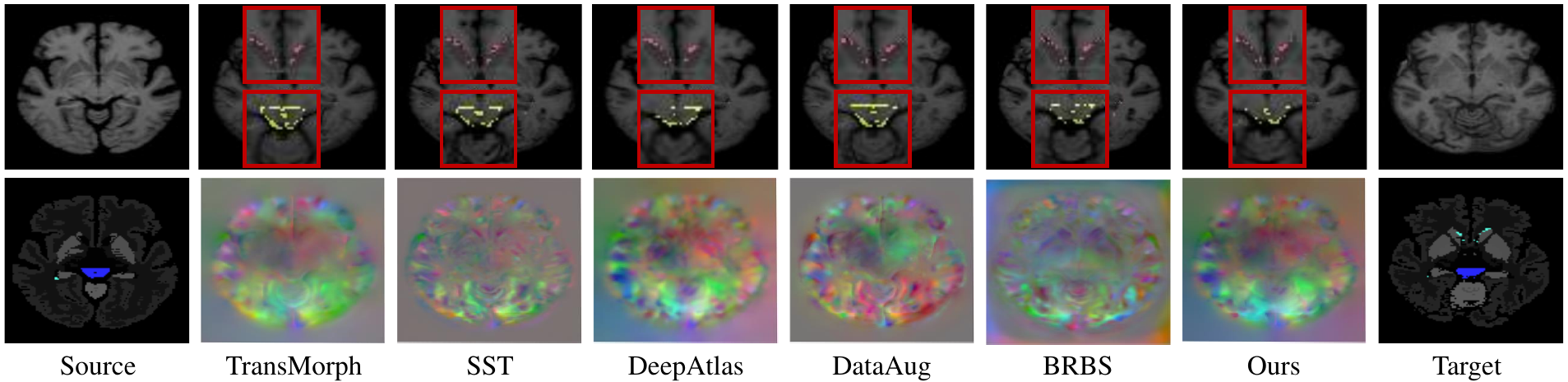} \\ 
		\end{tabular}
	\end{center}
	\vspace{-0.2cm}
	\caption{Visual results of state-of-the-art medical image registration methods for the lateral ventricle (LV) and brain stem (BS) with 2D visualization.}
	\label{fig:mis-0}
\end{figure*}

\begin{figure*}[htbp]
	\begin{center}
		\begin{tabular}{c@{\extracolsep{0.3em}}}
			\includegraphics[width=17.5cm]{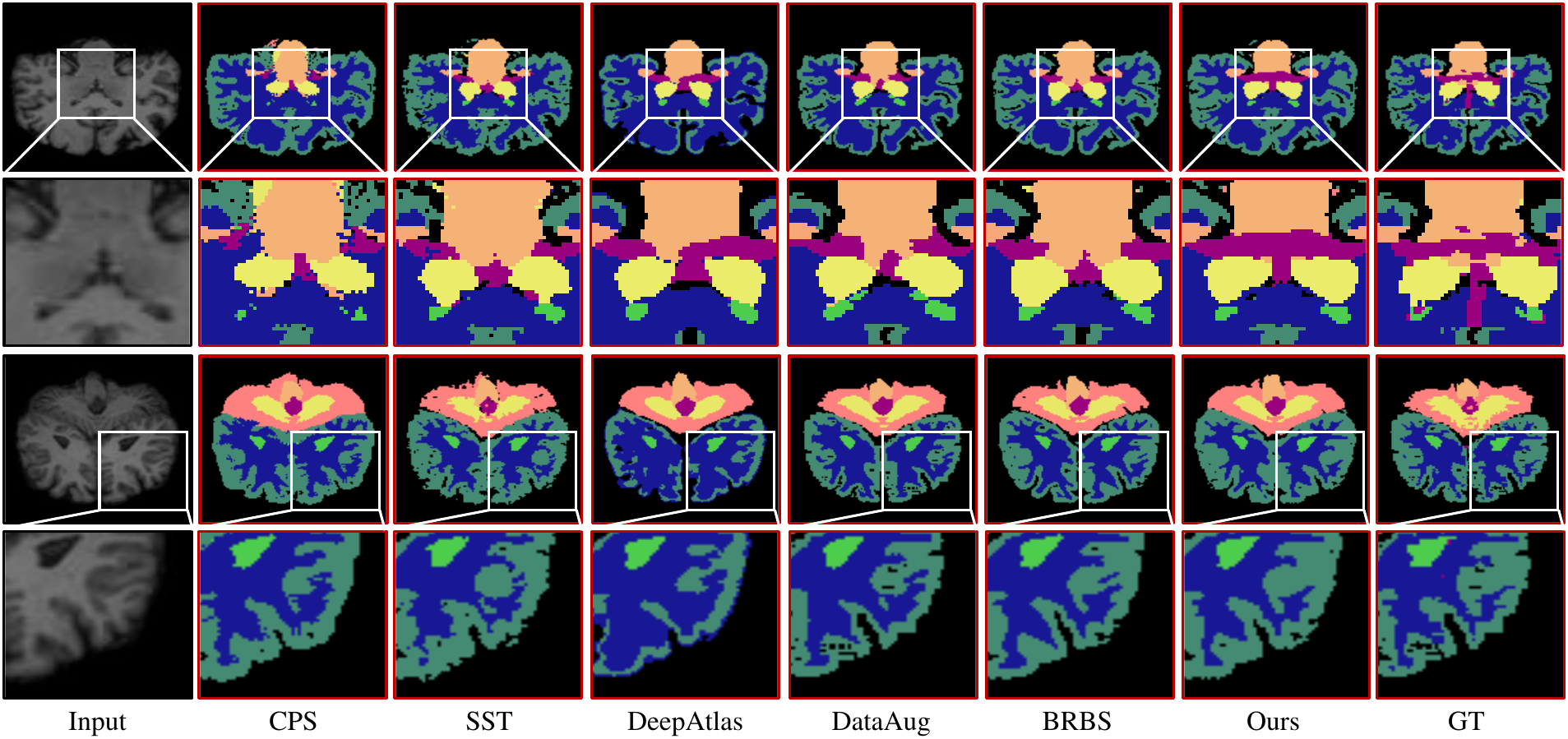} \\ 
		\end{tabular}
	\end{center}
	\vspace{-0.2cm}
	\caption{Visual results of state-of-the-art medical image segmentation methods for 2D slice of the brain. Our method ensure finer segmentation quality either in \textit{Top}: small brain structure 3rd/4th Ventricle (Ven) or in \textit{Bottom}: large brain structure Cerebral White Matter (CrWM).}
	\label{fig:mis-1}
\end{figure*}

\begin{table}[h!]
	%\scriptsize
	\begin{center}  \footnotesize
		\renewcommand{\arraystretch}{1.2}
		\caption{Comparison results of different methods for joint registration and segmentation tasks.}
		\label{tab:regseg}%
		\vspace{-0.2cm}
		\setlength{\tabcolsep}{3mm}{
			\begin{tabular}{|c|c|c|c|}
				\hline
				Method & Dice$\uparrow$ & HD95$\downarrow$ & ASD$\downarrow$\\
				\hline
				\hline
				\multicolumn{4}{|c|}{Medical Image Registration} \\			
				\hline
				Initial & 64.5 $\pm$ 6.0 & 2.99 $\pm$ 0.79 & 0.55 $\pm$ 0.15 \\
				%				\hline
				%				SyN  & 72.0 $\pm$ 3.0 & 2.99 $\pm$ 0.79 & 0.55 $\pm$ 0.15 \\
				%				NiftyReg & 74.1 $\pm$ 2.8 & 2.48 $\pm$ 0.53 & 0.47 $\pm$ 0.12\\
				%				% LDDMM\cite{beg2005computing} & 74.4 $\pm$ 2.5 & 2.73 $\pm$ 0.67 & 0.49 $\pm$ 0.13\\
				%				deedsBCV  & 75.6 $\pm$ 2.1 & 2.44 $\pm$ 0.51 & 0.45 $\pm$ 0.11\\
				\hline
				VxM  & 76.8 $\pm$ 1.5 & 2.39 $\pm$ 0.51 & 0.43 $\pm$ 0.09\\
				% VxM-diff\cite{DBLP:conf/miccai/DalcaBGS18} & 75.5 $\pm$ 1.9 & 2.63 $\pm$ 0.50 & 0.46 $\pm$ 0.10\\
				% ViT-V-Net\cite{DBLP:journals/corr/abs-2104-06468} & 77.3 $\pm$ 1.6 & 2.42 $\pm$ 0.48 & 0.43 $\pm$ 0.09\\
				LKU-Net  & 77.2 $\pm$ 1.5 & \textbf{\textcolor{blue}{2.28 $\pm$ 0.47}} & 0.41 $\pm$ 0.09\\
				TransMorph  & 77.9 $\pm$ 1.3 & 2.40 $\pm$ 0.48 & 0.42 $\pm$ 0.09\\
				\hline
				SST  & 74.2 $\pm$ 2.6 & 2.89 $\pm$ 0.58
				& 0.51$\pm$ 0.12\\
				DeepAtlas  & 76.9 $\pm$ 2.1 & 2.58 $\pm$ 0.57 & 0.46 $\pm$ 0.13 \\
				DataAug & 78.1 $\pm$ 1.9 & 2.34 $\pm$ 0.54 & \textbf{\textcolor{blue}{0.40 $\pm$ 0.11}} \\
				BRBS  & \textbf{\textcolor{blue}{80.1 $\pm$ 1.9}} & 2.35 $\pm$ 0.53 & 0.42 $\pm$ 0.13 \\
				\hline
				Ours & \textbf{\textcolor{red}{80.3 $\pm$ 1.4}} & \textbf{\textcolor{red}{2.23 $\pm$ 0.49}} &\textbf{\textcolor{red}{ 0.39 $\pm$ 0.10}} \\
				\hline
				\hline
				\multicolumn{4}{|c|}{Medical Image Segmentation} \\
				\hline
				UNet & 58.1 $\pm$ 7.5 & 12.8 $\pm$ 3.23 & 2.47 $\pm$ 0.98\\
				MASSL  & 66.2 $\pm$ 4.5 & 10.7 $\pm$ 2.95 & 1.79 $\pm$ 0.57\\
				CPS & 73.1 $\pm$ 4.2 & 3.66 $\pm$ 1.34 & 0.67 $\pm$ 0.19 \\
				\hline
				SST & 76.5 $\pm$ 2.2 & 2.93 $\pm$ 0.72 & 0.55 $\pm$ 0.18 \\
				DeepAtlas & 77.8 $\pm$ 1.7 & 2.89 $\pm$ 0.62 & 0.53 $\pm$ 0.15\\
				DataAug & 78.9 $\pm$ 3.0 & 2.97 $\pm$ 0.46 & 0.56 $\pm$ 0.16\\
				BRBS & \textbf{\textcolor{blue}{82.3 $\pm$ 1.8}} & \textbf{\textcolor{blue}{2.84 $\pm$ 0.59}} & \textbf{\textcolor{blue}{0.50 $\pm$ 0.19}} \\
				% Ours(scale 3) & 78.2 $\pm$ 1.6 &  & \\
				\hline
				Ours & \textbf{\textcolor{red}{83.4 $\pm$ 1.3}}  & \textbf{\textcolor{red}{2.55 $\pm$ 0.54}} & \textbf{\textcolor{red}{0.43 $\pm$ 0.15}}\\			
				\hline
			\end{tabular}
		}
	\end{center}
	\vspace{-0.2cm}	 
\end{table}

%\subsection{Validation on Real-world Vision Tasks}
%In this part, we evaluate the flexibility and effectiveness of our framework on several cutting-edge research topics covering both low-level and high-level four representative applications, i.e., Image Generation, Style Transfer,  Reinforcement Learning and Low-light Enhancement.
%\subsubsection{Image Generation} 
\subsubsection{Image Generation}  
In our experimental evaluation, we examine the performance of several well-established generative models, namely DCGAN~\cite{gao2018deep}, LSGAN~\cite{mao2017least}, and SNGAN~\cite{miyato2018spectral}.  
To assess their capabilities, we employ widely used benchmark datasets, including CIFAR10~\cite{krizhevsky2010convolutional}, CIFAR100~\cite{krizhevsky2009learning} and CelebA-HQ~\cite{liu2015deep}. 
%The CIFAR10 and CIFAR100 datasets consist of 60,000 color images with dimensions of $32\times32$. Among them, 50,000 images are allocated for training, while the remaining 10,000 images serve as the test set. CIFAR10 encompasses 10 distinct classes, while CIFAR100 contains 100 classes. 
%CelebA-HQ is a large scale face attributes dataset, which consists of more than 200K celebrity images. 
For evaluating the generative models, we employ two widely recognized metrics: Inception Score (IS) for evaluating generation quality and diversity, and Fréchet Inception Distance (FID) for capturing the issue of mode collapse.  
%During the training process, DCGAN, LSGAN, and SNGAN undergo 200k, 200k and 20k iterations, respectively. 
%Additionally, the network architecture and other training details follow the settings of \cite{kang2021ReACGAN}. We adopt the network architecture and training configurations outlined in \cite{kang2021ReACGAN}. 
Under our LwCL framework, we consider DCGAN, which incorporates standard binary cross-entropy loss for unsupervised training and employs a coupled game process of OL and CL. The objective functions $\mathcal{F}^{\bm{\omega}}_{\mathtt{OL}}$ and $\mathcal{F}^{\bm{\theta}}_{\mathtt{CL}}$ for DCGAN are defined as follows: 
	$\mathcal{F}^{\bm{\omega}}_{\mathtt{OL}}:= E_{\mathbf{v}\sim \mathcal{N}_{(0,1)}}\left[\log(1-D(G(\mathbf{v})))\right]$, and $\mathcal{F}^{\bm{\theta}}_{\mathtt{CL}}:=E_{\mathbf{u}\sim P_{data}}\left[\log(D(\mathbf{u}))\right] 
	+ E_{\mathbf{v}\sim \mathcal{N}_{(0,1)}}\left[\log(1-D(G(\mathbf{v})))\right]$.  Similarly, for LSGAN, which employs a least squares loss, the objective functions $\mathcal{F}^{\bm{\omega}}_{\mathtt{OL}}$ and $\mathcal{F}^{\bm{\theta}}_{\mathtt{CL}}$ are constructed as follows: $\mathcal{F}^{\bm{\omega}}_{\mathtt{OL}}:= E_{\mathbf{v}\sim \mathcal{N}_{(0,1)}}\left[D(G(\mathbf{v}))-c\right]^2$, and $\mathcal{F}^{\bm{\theta}}_{\mathtt{CL}}:=E_{\mathbf{u}\sim P_{data}}\left[D(\mathbf{u})-b\right]^2 
	+ E_{\mathbf{v}\sim \mathcal{N}_{(0,1)}}\left[D(G(\mathbf{v}))-a\right]^2$. As for SNGAN, it incorporates spectral normalization to ensure the 1-Lipschitz continuity constraint. The objective function $\mathcal{F}^{\bm{\theta}}_{\mathtt{CL}}$ for SNGAN is defined as follows:  $\mathcal{F}^{\bm{\theta}}_{\mathtt{CL}}:=\sup_{||D||\leq1}E_{\mathbf{u}\sim P_{data}}\left[D(\mathbf{u})\right] 
	- E_{\mathbf{v}\sim \mathcal{N}_{(0,1)}}\left[D(G(\mathbf{v}))\right]$.  Furthermore, in our face generation experiment on the high-resolution CelebA-HQ dataset~\cite{liu2015deep}, we employ StyleGAN as the backbone architecture.

Tab.~\ref{tab:real_is} further highlights the consistent performance improvements achieved by state-of-the-art GAN architectures when incorporated into our LwCL framework. 
Moreover, Fig.~\ref{fig:real_face} visually demonstrates the efficacy of our LwCL methodology in conjunction with StyleGAN. It showcases the superior style-content trade-off achieved, validating the versatility and effectiveness of our flexible solution strategy. Notably, our approach excels in generating realistic facial structures while effectively mitigating twist distortion.
%For other training details, we follow the same settings as StyleGAN.

\begin{figure*}[t!]
	\begin{center}
		\begin{tabular}{c@{\extracolsep{0.3em}}c@{\extracolsep{0.3em}}c@{\extracolsep{0.3em}}c@{\extracolsep{0.3em}}c@{\extracolsep{0.3em}}c@{\extracolsep{0.3em}}} 
			\includegraphics[width=0.155\linewidth,trim=0 0 0 0,clip]{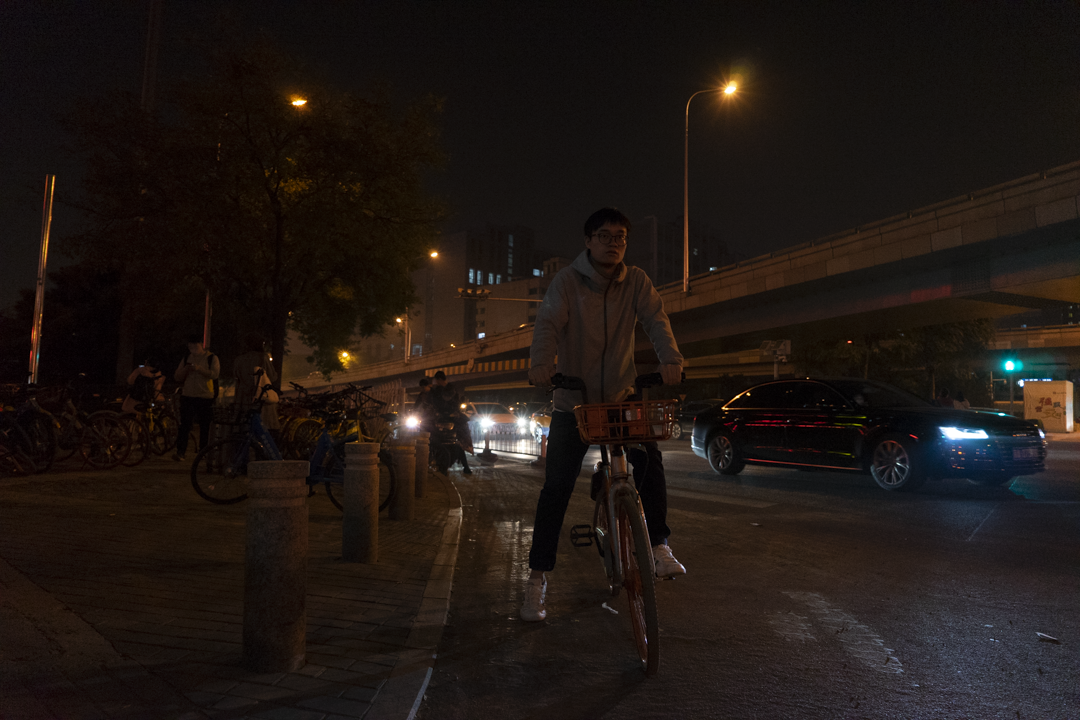}&\includegraphics[width=0.155\linewidth,trim=0 0 0 0,clip]{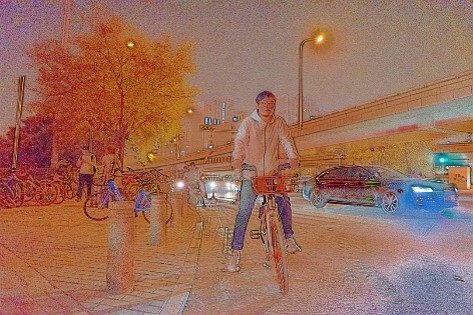}&\includegraphics[width=0.155\linewidth,trim=0 0 0 0,clip]{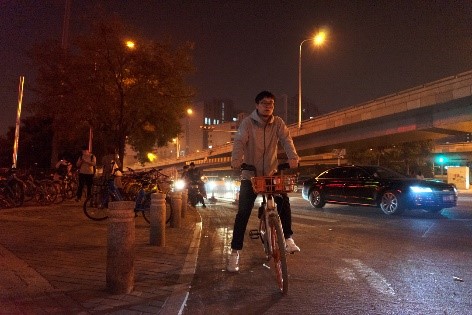}&\includegraphics[width=0.155\linewidth,trim=0 0 0 0,clip]{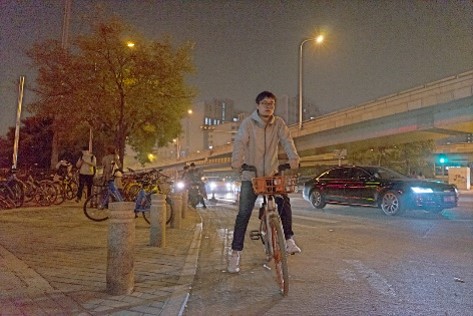}&\includegraphics[width=0.155\linewidth,trim=0 0 0 0,clip]{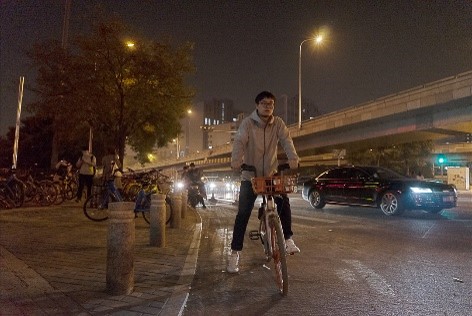}&\includegraphics[width=0.155\linewidth,trim=0 0 0 0,clip]{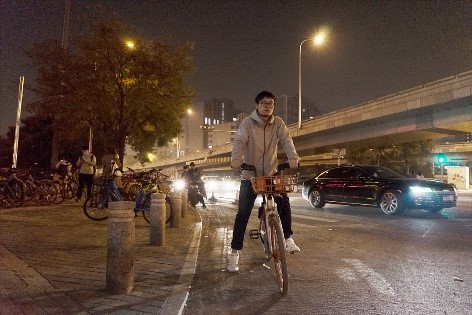}\\ \specialrule{0em}{-3pt}{-3pt}
			-&\footnotesize DE=6.710 & \footnotesize  DE=6.819 & \footnotesize  DE=6.817 & \footnotesize  DE=6.717 & \footnotesize  DE=7.106 \\ 					 
 	        \includegraphics[width=0.155\linewidth,trim=0 0 0 0,clip]{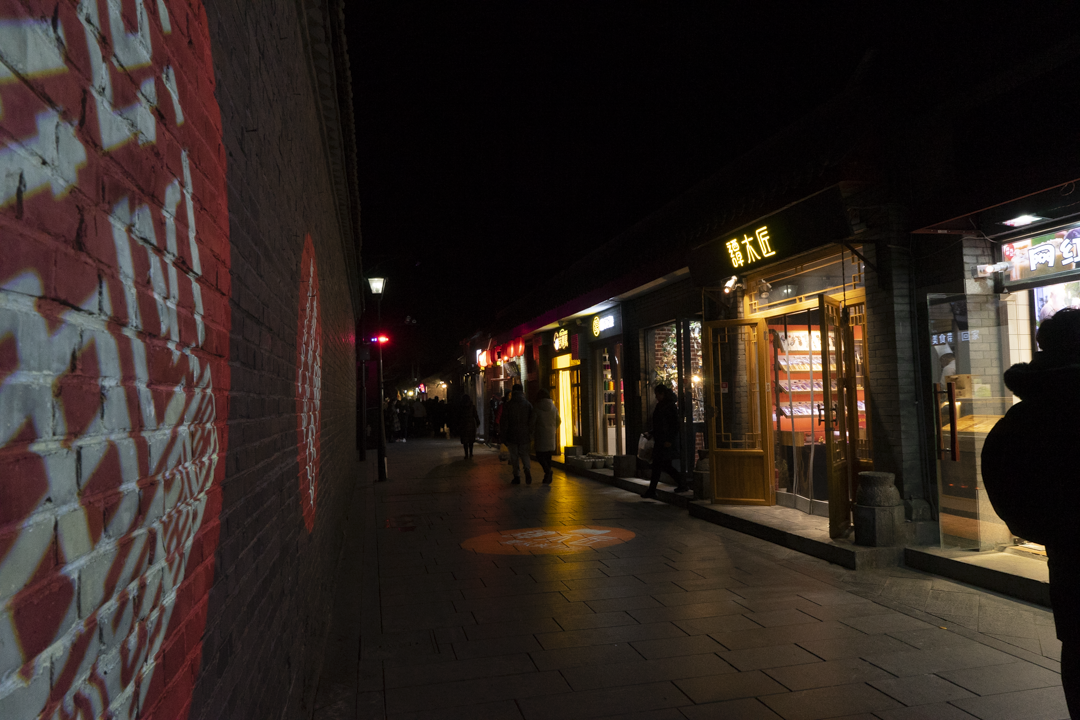}&\includegraphics[width=0.155\linewidth,trim=0 0 0 0,clip]{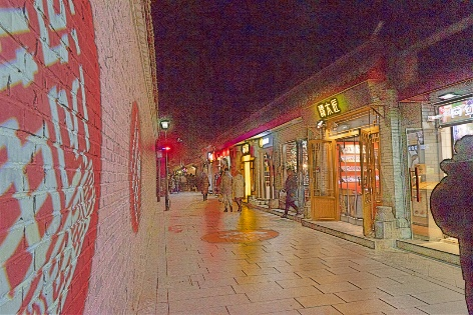}&\includegraphics[width=0.155\linewidth,trim=0 0 0 0,clip]{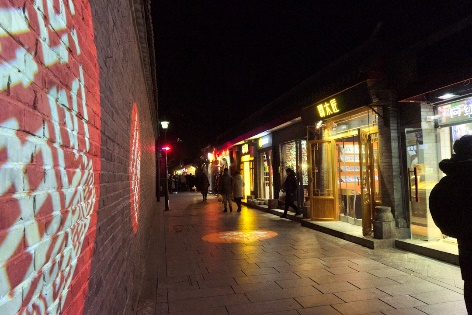}&\includegraphics[width=0.155\linewidth,trim=0 0 0 0,clip]{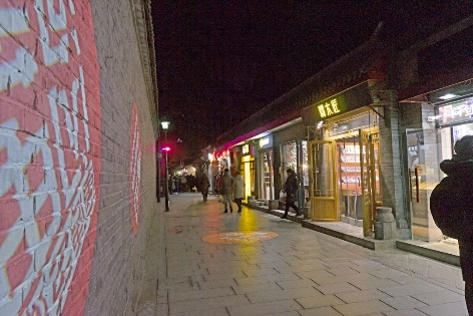}&\includegraphics[width=0.155\linewidth,trim=0 0 0 0,clip]{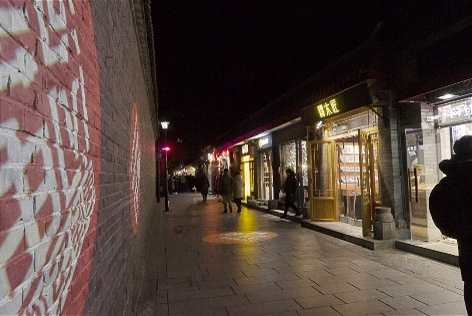}&\includegraphics[width=0.155\linewidth,trim=0 0 0 0,clip]{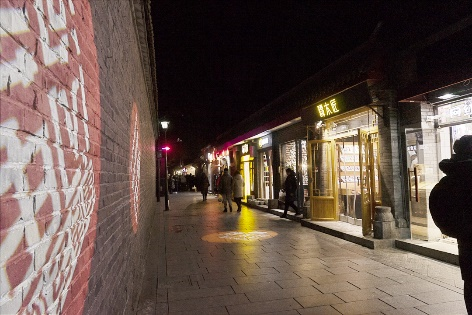}\\ \specialrule{0em}{-3pt}{-3pt}
			-&\footnotesize DE=7.331 & \footnotesize  DE=7.444 & \footnotesize  DE=7.334 & \footnotesize  DE=7.284 & \footnotesize  DE=7.537 \\ \specialrule{0em}{-0.5pt}{-1pt}	 
			\footnotesize Input & \footnotesize  RetinexNet & \footnotesize DeepUPE
			& \footnotesize  ZeroDCE
			 & \footnotesize  SCI & \footnotesize  Ours \\  
		\end{tabular}
	\end{center}
	\vspace{-0.2cm}
	\caption{Visual results of state-of-the-art low-light enhancement methods on the Darkface dataset.  Our method improves brightness for a more natural and realistic effect. 
		%The no-reference metric DE is used as the evaluated metric for unsupervised dataset.  
		Larger DE indicates more perceptually favored quality. Zoom-in regions are used to illustrate the visual differences. 
		%The best result is in red whereas the second best one is in blue.
	}
	\label{fig:real_darkface}\vspace{-0.2cm}
\end{figure*}

 \begin{figure*}[t!]
	\begin{center}
		\begin{tabular}{c@{\extracolsep{0.3em}}c@{\extracolsep{0.3em}}c@{\extracolsep{0.3em}}c@{\extracolsep{0.3em}}c@{\extracolsep{0.3em}}c@{\extracolsep{0.3em}}c@{\extracolsep{0.3em}}}
			\includegraphics[width=0.13\linewidth]{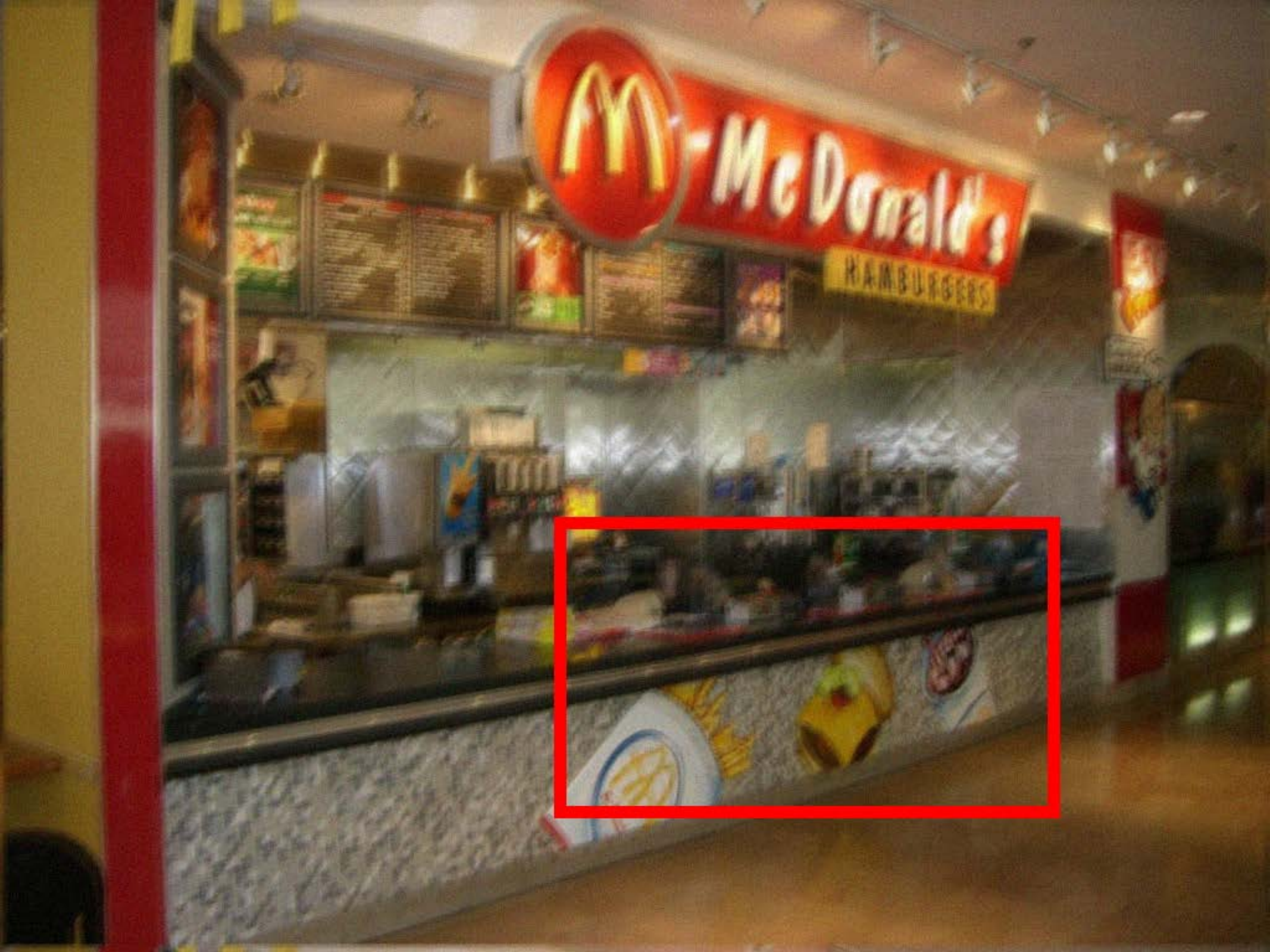}&\includegraphics[width=0.13\linewidth]{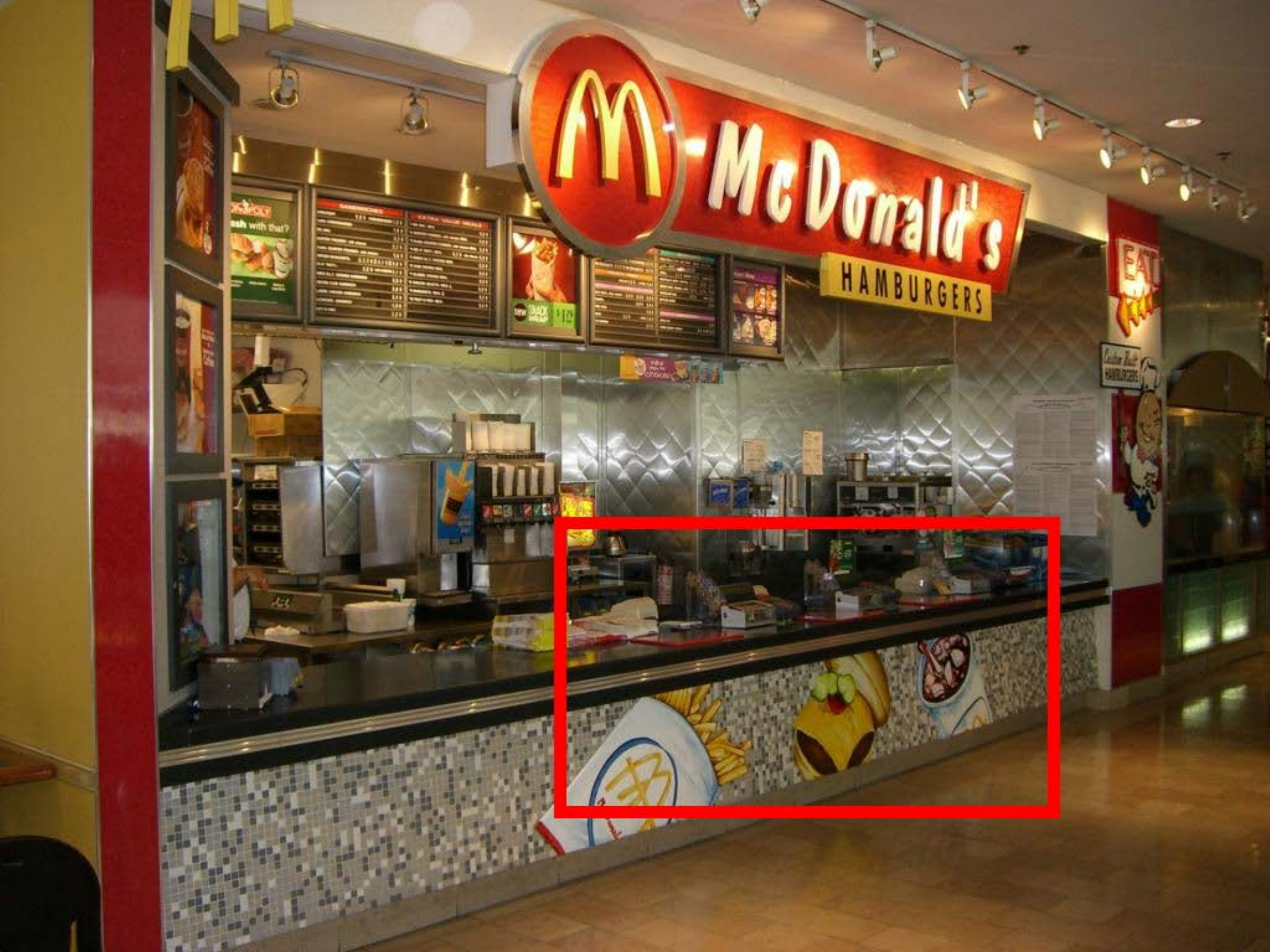}&\includegraphics[width=0.13\linewidth]{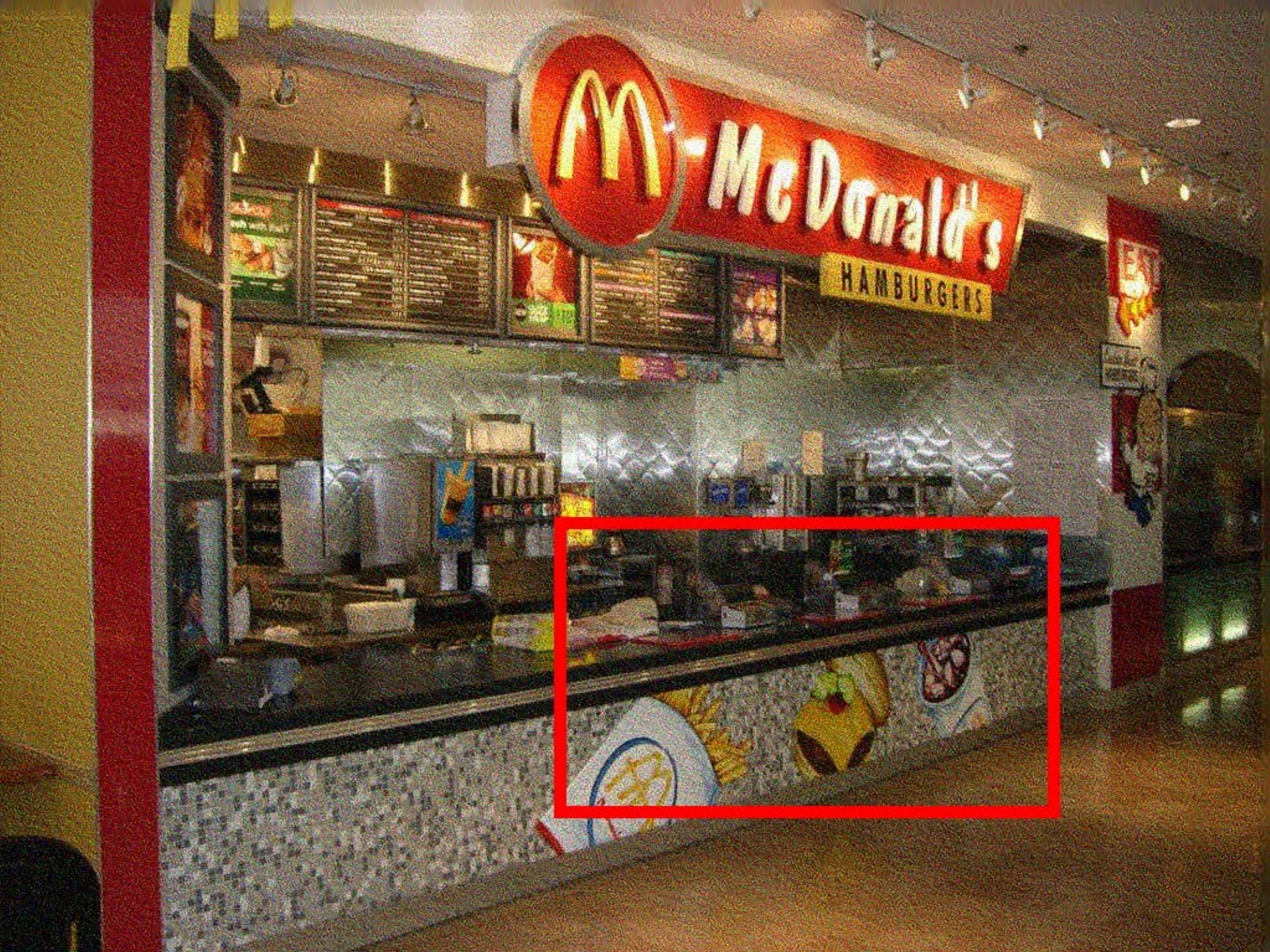}&\includegraphics[width=0.13\linewidth]{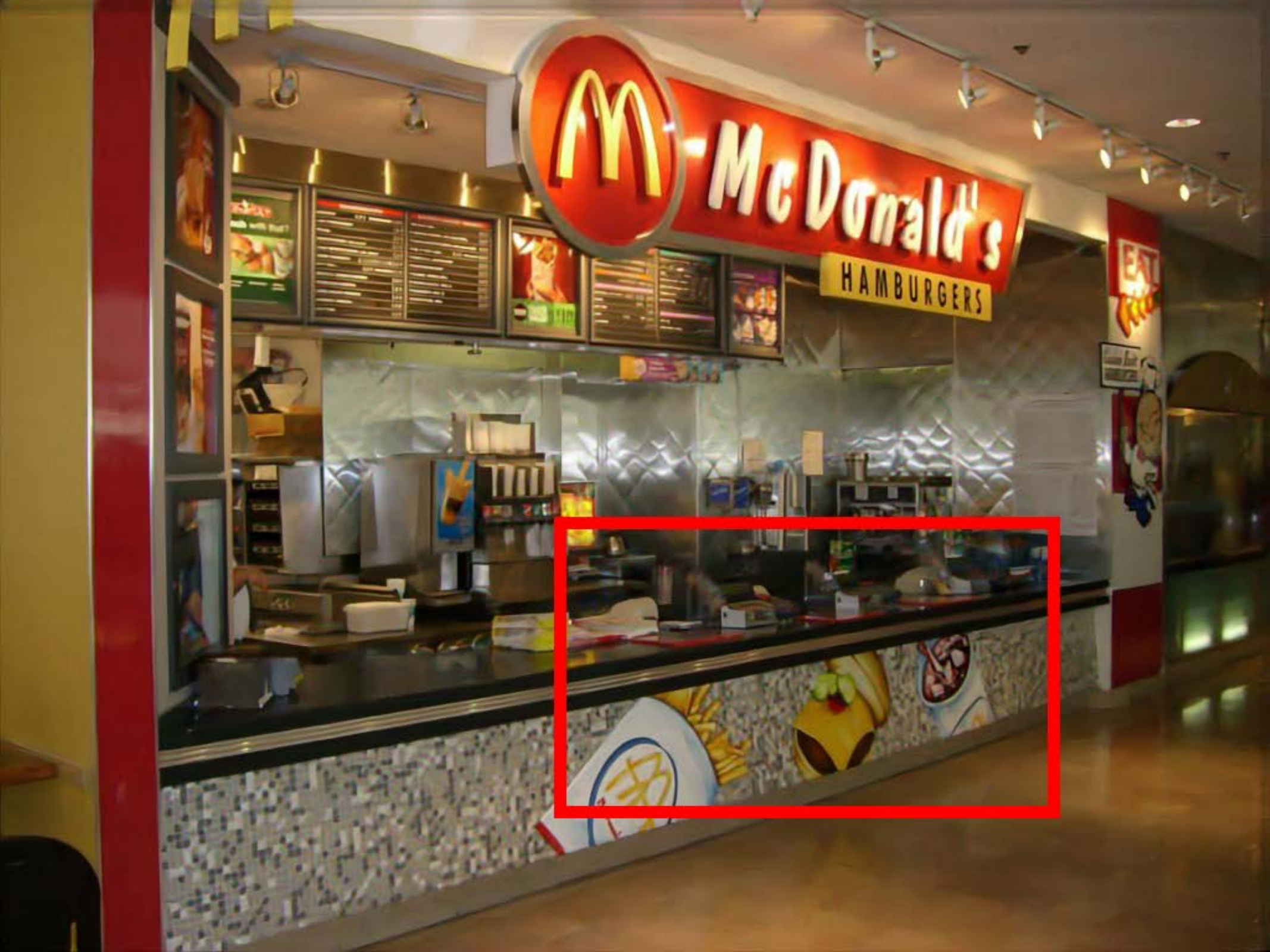}&\includegraphics[width=0.13\linewidth]{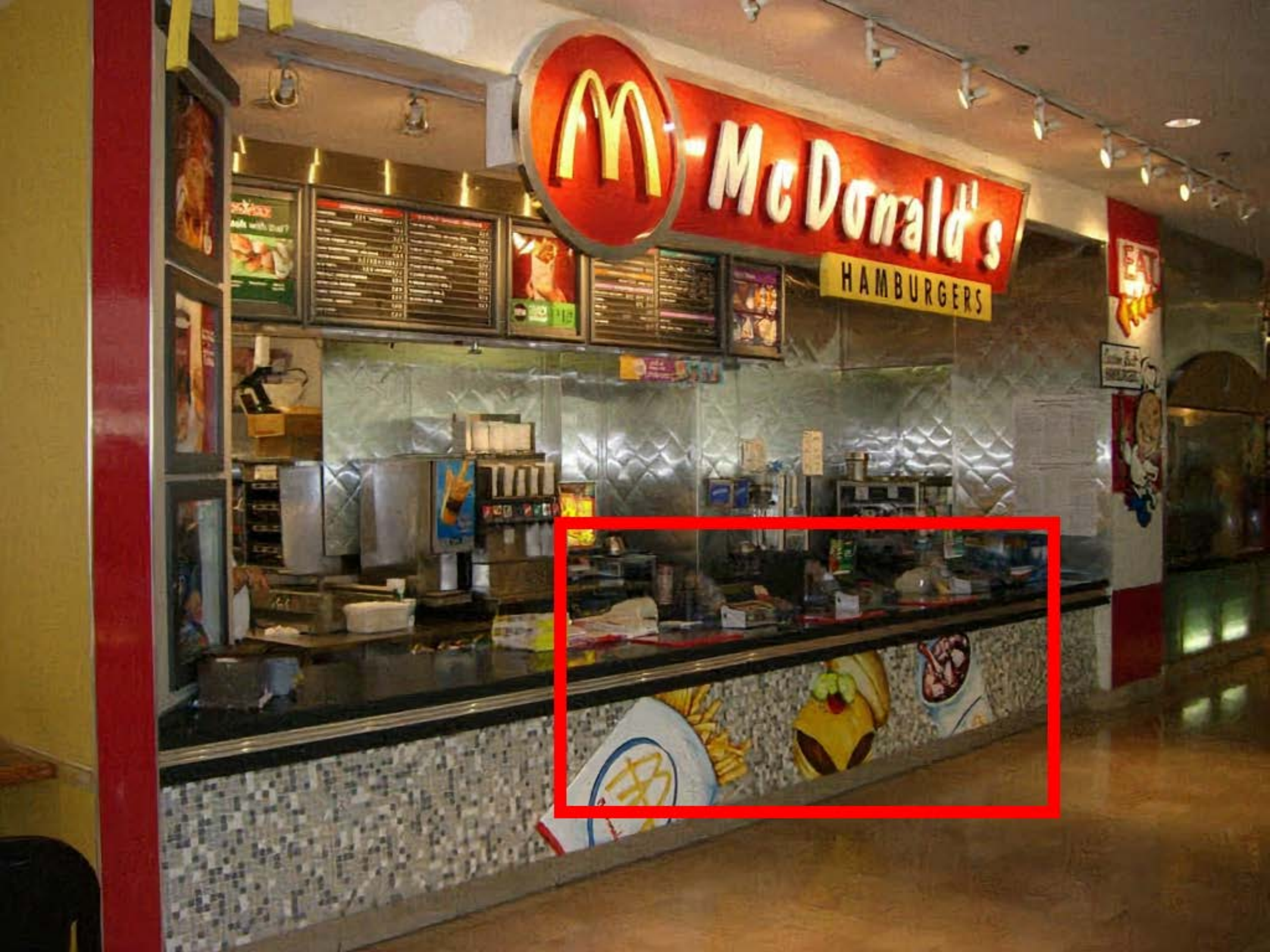}&\includegraphics[width=0.13\linewidth]{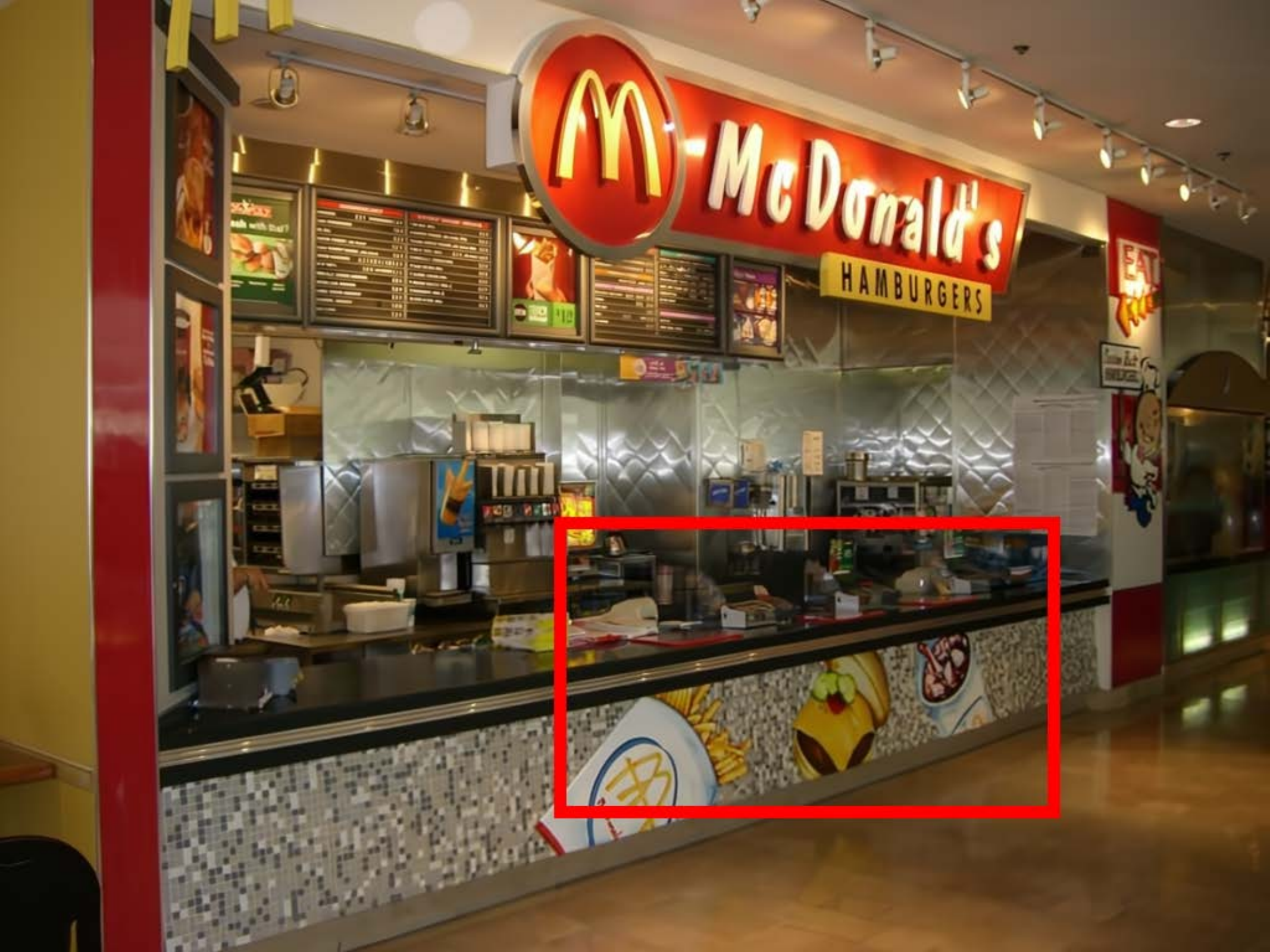} & \includegraphics[width=0.13\linewidth]{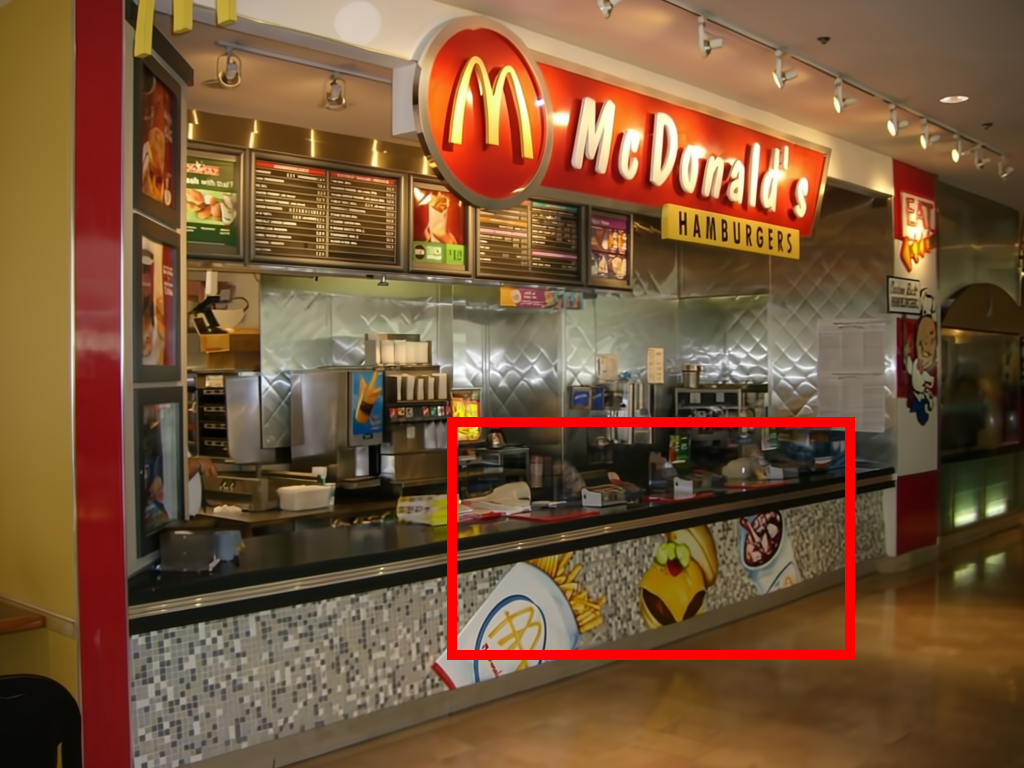} \\ \specialrule{0em}{-0.5pt}{-1pt}
			\includegraphics[width=0.13\linewidth]{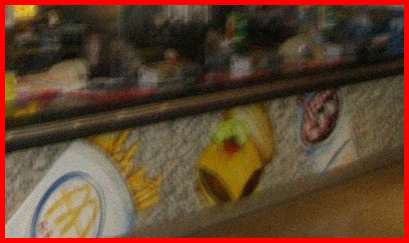}&\includegraphics[width=0.13\linewidth]{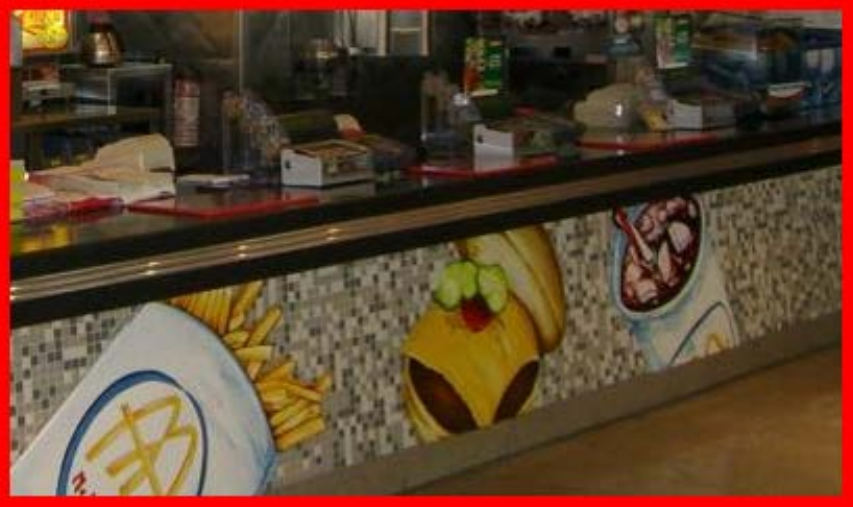}&\includegraphics[width=0.13\linewidth]{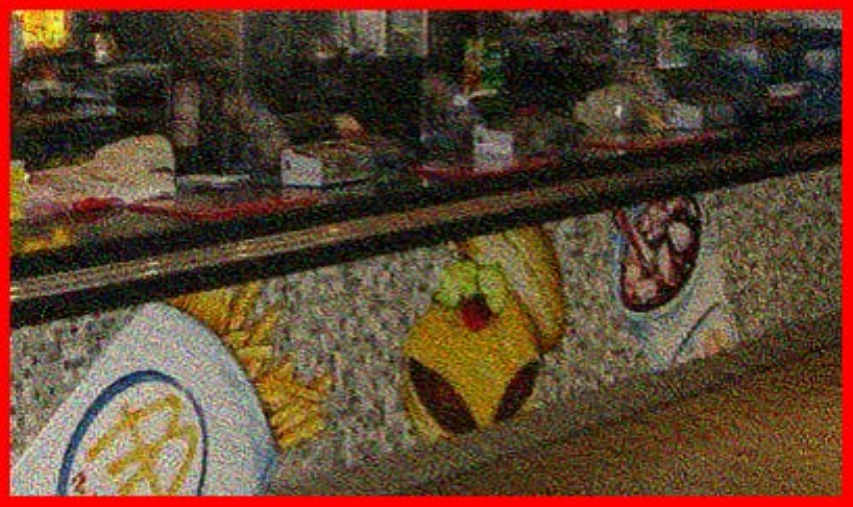}&\includegraphics[width=0.13\linewidth]{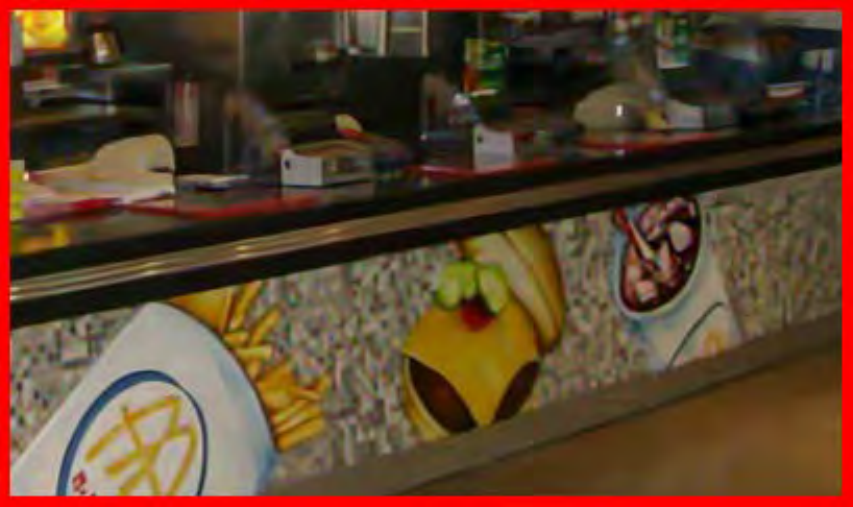}&\includegraphics[width=0.13\linewidth]{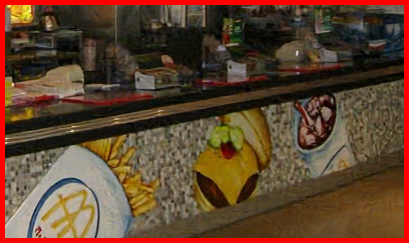}&\includegraphics[width=0.13\linewidth]{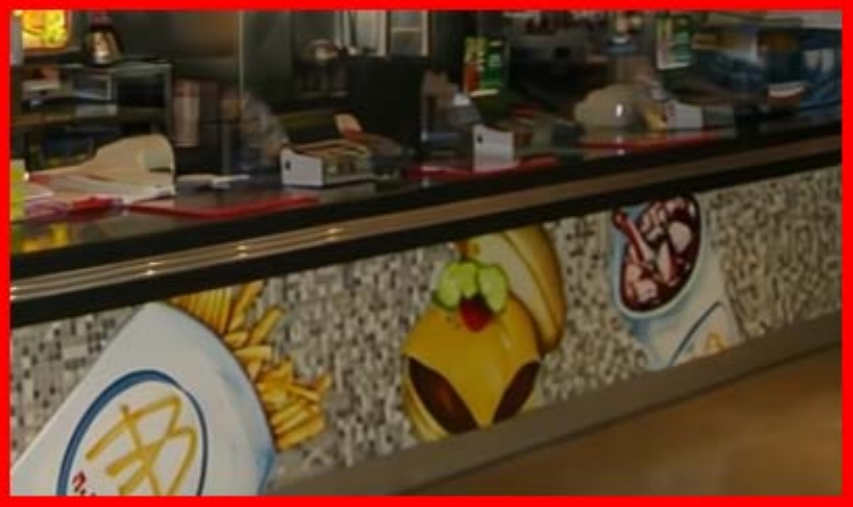} & \includegraphics[width=0.13\linewidth]{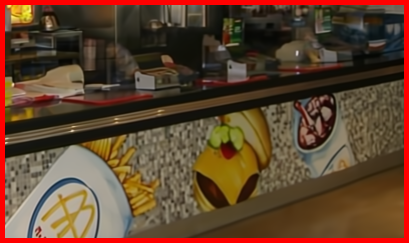} \\ \specialrule{0em}{-2pt}{-2pt} 
			\footnotesize -- & -- & \footnotesize (18.41, 0.61) & \footnotesize  (25.24, 0.86) & \footnotesize  (27.63, 0.91) & \footnotesize  (27.93, 0.91) & \footnotesize (28.01, 0.92)\\ 
			%\specialrule{0em}{-0.5pt}{-1pt} 
			%\footnotesize Input & \footnotesize  Ground Truth & \footnotesize FDN & \footnotesize  IRCNN & \footnotesize IRCNN & \footnotesize  DPIR & \footnotesize Ours\\  	

			\includegraphics[width=0.13\linewidth]{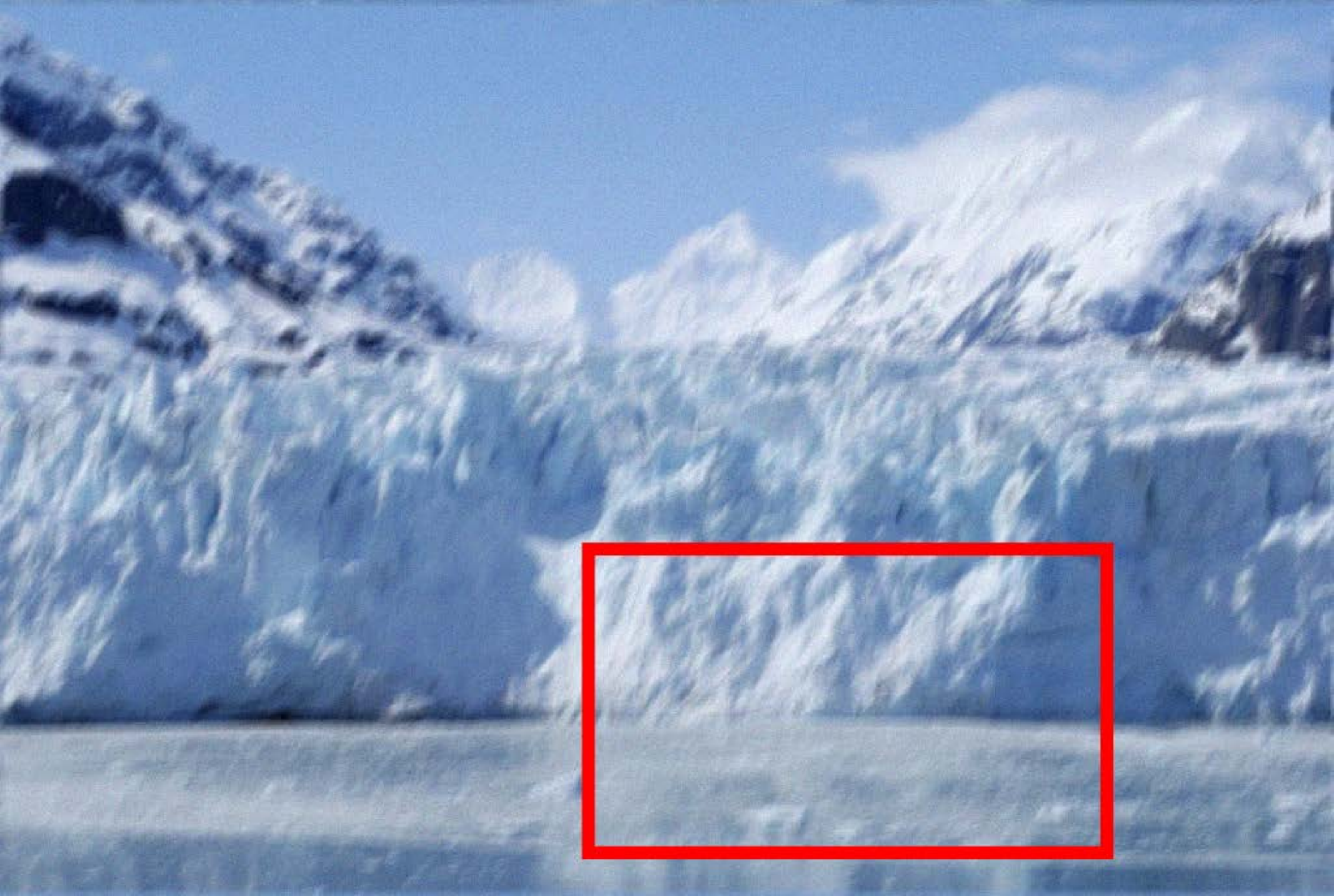}&\includegraphics[width=0.13\linewidth]{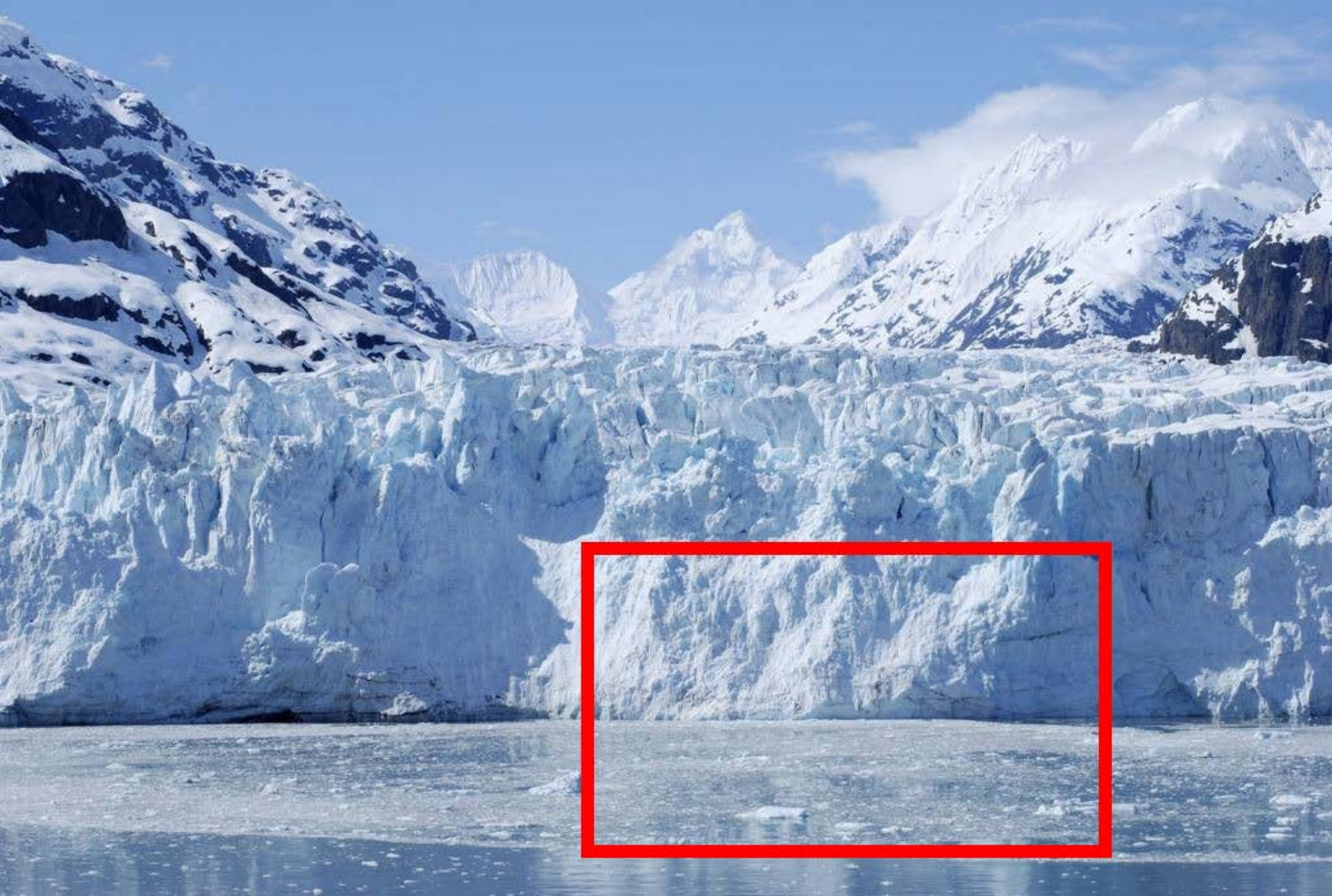}&\includegraphics[width=0.13\linewidth]{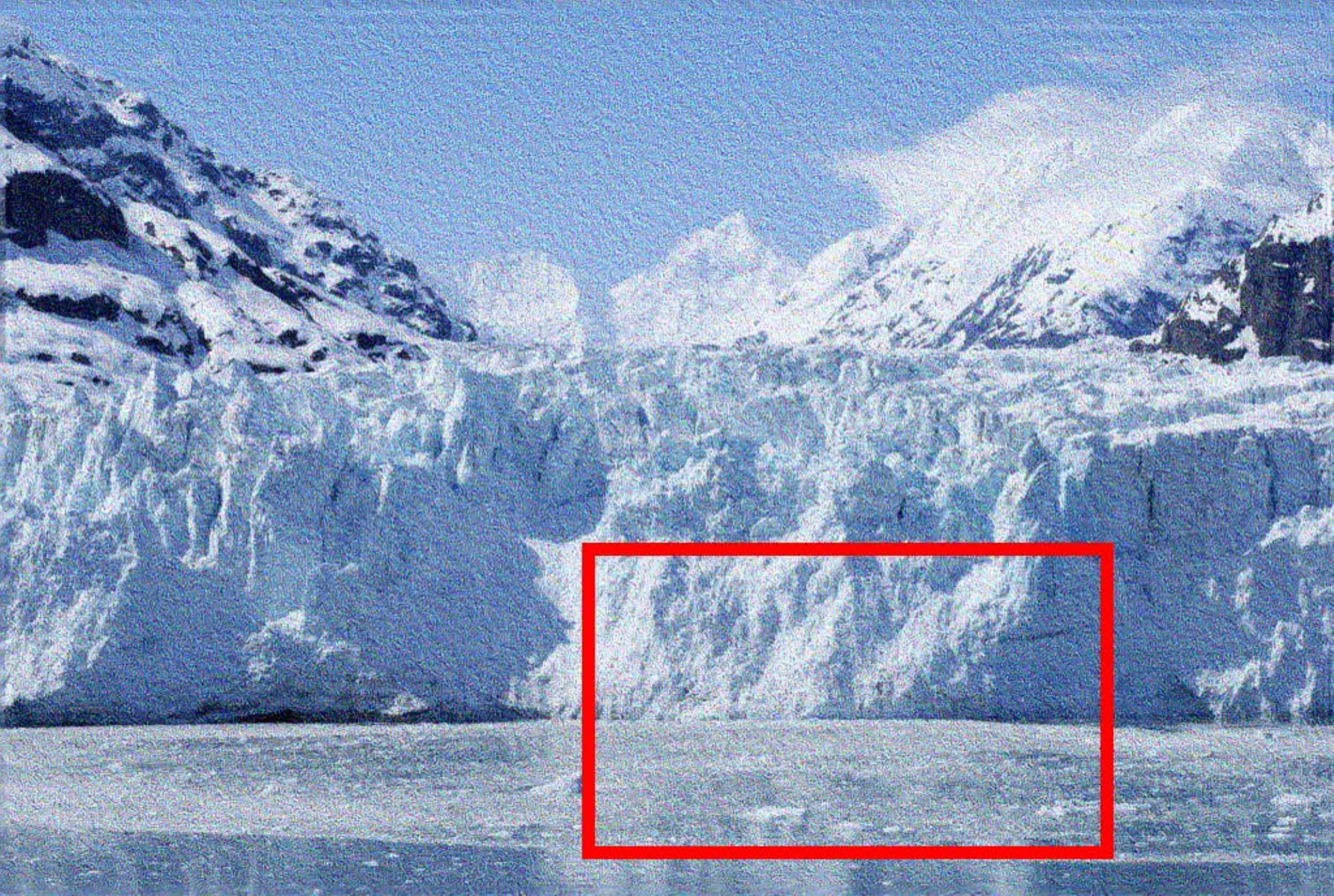}&\includegraphics[width=0.13\linewidth]{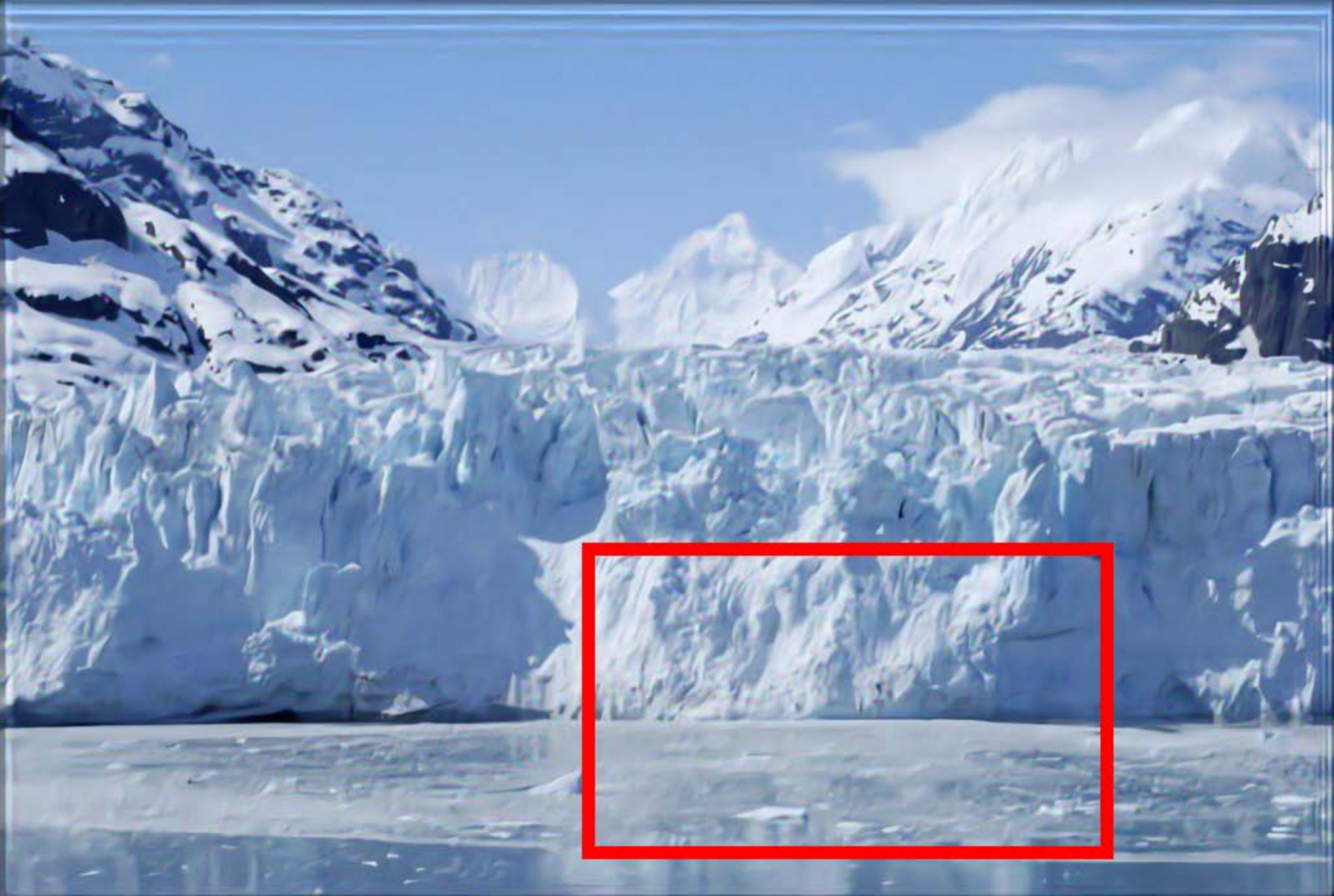}&\includegraphics[width=0.13\linewidth]{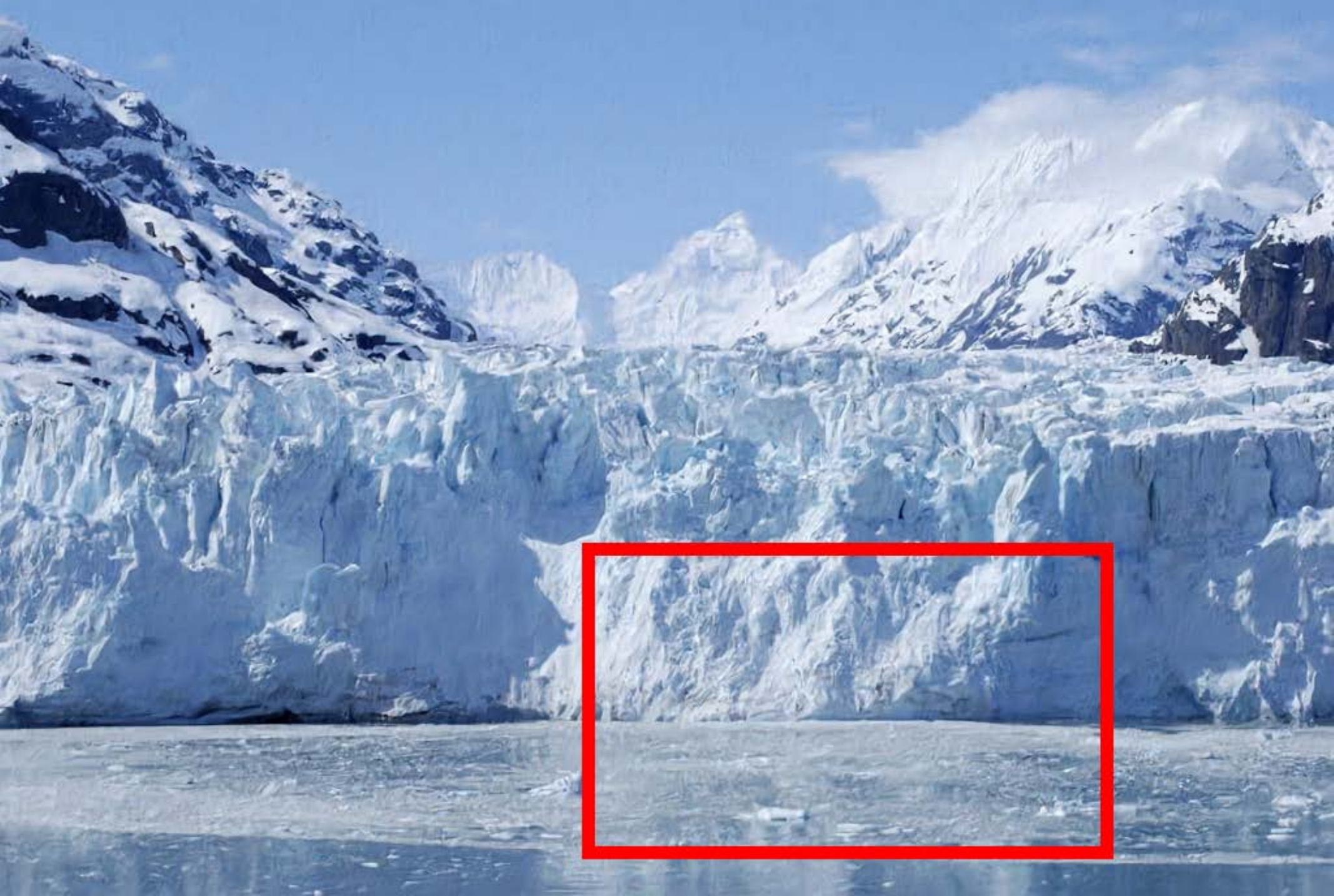}&\includegraphics[width=0.13\linewidth]{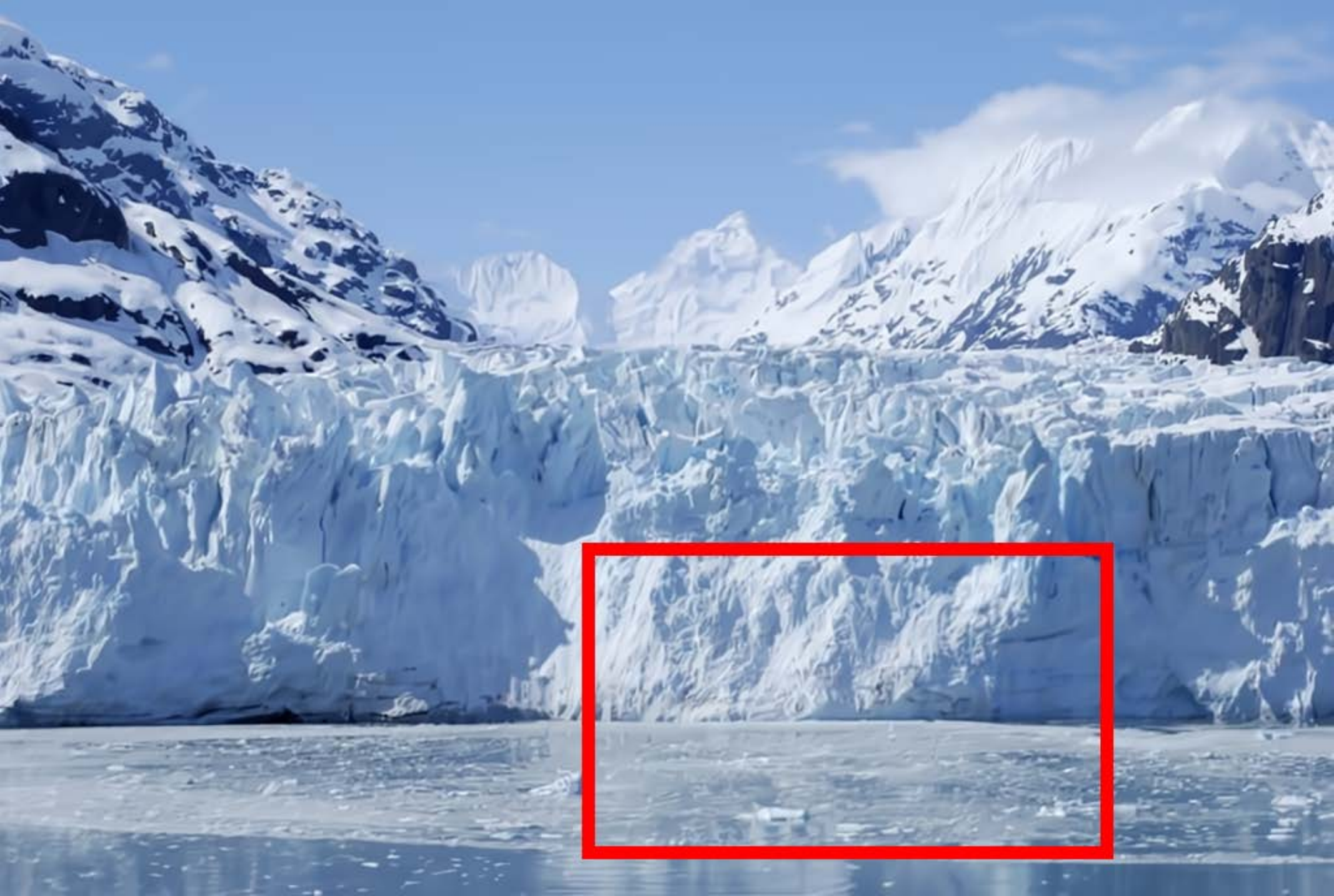} & \includegraphics[width=0.13\linewidth]{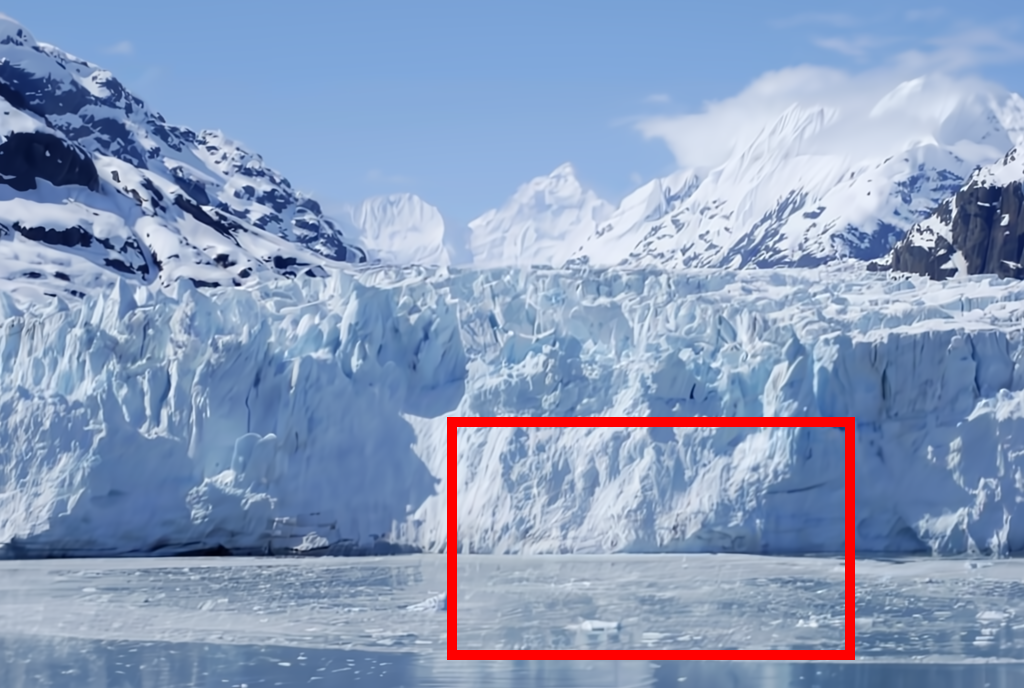} \\ 
			%\specialrule{0em}{-0.5pt}{-1pt}
			\includegraphics[width=0.13\linewidth]{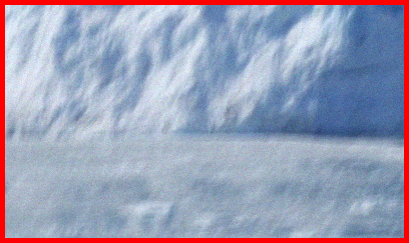}&\includegraphics[width=0.13\linewidth]{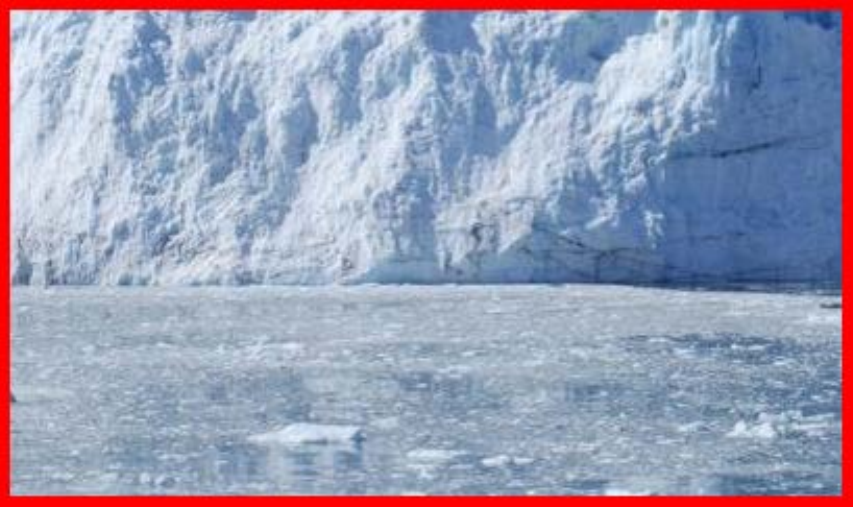}&\includegraphics[width=0.13\linewidth]{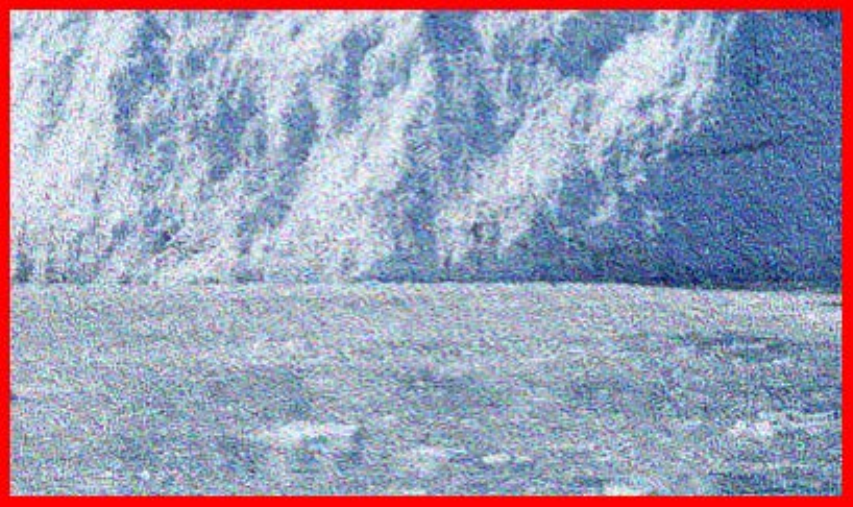}&\includegraphics[width=0.13\linewidth]{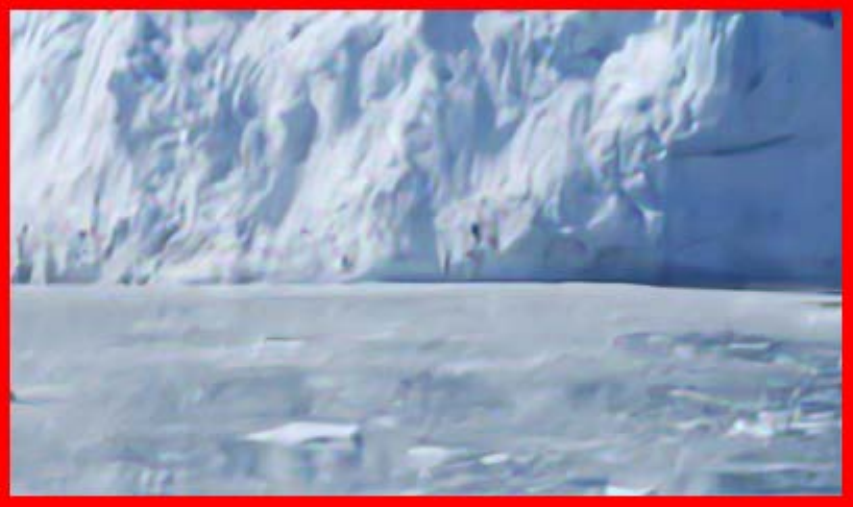}&\includegraphics[width=0.13\linewidth]{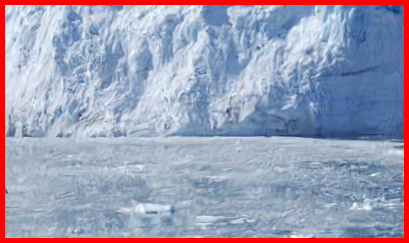}&\includegraphics[width=0.13\linewidth]{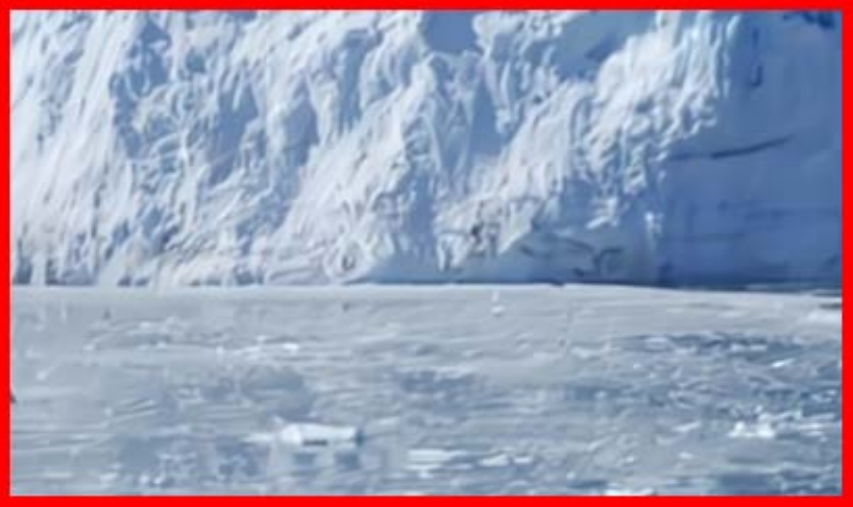} & \includegraphics[width=0.13\linewidth]{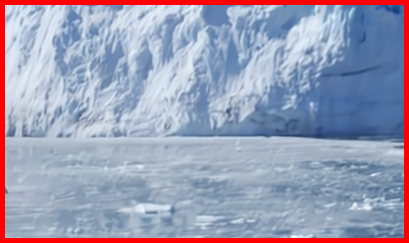} \\ \specialrule{0em}{-2pt}{-2pt} 
			\footnotesize -- & -- & \footnotesize (18.45, 0.55) & \footnotesize  (24.41, 0.83) & \footnotesize  (26.17, 0.88) & \footnotesize  (26.55, 0.88) & \footnotesize (26.61, 0.89)\\ \specialrule{0em}{-0.5pt}{-1pt} 
			\footnotesize Input & \footnotesize  Ground Truth & \footnotesize FDN & \footnotesize  IRCNN & \footnotesize IRCNN & \footnotesize  DPIR & \footnotesize Ours\\  	
		\end{tabular}
	\end{center}
	\vspace{-0.2cm}
	\caption{
		%Visual results of state-of-the-art image deblurring methods (i.e., FDN~\cite{kruse2017learning}, IRCNN~\cite{zhang2017learning}, IRCNN$+$ and DPIR~\cite{yang2019joint}).  Our method maintains a more realistic, natural and clear restoration result in terms of tone and texture. Two well-known metrics (PSNR, SSIM) are listed to quantify the generated image quality. Zoom-in regions are used to illustrate the visual differences. The best result is in red whereas the second best one is in blue. 
		Visual results of state-of-the-art image deblurring methods. Our method maintains a more realistic, natural and clear restoration result in terms of tone and texture. Two well-known metrics (PSNR$\uparrow$, SSIM$\uparrow$) are listed to quantify the generated image quality. 	
}
	\label{fig:deblur}\vspace{-0.2cm}
\end{figure*}

%\subsubsection{Style Transfer} 	
\subsubsection{Style Transfer}

We select the CycleGAN model~\cite{zhu2017unpaired} as the foundation for our architecture and conduct experiments on the FFHQ dataset~\cite{karras2019style} to explore unsupervised style transfer. Our approach leverages two Generator networks and two Discriminator networks, forming a bidirectional LwCL framework. This framework incorporates cycle-consistency loss in two loops to assess the ability to reconstruct an image from its transformed counterpart. 
The cycle consistency loss, denoted as $\mathcal{L}_{cyc}(G_1,G_2)$, captures the discrepancy between the original image and its reconstructed version in both the forward and backward mapping processes. Mathematically, it is defined as the summation of the $\ell_1$ norm of the difference between the transformed and reconstructed images, computed as $\mathcal{L}_{cyc}(G_1,G_2)=E_{\mathbf{x}\sim P_{data}(\mathbf{x})}\left[\|G_2(G_1(\mathbf{x}))\|_1\right]+E_{\mathbf{y}\sim P_{data}(\mathbf{y})}\left[\|G_1(G_2(\mathbf{y}))\|_1\right].$  
In the bidirectional mapping process, the optimization of variables in the two objective functions can be mutually exchanged and modeled. We ensure a fair comparison by following the experimental setups and model architecture detailed in \cite{zhu2017unpaired}.

Fig.~\ref{fig:real_transfer} visually demonstrates the remarkable advantages of our proposed framework in generating samples of higher quality and richer detailed textures. In comparison to the sparse and unrealistic textures produced by the standard CycleGAN, it is evident that the network integrated with our LwCL framework can generate zebra stripes that are more abundant, natural, and realistic. 
In Fig.~\ref{fig:real_transfer_epoch}, we observe significant fluctuations in the visual perceptual quality of the generated images by the standard CycleGAN throughout the training process. With the standard CycleGAN, high-quality zebra images can be generated at 40 epochs. However, at 60 epochs, a small amount of background stripes starts to appear, and by 120 epochs, the stripes that should be generated on the horse are completely transferred to the background. In contrast, when our LwCL strategy is incorporated, the generation of zebra patterns becomes gradually stable, and the authenticity of the texture is significantly enhanced. This improvement effectively mitigates the occurrence of mode collapse, ensuring the preservation of desirable characteristics in the generated images.

\begin{table}[htbp]
	\begin{center}  \footnotesize
		\renewcommand{\arraystretch}{1.2}
		\caption{Quantitative results (DE$\uparrow$, LOE$\downarrow$) on the Darkface dataset. The best result is in red whereas the second best one is in blue.}
		\label{tab:darkface}%
		\vspace{-0.2cm}
		\setlength{\tabcolsep}{3mm}{
			\begin{tabular}{|c|c|c|c|c|}
				\hline  
				Metric & RetinexNet & DeepUPE  & KinD & ZeroDCE \\
				\hline 
				\hline
				DE$\uparrow$  & 7.125 & 7.089 & 7.053 & 7.150 \\
				LOE$\downarrow$  & 555.990 & 144.242 & 277.523 & 142.544  \\
				\hline 
				\hline
				Metric & FIDE & DRBN& SCI & Ours \\
				\hline 
				\hline
				DE$\uparrow$ & 6.106 &\textcolor{blue}{\textbf{7.187}} &7.005 &\textcolor{red}{\textbf{7.254}} \\ 
				LOE$\downarrow$ & 280.616 &495.748 &\textcolor{blue}{\textbf{84.745}} &\textcolor{red}{\textbf{78.976}} \\
				\hline
				%BRC & - & -& - & - & - & - & - & - & w/o & w/ & w/ \\ 			
			\end{tabular}%
		}
	\end{center}
	\vspace{-0.2cm}	
\end{table}%

%\subsubsection{Complex Visual Perception Applications}
\subsubsection{Imitation Learning}

In the formulated Markov decision process, denoted as $\mathcal{M}=(\mathcal{S}, \mathcal{A}, \rho, P, \mathcal{R}, \gamma)$, we define the action space $\mathcal{A}$, state space $\mathcal{S}$, and reward function $\mathcal{R}: \mathcal{S} \times \mathcal{A} \mapsto \mathbb{R}$. The initial state $s_{0}$ is drawn from the distribution $\rho(\cdot)$, and the discount factor $\gamma \in(0,1)$ is applied. 
In this context, the actor $\pi$ interacts with the environment to learn the state-action value function $Q^{\pi}(s,a)$, followed by the update of the actor $\pi$ based on $Q^{\pi}(s,a)$. The objective of the policy is to maximize the expected discounted cumulative reward, given by: $
\pi^*=\arg\max_{\pi}\mathbb{E}_{s_0\sim p_0, a_0\sim \pi}\left(Q^{\pi}(s_0,a_0)\right),\label{eq:eqac1}
$ where $s_0$, $a_0$, and $p_0$ represent the initial state, initial action, and initial state distribution, respectively. For our experiment, we employ the PyBullet physics simulator and publicly available datasets tailored for data-driven deep reinforcement learning. Following the definitions and experimental settings of recent studies~\cite{fu2017learning,arulkumaran2021pragmatic}, we design an agent with an actor-critic structure to predict actions that deceive the discriminator. The discriminator, on the other hand, is trained to distinguish between samples generated by the policy $\pi^{*}$ and an expert policy $\hat{\pi}$. To compute the reward function $\mathcal{R}$, we adopt the form $h(s, a)=\log (D(s, a))-\log (1-D(s, a)).$  Additionally, both Generative Adversarial Imitation learning (GAIL) and our method incorporate the $R_{1}$ gradient penalty regularizers.

Fig.~\ref{fig:gail} presents the average policy return and standard error for two simulated environments, namely  ``Walker2D'' and ``Hopper''. The average policy return and its standard error are plotted to illustrate the performance of GAIL. It is evident that the GAIL curve exhibits significant instability and lacks a convergent trend. In contrast, our proposed solution technique ensures a more stable convergence during training with reduced deviation. Additionally, we provide the final return achieved throughout the training episodes, which serves as an indicator of the performance improvement achieved by GAIL when employing our novel solution techniques.

\subsection{ART-type Applications} 
%In the following, we explore three applications with introducing related task constraints, i.e., medical image segmentation, low-light image enhancement and hyper-parameter learning. 
In the subsequent sections, we delve into three distinct applications, each accompanied by the introduction of relevant task constraints. These applications encompass medical image analysis, low-light image enhancement, and hyper-parameter learning. 
%By incorporating these task constraints into our framework, we aim to enhance the performance and address specific challenges associated with these tasks.

\subsubsection{Medical Image Analysis} \label{sec:mis}
We evaluate the segmentation performance on a hybrid dataset by using 426 mixed medical scans, which is sampled from three standard datasets, ABIDE~\cite{di2014autism}, ADNI~\cite{mueller2005ways} and PPMI~\cite{marek2011parkinson}. During the training phase, the scanned images are divided into 346, 40, and 40 volumes for training, validation, and testing, respectively. To capture both global and local differences in appearance, we define the energy function $\mathcal{L}_{reg}=\mathcal{L}_{sim}+\lambda_{scc}*\mathcal{L}_{scc}$, where $\mathcal{L}_{sim}$ is the multi-scale local cross-correlation in appearance, and $\mathcal{L}_{scc}$ is the semantic content consistency loss, i.e.,  $\mathcal{L}_{scc}=\frac{1}{2}\big(\mathtt{KL}(p_t||\frac{p_w+p_t}{2})+\mathtt{KL}(p_w||\frac{p_w+p_t}{2})\big)$. Here $\mathtt{KL}$ denotes the Kullback-Leibler divergence, and $p_w$ and $p_t$ are the warped prediction and target prediction, respectively. For segmentation, the energy function is defined as a hybrid loss $\mathcal{L}_{seg}=\mathcal{L}_{dice}+\lambda_{mce}*\mathcal{L}_{mce}$ composed of multi-class cross entropy loss $\mathcal{L}_{mce}$ and Dice coefficient loss $\mathcal{L}_{dice}$. In our training, we empirically set the balancing parameters $\lambda_{scc}=10$ and $\lambda_{mce}=1$, and use ADAM optimizer with learning  rate of $4*10^4$. 
To evaluate the performance of registration and segmentation, we employed well-established metrics such as the Dice score, Hausdorff distance (HD95), and average surface distance (ASD). We compare our method to several state-of-the-art methods, including 
a) Deep learning registration-based segmentation
methods, VxM\cite{DBLP:journals/tmi/BalakrishnanZSG19}, LKU-Net\cite{DBLP:conf/miccai/JiaBZLQD22}, and TransMorph\cite{DBLP:journals/mia/ChenFHSLD22}. b) Deep learning segmentation methods, 
UNet\cite{DBLP:conf/miccai/CicekALBR16}, MASSL\cite{DBLP:conf/miccai/ChenBJTB19}, and CPS\cite{DBLP:conf/cvpr/ChenYZ021}. d) joint registration and segmentation methods, SST\cite{DBLP:conf/wacv/TomarBLVRT22}, DeepAtlas\cite{DBLP:conf/miccai/XuN19}, DataAug\cite{DBLP:conf/cvpr/ZhaoBDGD19}, and BRBS\cite{he2022learning}. 

Tab.~\ref{tab:regseg} presents quantitative comparisons for joint registration and segmentation tasks, demonstrating that our method achieved the highest Dice score and the lowest HD95 and ASD metrics in registration and segmentation, respectively. 
Fig.~\ref{fig:mis-0} illustrates the two-dimensional visualization results of the registration method compared to other approaches. The framework solely relying on registration and segmentation tends to exhibit more misalignment errors along anatomical boundaries. In contrast, our method demonstrates the smallest mislabeling regions on the lateral ventricle (LV) and brainstem (BS), as indicated by the yellow and pink areas. Additionally, we provide qualitative results in Fig.~\ref{fig:mis-1}, illustrating the robust segmentation performance of our method at complex termination sites in the structural white matter of the brain and the finer segmentation quality achieved on the cerebellar tissue and the 3/4 ventricles.

%\subsubsection{Low-light Enhancement}  
\subsubsection{Low-light Image Enhancement}

We conduct low-light enhancement experiments on DarkFace~\cite{yang2020advancing} dataset, and adopt the well-known no reference metrics (i.e., DE and LOE).  To benchmark our method, we compare it against several state-of-the-art approaches, including MBLLEN~\cite{lv2019attention},  RetinexNet~\cite{Chen2018Retinex},  KinD~\cite{zhang2019kindling}, ZeroDCE~\cite{guo2020zero}, DeepUPE~\cite{wang2019underexposed}, FIDE~\cite{xu2020learning}, DRBN~\cite{yang2020fidelity} and SCI~\cite{ma2022toward}. As mentioned  in~\cite{ma2022toward}, we introduce the  pixel fidelity term $\mathcal{L}_p$ and a smoothness term $\mathcal{L}_s$ for $\mathcal{N}_{\bm{\theta}}^{O}$, which are formulated as  $\mathcal{L}_p=\sum_{t=1}^{T}\|\mathbf{x}^t-\hat{\mathbf{y}}_{t-1}\|$, $\mathcal{L}_s=\sum_{i=1}^{N}\sum_{j\in \mathcal{N}(i)}w_{i,j}\|\mathbf{x}_{i}^t-\mathbf{x}_{j}^t\|$, where $\hat{\mathbf{y}}_{t-1}$ denotes the self-calibrated variable, $N$ is the total number of pixels, $w_{i,j}$ represents the weight function. As for $\mathcal{N}_{\bm{\omega}}^{C}$, we also introduce the anchor-based multi-task loss and progressive anchor loss~\cite{li2019dsfd}, defined as: $\mathcal{F}^{\bm{\theta}}_{\mathtt{CL}}:=\mathcal{L}_{SSL}(a)+\lambda \mathcal{L}_{SSL}(sa)$. 
%Here, a smaller anchor loss with $sa_i$ is introduced.  
Here, $\mathcal{L}_{SSL}(p_i,p_i^*,t_i,g_i,a_i)=\frac{1}{N_{conf}}((\sum_iL_{conf}(p_i,p_i^*))+\frac{\beta}{N_{loc}}\sum_ip_i^*L_{loc}(t_i,g_i,a_i)),$ where
$N_{conf}$  and $N_{loc}$ indicate the number of positive and negative anchors, and the number of positive anchors, respectively. $L_{loc}$ is the smooth loss between the predicted box $t_i$ and ground-truth box $g_i$ using the anchor $a_i$, and $L_{conf}$ is the softmax loss in terms of two classes.

\begin{table}[htbp]
	%\vspace{-0.4cm} %
	\centering
	\caption{Reporting results of existing methods for solving data hyper-cleaning tasks. F1 score denotes the harmonic mean of the precision and recall.}\label{tab:hyper_cleaning}
	\renewcommand\arraystretch{1.2}
	\setlength{\tabcolsep}{0.7mm}{
		\begin{tabular}{|c|c|c|c| c|c|c| }
			\hline
			\multirow{2}{*} { Method } & \multicolumn{2}{c|} { MNIST } & \multicolumn{2}{c|} { FashionMNIST }  & \multicolumn{2}{c|} { CIFAR10 }\\
			\cline { 2 - 7}
			&   F1 score& Time (s)  &  F1 score &Time (s) &  F1 score& Time (s)\\
			\hline
			\hline
			CG  & $85.96$ & $0.1799$  & $85.13$ &$0.2041$  &$69.10$& $0.4796$\\
			%\hline
			Neumann & $87.54$   &{\color{blue}$\mathbf{0.1723}$}  & $87.28$ &{\color{blue}$\mathbf{0.1958}$}   &$68.87$&{\color{blue}$\mathbf{0.4694}$}\\
			%\hline
			RHG & $89.36$&$0.4131$  & $87.12$  & $0.4589$ &$68.27$&$1.3374$\\
			%\hline
			T-RHG  & $89.77$ &$0.2623$  & $86.76$ & $0.2840$  &$68.06$&$0.8409$\\
			%\hline
			BDA   & {\color{blue}$\mathbf{90.38}$} & $0.6694$ & {\color{blue}$\mathbf{88.24}$}  &$0.8571$ &{\color{blue}$\mathbf{67.33}$}&$1.4869$\\
			%\cdashline{1-10}[1.5pt/2pt] 			
			%IAPTT& {\color{red}$\mathbf{90.88}$} & {\color{blue}$\mathbf{91.57}$}&$0.1948$& {\color{blue}$\mathbf{83.67}$} & {\color{blue}$\mathbf{90.37}$} &{\color{blue}$\mathbf{0.2032}$}&{\color{blue}$\mathbf{37.16}$}&{\color{blue}$\mathbf{71.57}$}&$0.6727$\\
			\hline
			%BRC&{\color{blue}$\mathbf{90.41}$}&{\color{red}$\mathbf{91.95}$}&$0.2086$&{\color{red}$\mathbf{83.80}$}&{\color{red}$\mathbf{90.40}$}&$0.2105$&{\color{red}$\mathbf{37.70}$}&{\color{red}$\mathbf{72.74}$}&$0.6776$\\
			Ours  & {\color{red}$\mathbf{91.41}$} & {\color{red}$\mathbf{0.0279}$}   & {\color{red}$\mathbf{90.03}$} & {\color{red}$\mathbf{0.0289}$} &{\color{red} $\mathbf{70.10}$} & {\color{red}$\mathbf{0.0797}$}\\
			\hline
		\end{tabular}
	}
	%\vspace{-0.4cm}
\end{table}

In Fig.~\ref{fig:real_darkface}, we compare the visualization results. It can be seen that although some methods can successfully enhance the brightness of the image, none of them can restore the clear image texture. The DE score is reported below, and a higher DE value indicates better visual quality. In comparison, our method produces the most visually pleasing results, can not only learns to enhance the dark area while restoring more visible details but also avoids over-exposure artifacts. We report the quantitative results in Tab.~\ref{tab:darkface}.  It can be seen that our method numerically outperforms existing methods by large margins and ranks first across all metrics.  This further endorses the superiority of our method over current state-of-the-art methods in generating high-quality visual results.

 \begin{table}
	\footnotesize
	\caption{Comparison of quantitative results (i.e., averaged PSNR and SSIM scores) among state-of-the-art image deblurring methods (i.e., FDN~\cite{kruse2017learning}, IRCNN~\cite{zhang2017learning}, IRCNN$+$ and DPIR~\cite{zhang2021plug}). The default noise intensity $\textbf{n}$ is set to 7.65. The quality scores of the various methods are compared under different sizes of blur kernels $\textbf{K}_{i}, i=1,\cdots,5$.} 
	%Bold red and bold blue refer to top 1 $\mathrm{^{st}}$ and top 2 $\mathrm{^{nd}}$, respectively.
	% \vskip -0.9in
	%\vskip 11pt
	\renewcommand\arraystretch{1.2}
	\setlength\tabcolsep{2.8pt}
	\begin{center}
		%\begin{small}
		%   \resizebox{0.5\textwidth}{!}{
		\begin{tabular}{|c|c|c|c|c|c|c|}
			\hline 
			%Method & FDN~\cite{kruse2017learning} &IRCNN~\cite{zhang2017learning}&IRCNN$+$~\cite{zhang2017learning}&DPIR~\cite{yang2019joint}& Ours\\  
			\multirow{2}{*}{Blur Kernel}	& \multirow{2}{*}{Metric} &\multicolumn{5}{c|}{Method}\\
			\cline{3-7}
			&	 &   FDN &IRCNN &IRCNN$+$ &DPIR & Ours\\ 
			\hline  
			\hline  
			\multirow{2}{*}{$\textbf{K}_1$}  & PSNR &18.741&	26.281&	27.761&
			\textcolor{blue}{\textbf{28.168}} &	\textcolor{red}{\textbf{28.236}}\\ 				
			& SSIM &0.472&	0.876&	0.880&\textcolor{blue}{\textbf{0.889}}&\textcolor{red}{\textbf{0.890}}\\
			%\hline 
			%\hline 
			\multirow{2}{*}{$\textbf{K}_2$}  & PSNR & 17.977&	27.361&	27.532&\textcolor{blue}{\textbf{28.066}}&	\textcolor{red}{\textbf{28.132}}\\
			& SSIM & 0.437&	0.881&	0.874& \textcolor{blue}{\textbf{0.887}}&	\textcolor{red}{\textbf{0.888}}\\
			%\hline 
			
			\multirow{2}{*}{$\textbf{K}_3$}  & PSNR & 17.775&	\textcolor{red}{\textbf{28.364}}&	27.586& \textcolor{blue}{\textbf{28.295}}&	\textcolor{red}{\textbf{28.364}} 	 \\
			& SSIM & 0.431& 	\textcolor{red}{\textbf{0.895}}& 	0.868&  0.890& 	\textcolor{blue}{\textbf{0.891}} \\
			%\hline 
			%\hline 
			\multirow{2}{*}{$\textbf{K}_4$}  & PSNR & 17.905 &	25.291 &	27.400 & \textcolor{blue}{\textbf{27.918}} &	\textcolor{red}{\textbf{27.970}} \\	 				
			& SSIM & 0.435&	0.858&	0.872& \textcolor{blue}{\textbf{0.884}}&	\textcolor{red}{\textbf{0.885}} \\ 
			%\hline 
			
			% 				\multirow{2}{*}{$k_5$}  & PSNR & 18.966&	29.548&	28.089& 28.904&	\textcolor{red}{\textbf{28.986}}\\ 					 				
			% 				& SSIM & 0.488&	\textcolor{red}{\textbf{0.916}}&	0.876& 0.899&	0.901\\
			% 				\hline 
			% 				\multirow{2}{*}{$k_6$}  & PSNR & 18.786&\textcolor{red}{\textbf{29.338}}&	27.945& 28.759&	28.846\\   								 				
			% 				& SSIM &0.480&	\textcolor{red}{\textbf{0.911}}&	0.876& 0.896&0.898\\
			% 				\hline 
			% 				\multirow{2}{*}{$k_7$}  & PSNR & 18.786&\textcolor{red}{\textbf{29.338}}&	27.945& 28.759&	28.846\\   								 				
			% 				& SSIM &0.457&	0.894&	0.869& 0.894&\textcolor{red}{\textbf{0.895}}\\
			%\hline 
			\multirow{2}{*}{$\textbf{K}_5$}  & PSNR & 18.273&	27.642&	27.640& \textcolor{blue}{\textbf{28.450}}&\textcolor{red}{\textbf{28.471}}
			\\
			
			& SSIM &0.436&	\textcolor{blue}{\textbf{0.879}}&	0.866& \textcolor{red}{\textbf{0.888}}&\textcolor{red}{\textbf{0.888}}\\ 
			%			\hline 
			%			\hline				
			%			% 				\multirow{2}{*}{Average}  & PSNR & 17.878&	26.938&	27.380&	 \textcolor{blue}{\textbf{28.089}} &\textcolor{red}{\textbf{28.092}}\\
			%			% 				
			%			% 				& SSIM &  0.455&0.889&0.873&\textcolor{blue}{\textbf{0.891}}&\textcolor{red}{\textbf{0.892}} \\
			%			\multirow{2}{*}{Average}  & PSNR & 17.878&	26.987&	27.380&	 \textcolor{blue}{\textbf{28.179}} &\textcolor{red}{\textbf{28.234}}\\
			%			
			%			& SSIM &  0.455&0.877&0.873&\textcolor{blue}{\textbf{0.887}}&\textcolor{red}{\textbf{0.888}} \\
			\hline
		\end{tabular}
		%   }
		%	\end{small}
	\end{center}
	\label{tab:deblur}
	%\vskip 11pt
\end{table}

\begin{table*}[t]
	\caption{Mean test accuracy of various methods (model-based methods and gradient-based bi-level methods) on few-shot classification classification problems (1-shot and 5-shot, i.e., $M=1,5$, $N=5,20,30,40$) on Omniglot. We use $\pm$ to represent $95\%$ confidence intervals over tasks.}
	\label{tab:omniglot}
	\centering  
	% \vspace{0.5em}
	\renewcommand\arraystretch{1.2}
	% \vskip 0.12in
	\setlength{\tabcolsep}{0.3mm}{
		\begin{tabular}{|c|c|c|c|c|c|c|c|c|}
			\hline
			\multirow{2}{*}{Method}& \multicolumn{2}{c|}{ $5$-way}&\multicolumn{2}{c|}{$20$-way}&\multicolumn{2}{c|}{$30$-way} &\multicolumn{2}{c|}{$40$-way}\\
			\cline{2-9}
			&$1$-shot & $5$-shot &  $1$-shot &  $5$-shot & $1$-shot &  $5$-shot & $1$-shot &  $5$-shot\\
			\hline\hline
			MAML &98.70 $\pm$ $0.40\%$&{\color{red}$\mathbf{ 99.91\pm 0.10}\%$} &95.80 $\pm$ $0.30\%$ &98.90 $\pm$ $0.20\%$ & 86.86 $\pm\ 0.49\%$ & 96.86 $\pm\ 0.19\%$ & 85.98 $\pm\ 0.45\%$& 94.46 $\pm\ 0.13\%$  \\
			%\hline
			Meta-SGD  & 97.97 $\pm\ 0.70\%$ & 98.96 $\pm\ 0.20\%$ & 93.98 $\pm\ 0.43\%$& 98.42 $\pm\ 0.11\%$ & 89.91 $\pm\ 0.04\%$ & 96.21 $\pm 0.15\%$ & 87.39 $\pm\ 0.43\%$ & 95.10 $\pm\ 0.15\%$\\
			%\hline
			Reptile & 97.68 $\pm\ 0.04\%$ & 99.48 $\pm\ 0.06\%$ & 89.43 $\pm\ 0.14\%$ & 97.12 $\pm\ 0.32\%$ & 85.40 $\pm\ 0.30\%$ & 95.28 $\pm\ 0.30\%$ & 82.50 $\pm\ 0.30\%$ & 92.79 $\pm\ 0.33\%$\\
			%\hline
			iMAML  & {\color{blue}$\mathbf{ 99.16\pm 0.35}\%$} & 99.67 $\pm$  0.12$\% $ & 94.46 $\pm$ 0.42$\%$ & 98.69 $\pm \ 0.10\%$ & 89.52 $\pm\ 0.20\%$ & 96.51 $\pm\ 0.08\%$ & 87.28 $\pm\ 0.21\%$ & 95.27 $\pm\ 0.08\%$ \\
			%\hline\hline
			RHG & 98.64 $\pm$ $ 0.21\%$ & 99.58 $\pm$ $ 0.12\%$ & {\color{blue}$\mathbf{ 96.13\pm 0.02}\%$} & {\color{red}$\mathbf{ 99.09\pm 0.08}\%$}  & 93.92 $\pm\ 0.18\%$ & {\color{blue}$\mathbf{ 98.43\pm 0.08}\%$}  & {\color{blue}$\mathbf{ 90.78\pm 0.20}\%$} & {\color{blue}$\mathbf{ 96.79\pm 0.10}\%$} \\
			%\hline
			T-RHG  & 98.74 $\pm$ $ 0.21\%$& {\color{blue}$\mathbf{ 99.71\pm 0.07}\%$} & 95.82 $\pm$ $ 0.20\%$ & 98.95 $\pm$ $ 0.07\%$& {\color{blue}$\mathbf{ 94.02\pm 0.18}\%$} & 98.39 $\pm\ 0.07\%$ & 90.73 $\pm\ 0.20\%$  & {\color{blue}$\mathbf{ 96.79\pm 0.10}\%$}  \\
			%\hline
			%BDA & 99.04 $\pm$ $ 0.18\%$ & 99.74 $\pm$ $ 0.05\%$ & \textbf{96.50} $\pm$ $ 0.16\%$ & \textbf{99.19} $\pm$ $ 0.07\%$ & \textbf{94.37} $\pm\ 0.18\%$ & \textbf{98.53} $\pm\ 0.07\%$ & \textbf{92.49} $\pm\ 0.18\%$ & \textbf{97.12} $\pm$ 0.09 $\%$\\
			\hline
			Ours & {\color{red}$\mathbf{ 99.67\pm 0.05}\%$} &99.67 $\pm$ $ 0.06\%$ & {\color{red}$\mathbf{ 96.33\pm 0.21}\%$} & {\color{blue}$\mathbf{ 99.07\pm 0.11}\%$} & {\color{red}$\mathbf{ 95.53\pm 0.25}\%$} &{\color{red}$\mathbf{ 98.50\pm 0.14}\%$} & {\color{red}$\mathbf{ 91.21\pm 0.28}\%$} & {\color{red}$\mathbf{ 97.33\pm 0.16}\%$}\\
			\hline
		\end{tabular}
	}
\end{table*}

\subsubsection{Hyper-parameter Learning} 
We consider a specific data hyper-cleaning example~\cite{liu2022towards}. In this scenario, we aim to train a linear classifier using a given image dataset, but encounter the issue of corrupted training labels. To address this, we adopt softmax regression with parameters $\bm{\omega}$ as our classifier and introduce hyperparameters $\bm{\theta}$ to assign weights to the training samples. 
Initially, we define the cross-entropy function  $\ell(\bm{\omega};\mathbf{u}_i,\mathbf{v}_i)$ which measures the classification loss using the classification parameter $\bm{\omega}$ and the data pairs  $(\mathbf{u}_i,\mathbf{v}_i)$. The training and validation sets are denoted as 
 $\mathcal{D}_{\mathtt{tr}}$ and $\mathcal{D}_{\mathtt{val}}$, respectively. Next, we introduce the CL function $\mathcal{F}^{\bm{\theta}}_{\mathtt{CL}}$ as the following weighted training loss, given by 
$\mathcal{F}^{\bm{\theta}}_{\mathtt{CL}}(\bm{\omega})=\sum_{(\mathbf(u)_i,\mathbf(v)_i)\in \mathcal{D}_{\mathtt{tr}}}[\delta(\bm{\theta})]_i\ell(\bm{\omega};\mathbf{u}_i,\mathbf{v}_i)$. Here $\bm{\theta}$ represents the hyperparameter vector that penalizes the objective for different training samples.  The element-wise sigmoid function  $\delta(\bm{\theta})$ is applied to restrict the weights within the range of $[0,1]$.  Furthermore, we define $\mathcal{F}_{\mathtt{OL}}$ as the cross-entropy loss with $\ell_2$ regularization on the validation set, i.e., $\mathcal{F}_{\mathtt{OL}}(\bm{\theta})=\sum_{(\mathbf(u)_i,\mathbf(v)_i)\in \mathcal{D}_{\mathtt{val}}}\ell(\bm{\omega}(\bm{\theta});\mathbf{u}_i,\mathbf{v}_i)+\eta||\bm{\omega}(\bm{\theta})||^2$, where $\eta$ is the trade-off parameter. 
%We applied our algorithm with the baselines among hyper-parameter optimization methods, CG~\cite{pedregosa2016hyperparameter}, Neumann~\cite{lorraine2020optimizing}, RHG~\cite{franceschi2017forward}, Truncated RHG (T-RHG)~\cite{shaban2019truncated}, and BDA~\cite{liu2021general} to solve the above problem. 
Three well known datasets including MNIST, FashionMNIST and CIFAR10 are used to conduct the experiments.  Specifically, the training, validation and test sets
consist of 5000, 5000, 10000 class-balanced samples randomly selected to construct $\mathcal{D}_{\mathtt{tr}}$, $\mathcal{D}_{\mathtt{val}}$ and $\mathcal{D}_{\mathtt{test}}$, then half of the labels in $\mathcal{D}_{\mathtt{tr}}$ are tampered.  We adopted the architectures proposed by~\cite{franceschi2017forward} as the feature extractor for all the compared methods. 
%For T-RHG, we chose 25-step truncated back-propagation to guarantee its convergence. 

Tab.~\ref{tab:hyper_cleaning} presents a comprehensive comparison of our LwCL framework with various hyperparameter optimization methods, including CG~\cite{pedregosa2016hyperparameter}, Neumann~\cite{lorraine2020optimizing}, RHG~\cite{franceschi2017forward}, Truncated RHG (T-RHG)~\cite{shaban2019truncated}, and BDA~\cite{liu2021general} in terms of F1 score and running time.  The results clearly demonstrate the superior performance of our method in terms of F1 score compared to the other hyperparameter optimization techniques. Notably, our method significantly outperforms all relevant algorithms in terms of running time, achieving a substantial improvement of nearly an order of magnitude.

\subsection{TDC-type Applications}
%As described in Sec.~\ref{sec:tdc},  we follow the idea of divide-and-conquer, which allows us to flexibly decompose a complete learning task into a coupled learning process with objective learner and constraint learner, whereby we conduct the following experiments, i.e., \textit{multi-tasking and meta-learning and image deblurring}. 

%Following the idea of divide-and-conquer, we conduct the following experiments, i.e.,  multi-tasking and meta-learning and image deblur. 
%
%In the following, we further demonstrate the performance of LwCL with two typical learning and vision applications, including multi-tasking and meta-learning and image deblurring. 
In the subsequent sections, we proceed to showcase the efficacy of LwCL through the evaluation of its performance in two prominent TDC-type LwCL  applications:  image deblurring and multi-task meta-learning.

\subsubsection{Image Deblurring} 
We conduct image deblurring experiment on a data benchmark, containing 400 images from the Berkeley Segmentation dataset, 4744 images from the Waterloo Exploration database, 900 images from the DIV2K dataset and 2750 images from the Flick2K dataset. More specifically, we used the DRUNet in DPIR~\cite{zhang2021plug} containing four scales as the base network. In the specific implementation, the subproblem for $\textbf{u}$ is solved by closed-form solution based on the fast Fourier transform, and the subproblem on $\textbf{z}$  is obtained by an updated denoiser $\mathtt{Net}_{\bm{\theta}}(\textbf{u})$. Unlike the original method that treats the denoiser as a fixed pre-trained network, under our LwCL framework, the parameters of the deblurring network are dynamically updated as learnable variables. As discussed earlier, it can be understood that the lower variable $\bm{\omega}$ is  a combination about two optimized variables, i.e., $\bm{\omega}=\{\textbf{u}, \textbf{z}\}$.  Thus, by jointly learning two sub-tasks, our method can improve robustness and generalization to various complex noise scenarios. Each scale has a skip connection between the 2 × 2 stride convolution downscale and the 2 × 2 transpose convolution upscale operations. The number of channels per layer from the first to the fourth scale is 64, 128, 256 and 512 respectively. Four consecutive blocks of residuals are used in the down-sampling operation and up-sampling operation at each scale. 

We qualitatively and quantitatively evaluated the performance of a series of relevant methods, including  FDN~\cite{kruse2017learning}, IRCNN~\cite{zhang2017learning}, IRCNN$+$ and DPIR~\cite{zhang2021plug}). 
As shown in Tab.~\ref{tab:deblur}, we report the performance of the current methods under five different sizes of blur kernel settings (i.e., $\textbf{K}_{i}, i=1,\cdots,5$). All experiments were performed under a uniform noise criterion with a default noise level  $\textbf{n}=7.65$. In comparison, our method achieves the best PSNR scores under all five blurs and performs best in both average PSNR and SSIM scores. In addition, we show a visualization of the perceptual results in Fig.~\ref{fig:deblur}. As can be seen, for the deblurring task, our method outperforms other methods in terms of color recovery, detail retention and the quantitative metric PSNR, and achieves the best visual performance.  This further endorses the superiority of our method over current state-of-the-art methods in image deblurring.

\subsubsection{Multi-task Meta-learning} 
%Meta-learning aims to exploit a large number of similar few-shot tasks to learn an algorithm that should perform well on new tasks. 
We conduct the $N$-way $M$-shot classification experiments where each task is to discriminate $N$ separate classes and it is to learn the hyper-parameter $\bm{\theta}$ such that each task can be solved only with $M$ training samples. Typically, we separate the network architecture into two parts: the  cross-task intermediate representation layers (parameterized  by $\bm{\theta}$) outputs the meta features and the multinomial logistic regression layer (parameterized by $\bm{\omega}^j$ ) as our ground classifier for the $j$-th task. During training, we conduct our experiment on a meta training data set $\mathcal{D}=\{\mathcal{D}^j\}$, where $\mathcal{D}^j= \mathcal{D}_{\mathtt{tr}}^j \bigcup \mathcal{D}_{\mathtt{val}}^j$ is linked to the $j$-th task. Then, we consider the cross-entropy function $\ell(\bm{\theta},\bm{\omega}^j;\mathcal{D}_{\mathtt{tr}})$ as the task-specific loss for the $j$-th task and thus the $\mathcal{F}^{\bm{\theta}}_{\mathtt{CL}}$ can be defined as $\mathcal{F}^{\bm{\theta}}_{\mathtt{CL}}(\{\bm{\omega}^j\})=\sum_{j}\ell(\bm{\theta},\bm{\omega}^j;\mathcal{D}_{\mathtt{tr}}^j)$. Similarly, we also utilize the cross-entropy function but define it based on $\mathcal{D}_{\mathtt{val}}^j$ as $\mathcal{F}^{\{\bm{\omega}^j\}}_{\mathtt{OL}}(\bm{\theta})=\sum_{j}\ell(\bm{\theta},\bm{\omega}^j;\mathcal{D}_{\mathtt{val}}^j)$. We validate the performance based on the widely used  Omniglot~\cite{lake2015human} dataset, and consider ResNet-12 with Residual blocks as the backbone. Besides, we introduce the task-and-layer-wise attenuation~\cite{baik2020learning} to control the influence of prior knowledge for each task and layer.  

As illustrated in Tab.~\ref{tab:omniglot}, we followed the experimental protocol~\cite{collins2020task} and compared our algorithm to several state-of-the-art approaches, such as MAML~\cite{collins2020task}, Meta-SGD~\cite{li2017meta}, Reptile~\cite{nichol2018first}, iMAML~\cite{rajeswaran2019meta}, RHG~\cite{franceschi2017forward}, and T-RHG~\cite{shaban2019truncated}. In comparison, our LwCL achieved the highest classification accuracy except in the 5-way 5-shot and 20-way 5-shot tasks. Indeed, with more complex few-shot classification problems (such as 30-way and 40-way), our LwCL showed significant advantages over other methods.

\section{Conclusions and Future Works} \label{sec:sec5}

In this work, we have introduced a novel perspective called Learning with Constraint Learning (LwCL) to provide a deeper understanding of their underlying coupling mechanisms for efficiently solving contemporary complex problems in machine learning and computer vision. Our proposed framework provides a unified understanding of the intrinsic mechanisms behind diverse problems. By establishing a general hierarchical optimization framework and a dynamic best response-based fast solution strategy, we have demonstrated the effectiveness of our approach in formulating and addressing LwCLs. Through extensive experiments we have verified the efficiency of our proposed framework in solving a wide range of LwCL problems, spanning three categories and nine different types.  Future research can focus on further exploring and extending the capabilities of LwCL in addressing even more challenging problems and advancing the state-of-the-art in machine learning and computer vision.
The findings presented in this paper contribute to a deeper understanding and efficient resolution of complex problems in learning and vision, providing valuable insights for future research and applications in the field.

	\ifCLASSOPTIONcompsoc
	\section*{Acknowledgments}
	\else
	\section*{Acknowledgment}
	\fi

	This work is partially supported by the National Key R\&D Program of China (2022YFA1004101), and the National Natural Science Foundation of China (No. U22B2052).
	
	\ifCLASSOPTIONcaptionsoff
	%\newpage
	\fi

	\bibliographystyle{IEEEtran}
	\bibliography{reference}

	%	Biography text here.
	%\end{IEEEbiography}
	%	Biography text here.
	%% insert where needed to balance the two columns on the last page with
	%%\newpage
	%	Biography text here.
	
	
	%\vfill
	

\end{document}